\documentclass[12pt,PhD]{muthesis}

\usepackage[pdftex]{graphicx}
\usepackage[pdftex,colorlinks=false]{hyperref}
\usepackage{amsmath}

\usepackage{verbatim}
\usepackage{listings} 
\usepackage{epic,ecltree} 
\usepackage{url} 

\usepackage{pdftricks}
\begin{psinputs}
\usepackage{pst-all}
\usepackage{epsfig} 
\end{psinputs}

\usepackage{tabls}
\usepackage{amssymb}
\usepackage[active]{srcltx}

\begin{document}

\title{Complex-valued Convolutional Neural Network Classification of Hand Gestures from Radar Images}
\author{Shokooh Khandan}
\principaladviser{Fumie Costen}

\beforeabstract
\prefacesection{Abstract}
 Hand gesture recognition systems have yielded many exciting advancements in the last decade and become more popular in HCI (human-computer interaction) with several application areas, which spans from safety and security applications to automotive field \cite{g9841948}. 

Various deep neural network architectures have already been inspected for hand gesture recognition systems, including multi layer perceptron (MLP) \cite{gshasht}, convolutional neural network (CNN) \cite{gstwl}, recurrent neural network (RNN) \cite{gstten} and a cascade of the last two architectures known as CNN-RNN \cite{gsften}.

However a major problem still exists, which is most of the existing ML algorithms are designed and developed the building blocks and techniques for real-valued (RV). Researchers applied various RV techniques on the complex-valued (CV) radar images, such as converting a CV optimisation problem into a RV one, by splitting the complex numbers into their real and imaginary parts. However, the major disadvantage of this method is that, the resulting algorithm will double the network dimensions. 

Recent work on RNNs and other fundamental theoretical analysis
suggest that, CV numbers have a richer representational capacity, but due to the absence of the building blocks required to design such models, the performance of CV networks are marginalised.

In this report, first we review the background of ML and artificial neural networks (ANNs) in chapter two,  then in the third chapter, we explain the characteristics of our utilised two sets of CV datasets. In the forth chapter,we propose a fully CV-CNN, including all building blocks, forward and backward operations, and derivatives all in complex domain. Then we implement the designed model in Python from scratch and fully in complex domain. We explore the proposed classification model on two sets of CV hand gesture radar images in comparison with the equivalent RV model.

In chapter five, we propose a CV-forward residual network, for the purpose of binary classification of the two sets of CV hand gesture radar datasets. We demonstrate the blocks and operations that implement the CV simulated calculations, however,  the BP(back propagation) calculation is all in RV domain. Then, we explore the proposed classification model on two sets of CV hand gesture radar images in comparison with the equivalent RV residual model.

In chapter six, we propose a CV-forward CNN, which implements the simulated CV operations in the building blocks, however the BP operations are all in RV domain. Then, we explore the proposed classification model on two sets of CV hand gesture radar images in comparison with the equivalent RV-CNN model. 

At the end, we compare and analyse all three proposed models results and recommend future works.

\afterabstract
\prefacesection{Acknowledgements}

I would like to thank my supervisor Dr Fumie Costen for her valuable advice and continues support through all the years of my PhD research.
I would like to thank my kind family: my husband Osman for his continued patience and support and my lovely kids, Dina, Yusuf and Asmaa for giving me the best emotional support and motivation.

\afterpreface

\newcommand{\eq}[1]{~(\ref{eq:#1})}
\newcommand{\fig}[1]{~Figure~\ref{fig:#1}}
\newcommand{\tbl}[1]{~Table~\ref{tbl:#1}}
\newcommand{\sect}[1]{Section~\ref{sec:#1}}
\newcommand{\chap}[1]{~Chapter~\ref{cha:#1}}
\newcommand{\eeq}[1]{~\ref{eq:#1}}

\newcommand{\figg}[1]{~\ref{tbl:#1}}

\newcommand{\partialt}[1]{\displaystyle  \frac{\partial #1}{\partial t}}
\newcommand{\partialtt}[1]{\displaystyle  \frac{\partial^{2} #1}{\partial t ^{2}}}

\newcommand{\partialx}[1]{\displaystyle  \frac{\partial #1}{\partial x}}
\newcommand{\partialy}[1]{\displaystyle  \frac{\partial #1}{\partial y}}
\newcommand{\partialz}[1]{\displaystyle  \frac{\partial #1}{\partial z}}

\newcommand{\myvar}[1]{$#1$}
\newcommand{\lr}{\eta}
\newcommand{\X}{\mbox{\boldmath$x$}}

\newcommand{\myyi}{y_m}
\newcommand{\myxi}{{\X}_m}

\newcommand{\myL}{L}
\newcommand{\mylog}{20 \log_{10}}
\newcommand{\mym}{\cal{M}}

\newcommand{\myw}{\mbox{\boldmath$w$}}
\newcommand{\mylw}{L(\myw)}

\newcommand{\myf}{f}
\newcommand{\wpowert}{{\myw}^{(t)}}
\newcommand{\myeta}{\eta}
\newcommand{\xpowerT}{{\X}^T}
\newcommand{\ipowi}{{i=1}^m}
\newcommand{\wpowt}{{\myw}^{(t+1)}}
\newcommand{\myDelta}{{\Delta}_w}
\newcommand{\mysum}{\displaystyle \sum_{m=1}^{\cal{M}}}
\newcommand{\iinm}{m\in[\cal{M}]}
\newcommand{\xy}{(\X,\myy)}

\newcommand{\mynabla}{{\nabla}_{\vecw}}

\newcommand{\myy}{y}
\newcommand{\yhat}{{\hat{y}}^{(m)}}
\newcommand{\xone}{x_1}
\newcommand{\xtwo}{x_2}
\newcommand{\xk}{x_{\kappa}}

\newcommand{\wone}{w_1}
\newcommand{\wtwo}{w_2}
\newcommand{\wk}{{w}_{\kappa}}

\newcommand{\vecw}{\mbox{\boldmath $w$}}
\newcommand{\vecx}{\mbox{\boldmath $x$}}
\newcommand{\xm}{{\vecx}^{(m)}}
\newcommand{\vecym}{{\vecy}^{(m)}}
\newcommand{\ymm}{{y}^{(m)}}

\newcommand{\yone}{y^{(1)}}
\newcommand{\myfrac}{\frac{1}{\cal{M}}}
\newcommand{\partialwk}{\displaystyle \frac{\partial}{\partial {\wk}}}
\newcommand{\partialone}{\displaystyle \frac{\partial}{\partial {\wone}}}
\newcommand{\Q}{\varphi}
\newcommand{\Qvec}{\mbox{\boldmath $\Q$}}
\newcommand{\myLw}{\cal{L}(\vecw)}
\newcommand{\trainx}{\cal {X}}
\newcommand{\xmvecw}{(\xm,\vecw)}

\newcommand{\veccalwone}{\mbox{\boldmath $\calwone$}}
\newcommand{\veccalwo}{\mbox{\boldmath $\calwo$}}
\newcommand{\calwone}{{\cal{W}}_1} 
\newcommand{\calwo}{{\cal{W}}_o}
\newcommand{\Gone}{G_1}
\newcommand{\mykappa}{\kappa}
\newcommand{\vecf}{\mbox{\boldmath $f$}}
\newcommand{\vecfone}{{\vecf}_1}
\newcommand{\vecfo}{{\vecf}_o}
\newcommand{\mydelta}{\delta}
\newcommand{\vecloss}{\mbox{\boldmath $L$}}
\newcommand{\vecy}{\mbox{\boldmath $y$}}
\newcommand{\mycalv}{\cal{\nu}}
\newcommand{\vaddone}{{\mycalv}+1}

\newcommand{\mtrsum}{\displaystyle \sum_{m=1}^{{\mym} -1 - {\mym}_{tr}}}
\newcommand{\myro}{\rho}
\newcommand{\Zv}{\mbox{\boldmath $Z$}}
\newcommand{\Wv}{\mbox{\boldmath $W$}}
\newcommand{\Rv}{\mbox{\boldmath $R$}}
\newcommand{\sone}{{S}_{1}^{(i)}}
\newcommand{\pp}{\vecw ,{\vecx}^{(i)} ,{\vecy}^{(i)}}
\newcommand{\myS}{\mbox{\boldmath $S$}}
\newcommand{\Ovar}{O}
\newcommand{\Ov}{\mbox{\boldmath $O$}}
\newcommand{\Vvar}{V}
\newcommand{\Vv}{\mbox{\boldmath $V$}}
\newcommand{\Xvar}{\mbox{\boldmath $X$}}
\newcommand{\Yvar}{Y}

\newcommand{\vecwonekone}{\mbox{\boldmath ${W_{1}}_{\kpl{1}}$ }}
\newcommand{\wonekone}{{W_{1}}_{\kpl{1}}}
\newcommand{\bonekone}{{b_{1}}_{\kpl{1}}}
\newcommand{\vecwtwoktwo}{\mbox{\boldmath ${{W}_{2}}_{\kpl{1},\kpl{2}}$}}
\newcommand{\wtwoktwo}{{{W}_{2}}_{\kpl{1},\kpl{2}}}
\newcommand{\vecW}{\mbox{\boldmath{$W$}}}
\newcommand{\vecwthree}{\mbox{\boldmath{${W}_{3}$}}}
\newcommand{\wthree}{{W}_{3}}
\newcommand{\vecoonekone}{\mbox{\boldmath ${{O}_{1}}_{\kpl{1}}$}}
\newcommand{\oonekone}{{{O}_{1}}_{\kpl{1}}}
\newcommand{\vecvonekone}{\mbox{\boldmath ${{V}_{1}}_{\kpl{1}}$}}
\newcommand{\vecs}{\mbox{\boldmath{$S$}}}
\newcommand{\otwoktwo}{{{O}_{2}}_{\kpl{2}}}

\newcommand{\vecotwoktwo}{\mbox{\boldmath ${{O}_{2}}_{\kpl{2}}$}}
\newcommand{\btwoktwo}{{b_{2}}_{\kpl{2}}}
\newcommand{\vecbtwoktwo}{\mbox{\boldmath ${b_{2}}_{\kpl{2}}$}}
\newcommand{\vecbonekone}{\mbox{\boldmath $b_{1}$}}
\newcommand{\vecvtwoktwo}{\mbox{\boldmath ${{V}_{2}}_{\kpl{2}}$}}
\newcommand{\vtwoktwo}{{{V}_{2}}_{\kpl{2}}}
\newcommand{\vonekone}{{{V}_{1}}_{\kpl{1}}}
\newcommand{\vthree}{{V}_{3}}
\newcommand{\vecvthree}{\mbox{\boldmath ${{V}_{3}}$}}
\newcommand{\vecb}{\mbox{\boldmath $b$}}
\newcommand{\bthree}{{b}_{3}}
\newcommand{\haty}{\hat{y}}
\newcommand{\Ivec}{\mbox{\boldmath{$I$}}}
\newcommand{\kpl}[1]{ \kappa_{#1}}
\newcommand{\dl}[1]{d_{#1}}
\newcommand{\kl}[1]{ K_{#1}}
\newcommand{\relu}{\mathrm{ReLU}}
\newcommand{\alphl}[1]{{\alpha}_{#1}}
\newcommand{\betl}[1]{{\beta}_{#1}}
\newcommand{\nabll}[1]{{\nabla}_{#1}^{\myL}}
\newcommand{\myvec}[1]{\mbox{\boldmath{$#1$}}}
\newcommand{\reluprime}[1]{{\relu}^{\prime}{#1}}
\newcommand{\rotl}[1]{ {#1}_{\mathrm{,rot180}}}

\chapter{Introduction}

Hand gesture recognition is a branch of human-computer interaction (HCI), hand gesture recognition systems have yielded many exciting advancements in the last decade and be come more popular in HCI in several application areas which spans from safety and security applications to automotive field \cite{g9841948}, mobile phones, home automation and biomedical engineering \cite{g9266565}. Interaction with the infotainment system inside the car cockpit is realised more intuitively nowadays than in the past when a driver had to press various buttons or touch the surface of a screen to perform some fundamental tasks such as turning up and down the radio or activating the air-conditioning system. Virtual reality (VR) \cite{gsyek}\cite{g9841948} and augmented reality (AR) \cite{gsdo} are emerging scientific areas that have utilised the hand gesture. Camera-based and radar-based techniques are two main categories of contactless hand gesture recognition methods \cite{g9841948}. While the performance of the former is affected by the ambient light, the latter is insensitive to varying illumination conditions. In addition, some people do not find it convenient when they are kept under constant surveillance by a camera \cite{gsse}, \cite{gschar}. Furthermore, contactless hand gesture recognition, prevents from the risk of infection or contamination, which is an important issue especially in the clinical field\cite{g9266565}. 

The recent substantial progress in the semiconductor industry, introduced the millimeter-wave radars can beamed into small-sized and portable gadgets, making them more efficient than cameras in terms of power consumption\cite{g9841948}. Various deep neural network architectures have already been inspected for hand gesture recognition systems, including multi layer perceptron (MLP) \cite{gshasht}, convolutional neural network (CNN) \cite{gstwl}, recurrent neural network (RNN) \cite{gstten} and a cascade of the last two architectures as CNN-RNN \cite{gsften}. Previous scientific works have excessively used long-short-term memory (LSTM) networks as RNN. The gated recurrent unit (GRU) proposed in 2014 has demonstrated promising results comparable to its LSTM counterpart in video-based gesture identification \cite{gsfften}s\cite{g9841948}.

 In many scientific and engineering problems, the unknown variables are complex vectors and the main task is to find these variables that minimise complex-variable optimisation problems. Applications of the complex-variable optimisation problem can be found in communications, adaptive filtering, medical imaging, remote sensing\cite{g7152951}, Magnetic Reasoning Images (MRI), Synthetic Aperture Radar Data (SAR) and Very-Long-Baseline Interferometry (VLBI), Antenna Design (AD), Radar Imaging (RI), Acoustic Signal Processing (ASP) and Ultrasonic Imaging (UI), Communications Signal Processing (CSP), Traffic and Power Systems (TPS)\cite{complexnn}. Traditional optimisation methods are used to solve real-valued optimisation problems and cannot be directly applied to CV optimisation problems. To solve the optimisation problems over the complex field, it is required to convert a CV optimisation problem into a real-valued one by splitting the CV numbers into their real and imaginary parts. However, the major disadvantage of this method is that the resulting algorithm will double the dimension compared with the original problem and may break the special data structure. Moreover, they will suffer from high computational complexity and slow convergence when the problem size is large\cite{g7152951}.

In this report we demonstrate and compare three CV network and explore their accuracy and the effect of some hyper parameters setting on the classification accuracy, computation time and number of trainable parameters. A fully CV-CNN including all building blocks forward and backward operations and derivatives all in complex domain, then we implement the designed model in Python from scratch and without the use of any libraries fully in complex domain. We test the designed and implemented network on $2$ sets of  CV hand gesture radar images and the results of our binary gesture classification model is $100\%$. The second CV network is the complex-forward residual network which separates the imaginary and real parts of dataset, the convolutional block result as a CV convolution (with separated imaginary and real part convolution output ), however the back propagation is RV and the network implemented by using Python RV libraries. The third network that we implement and explore is complex-forward CNN which is also implemented by using the Python RV libraries.

\section{Aims}
This report aims to design and implement bellow three binary classification models:
\begin{itemize}
\item  a fully CV-CNN model for an accurate binary classification method of hand gestures, based
on CV 2D radar images. We aim to have every block in the network and all mathematical operations to utilise both real and imaginary parts of the data. The network blocks include convolutional layer, pooling layer, activation function, fully connected layer. The mathematical operations during the training, optimisation and BP the real and imaginary part
are all applied on CV numbers and real and imaginary parts are not splitted during the mathematical operations at any stage.
\item a CV-forward CNN model for binary classification of two CV hand gesture datasets. This model, utilises the simulated CV operations, including convolutional, pooling and activation function. However the BP derivatives are all in the RV domain and the real and imaginary parts of data are always splitted.
\item a CV-forward residual network, which include one residual block with two convolutional layers. Similar to CV-forward CNN model, the model, utilises the simulated CV operations, including convolutional, pooling and activation function. However the BP derivatives are all in the RV domain and the real and imaginary parts of data are always splitted.
\end{itemize}

\section{Objectives}
 
\begin{itemize}
  \item Prepare two CV $2$D radar images datasets each consist of two hand gestures in order to prepare the binary classification CV-CNN's training and test dataset.
  \item Design a mathematical 2-layer CV-CNN, CV-forward CNN and a CV-forward residual network, with details of every CV layer and their CV derivatives.  
\item Implement the designed models in Python. 
 \item Explore and tune the hyper-parameters such as learning rate, convolution feature map size and number of filters. 
\item Analyse the results of the implemented models in order to develop an accurate CV model to classify the hand gesture images.

\end{itemize}

\chapter{Background}
\label{cha:background}
In this chapter we explain the background of machine learning (ML) and five main types of ML. Then we review the artificial neural networks biological origin and discuses their model training methods. In the direction of discussing different model training methods, first we introduce various loss functions and optimisation algorithms and explain their differences. Afterwards, we explain the detailed mathematical operations in back propagation and various activation functions in order to create an insight to the optimisation and evaluation process of a selected training model. Next, we overview the validation techniques and regularisation methods. Finally, we concentrate on the radar-based hand gesture classification literature review and challenges with focus on convolutional neural network methods.

\section{Machine learning }
The idea of ML has been around over the last six decades. In $1950$, Alan Turing brought the idea of ''Can machine think?''. In 1959, Arthur Samuel defined ML as a ''field of study that gives
computers the ability to learn without being explicitly programmed''. Samuel is
credited with creating one of the first self-learning computer programs with his work
at IBM \cite{alma992975995682301631}. Tom M. Mitchell is the chair of ML at Carnegie Mellon University and the author of the book ''Machine Learning''. He defines ML as ''a computer program which is said to learn from experience with respect to some class of tasks and performance measure, if its performance at tasks in, as measured by performance measure, improves with the experience''.

ML is a branch of artificial intelligence and it is a multi-disciplinary subject, related to a wide range of fields. Usually ML pipeline consists of data, learning algorithm and a model. Using computers and software we design
systems that can learn from data in a manner of being trained. The systems
may learn and improve with experience, thus, with time the system refine a model that
can be used to predict outcomes of questions based on the previous learning \cite{alma992975995682301631}. 
 ML evolved as a sub-field of artificial intelligence that involved the development of self-learning algorithms to gain knowledge from that data in order to make predictions. There are five main types of machine learning, supervised, semi-supervised, unsupervised, transfer learning and reinforcement.

\subsection{Supervised learning}

Supervised learning refers to working with a set of labeled training data. For every sample in the training data we have an input and an output object. The main goal in supervised learning is to learn a model from labeled training data that allows us to make predictions about unseen or future data. In unstructured environments, such as an agricultural field, conditions are variable, so robustness of unsupervised algorithms may be at risk \cite{Kotsiantis}. Therefore supervised classification techniques are of special interest in this field, since a training set can be prepared by a priori establishing what features will correspond to the elements of a class, which, in turn, reduces uncertainty and leads to the possible solutions.

Examples of supervised learning algorithms include linear regression, logistic regression, decision trees, support vector machines (SVM), and neural networks. A supervised learning task with discrete class labels is also called a classification task. Another subcategory of supervised learning is regression, where the output is a continues value \cite{pythonbook}.

\subsubsection{Classification}
Classification is subcategory of supervised learning where the goal is to predict the categorical class labels of new instances based on past observations. Those class labels are discrete, un-ordered values that can be understood as the group memberships of the samples. The classification can be binary (two classes), multi-class classification or multi-label \cite{pythonbook}. In summary, the main differences between the classification types are in the number of classes or labels assigned to each sample. In binary classification, there are two classes, while multi-class classification involves assigning a single class label from multiple classes. In multi-label classification, multiple labels can be assigned to each sample. 

\subsubsection{Regression}
\label{sec:reg}
The term regression was devised by Francis Galton in his article ''Regression towards mediocrity in hereditary stature in 1886''.  Another type of supervised learning is the prediction of continuous outcomes, which is also called ''regression analysis''. In regression learning, we are given a number of the predictor (explanatory) variables and a continuous variable (outcome) and we try to find a relationship between those variables that allows us to predict an outcome \cite{pythonbook}. In statistics, the regression analysis is a basic type of predictive analysis which is used to quantify the relationship between a dependent variable known as the ''system output'' and one or more independent variables \cite{regression1}.

\subsection{Unsupervised learning}
Unsupervised learning is where we let the algorithm find a hidden pattern in a set of data. With unsupervised learning there is no right or wrong answer, it’s just a case of running the ML algorithm and seeing what patterns and outcomes occur \cite{alma992975995682301631}. When using unsupervised learning techniques, we are able to explore the structure of our data to extract meaningful information without the guidance of a known outcome variable or reward function \cite{pythonbook}.

\subsection{Semi-supervised learning}
Semi-supervised learning lies between supervised and unsupervised learning. It combines a small amount of labeled data with a larger set of unlabeled data. The goal is to leverage the unlabeled data to improve the learning process and enhance the model's performance. Techniques such as self-training, co-training, and graph-based methods are commonly used in semi-supervised learning.

\subsection{Transfer learning}
Transfer learning involves using knowledge or representations learned from one task or domain to improve performance on a different but related task or domain. Instead of starting from scratch, the model leverages the pre-existing knowledge to bootstrap the learning process in the new task. This can be especially useful when the target task has limited labeled data, as the model can transfer knowledge from a source task with abundant labeled data. Transfer learning is often employed in deep learning models, where pre-trained models (e.g., ImageNet pre-trained models) are fine-tuned on new tasks.
\subsection{Reinforcement learning}
The aim of reinforcement learning is to develop a system (agent) that improves its performance based on interactions with the environment. Since the information about the current state of the environment typically also includes a reward signal, we can think of a reinforcement learning as a field related to supervised learning. However, in reinforcement learning, this feedback is not the correct ground truth label or value, but a measure of how well the action was measured by the reward function. Through the interaction with the environment, an agent can then use reinforcement learning to learn a series of actions that maximises this reward via an exploratory trial-and-error approach \cite{pythonbook}.

\section{Artificial neural network}

\begin{figure}
\centering\includegraphics[height=7cm, width=15cm]{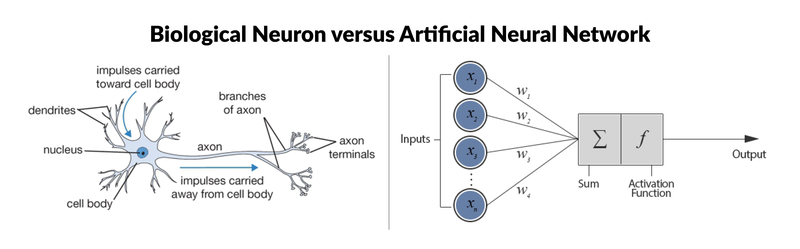}
\caption{A single artificial neuron and a biological neuron \cite{datacamp}}
\label{fig:biologicalneuron}
\end{figure}

The idea of building an artificial brain has existed for a long time. ANN is a possible method to help to better understand artificial intelligence \cite{ANN610}. ANN is a supervised learning algorithm inspired by biological operations consisting of a group of interconnected artificial neurons that work together to solve a specific problem (\fig{biologicalneuron}). Although ANN has gained more popularity in recent years, the earliest studies of neural networks date back to the $1940$s, when Warren McCulloch and Walter Pitt first described how neurons work. However, even after Rosenblatt's perceptron in the $1950$'s, which was the first implementation of McCulloch and Pitt's model, no one had a good solution for training a neural network with multiple layers. Eventually in 1983, Michalski proposed a machine that can learn from labeled samples \cite{ANN11}.

 In 1986, when Rumelhart, G.E. Hilton and R.J. Williams introduced a back-propagation (BP) algorithm to train ANN online automatically \cite{ANN12}, \cite{ANN13}, subsequently, some studies showed that memristors could be used as electronic synapses in ANN \cite{ANN16}, \cite{ANN18}. For example, ANN consisting of neurons and memristor-based synapses was used to mimic the associate function of human brain \cite{ANN16} , \cite{ANN19}. ANN shows a powerful and robust performance in modelling a complex system. Since then, researchers have made many amazing achievements in the applications of the ANN like pattern recognition \cite{ANNone}, \cite{ANN2}, \cite{ANN3}, face recognition \cite{ANN4}, \cite{ANN5}, learning cat concept from cat videos on the internet \cite{ANN6}, classifying \cite{ANN7}, and playing 'Pokemon Go' game \cite{ANN8}. ANN is a hot topic not only in academic research, but also in big technology companies such as Facebook, Microsoft and Google who invest heavily in ANN and deep learning research.

\subsection{Training an ANN}
The multi-layer neural network is a typical example of a feed-forward ANN. The term feed-forward refers to the fact that each layer serves as the input to the next one without loops. The training procedure starts at the input layer, we forward propagate the patterns of the training data through the network to generate an output. The second step, based on the network's output, we calculate the error that we want to minimise by using a loss (cost) function. The third step is to back-propagate the loss, find its derivative with respect to each weight in the  network and update the model\cite{pythonbook}.

\subsubsection{Loss function}
Machines learn by means of a ''loss function''. It’s a method of evaluating how well a specific algorithm models the given data. If prediction deviates too much from actual results, the loss function would cough up a very large number. Gradually, with the help of some optimisation functions, the loss function learns to reduce the error in prediction. There are various factors involved in choosing a loss function for specific problems, such as type of ML algorithm chosen, ease of calculating the derivatives and to some degree the percentage of outliers in the data set. There are many loss functions which are commonly used for different purposes in ML, such as mean square error (MSE), mean absolute error (MAE) and mean bias error (MBE):
\begin{eqnarray}
\label{eq:lossmse}
\mathrm{MSE} & = &  (y-\hat{y})^{2}.
\end{eqnarray}
Equation\eq{lossmse} calculates the MSE, where $y$ is its label of a training sample and $\hat{y}$ is the predicted output  of the training sample. MSE is measured as the average of squared difference between predictions and actual observations . It is only concerned with the average magnitude of error irrespective of their direction. However, due to squaring, predictions which are far away from actual values are penalised heavily in comparison to less deviated predictions. Moreover, it is computationally easy to calculate the gradients.
\begin{eqnarray}
\label{eq:lossmae}
\mathrm{MAE} & = & \mid y-\hat{y}\mid.
\end{eqnarray}
MAE as in equation\eq{lossmae}, is measured as the absolute differences between predictions and actual observations. Like MSE, it measures the magnitude of error without considering their direction. Unlike MSE, MAE needs more complicated tools such as linear programming to compute the gradients. In addition, MAE is more robust to outliers since it does not make use of the square.
\begin{eqnarray}
\label{eq:lossmbe}
\mathrm{MBE} & = & y-\hat{y}.
\end{eqnarray}
MBE as in equation\eq{lossmbe}, is less popular in the ML domain. There is a need for caution as positive and negative errors could cancel each other out. Although less accurate in practice, it could determine if the model has positive or negative bias.

\subsubsection{Gradient based optimisation algorithm}
\begin{figure}
\includegraphics[width=0.7\textwidth]{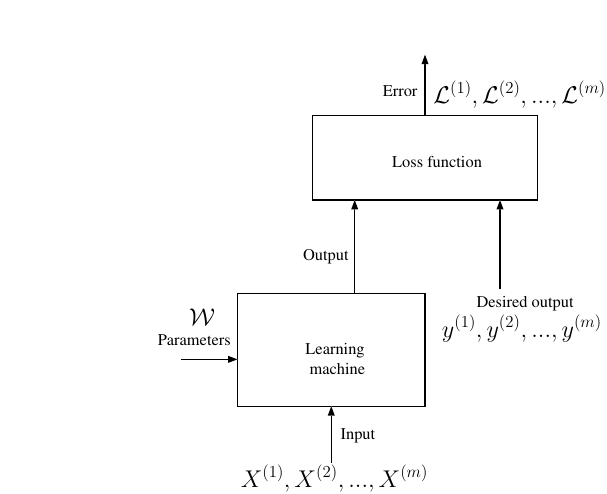}
\caption{Gradient-based learning machine}
\label{fig:gradientbox}
\end{figure}

Gradient descent is an iterative algorithm for finding the local or global minimum of the ''loss function''. It measures the closeness of a desired output for an input to the output of the network (predicted output). As the model iterates, it gradually converges towards a minimum where further tweaks to the parameters produce little or zero changes in the loss, which is also referred to as convergence. Let us start with a training set which is a set of samples, each sample consisting of a pair of an input and a desired output. The pairs are the samples of the function to be learned. There are several algorithms in ML, most of the successful algorithms can be categorised as gradient-based learning methods. The
learning machine, as represented in\fig{gradientbox}, computes a function $\myf(\xm, \vecw)$
where $\xm$ is the vector of $m$-th input, and $\vecw$ represents the vector collection of adjustable
parameters in the system.
\begin{eqnarray}
\label{eq:regfirst}
{\myL}^{(m)} & = & {\myL}({\myf}({\xm},{\vecw}),{\vecym})
\end{eqnarray}
A loss function ${\myL}^{(m)}$\eq{regfirst} measures the discrepancy between ${\vecym}$ the ''correct'' or desired output for the $mth$ input $\xm$, and
the predicted output by the system $\yhat=$ $(\myf(\xm, \vecw))$. The average loss function ${\myL}$ is the
average loss function over a set of input and output pairs called the training set
${(\vecx^{(1)}, \vecy^{(1)}), ....(\xm, \vecy^{(m)})}$. In the simplest setting, the learning algorithm consists in
finding the value of $\myw$ that minimises loss \cite{backprop}. In practice, the performance
of the system on a training set is of little interest. The more relevant measure
is the loss rate of the system in the field, where it would be used in practice. This performance is estimated by measuring the accuracy on a set of samples
disjoint from the training set, called the test set. The MSE loss function, which is used commonly in regression problems, measures the average squared difference between an desired actual and predicted values, as in\eq{lossmse}.
\begin{eqnarray}
\label{eq:regsecond}
{\myL} & = & \myfrac \mysum (\myf {( \xm , \vecw ) - \vecy^{(m )})}^{2}
\end{eqnarray}
The output of equation\eq{regsecond} is a single scalar representing the average loss, associated with the current set of weights. Our goal is to minimize MSE to improve the accuracy of our model. 
The momentum method \cite{polyak}, which we refer to as classical momentum (CM), is a technique for accelerating gradient descent that accumulates a velocity vector in directions of persistent reduction in the objective across iterations. Intuitively, the rational for the use of the momentum term is that the steepest descent is particularly slow when there is a long and narrow valley in the error function surface. 

In this situation, the direction of the gradient is almost perpendicular to the long axis of the valley. The system thus oscillates back and forth in the direction of the short axis, and only moves very slowly along the long axis of the valley. The momentum term helps average out the oscillation along the short axis while at the same time adds up contributions along the long axis \cite{Rumelhart}.
 \cite{polyak} showed that CM can considerably accelerate convergence to a local minimum, requiring fewer iterations than steepest descent to reach the same level of accuracy.

There are three approaches to gradient descent algorithm: batch, stochastic and mini-batch gradient descent \cite{Sebastian}. The amount of data we use to compute the gradient of the loss function differs between each type. Depending on the amount of the training set, we make a trade-off between the accuracy of the parameter update and the time it takes to perform an update.

\paragraph{Batch gradient descent}
Batch gradient descent computes the gradient of the loss function with respect to all the weights ($\nabla_{\vecw} \myL$) for the entire training dataset to update the weights of the network.
\begin{eqnarray}
\label{eq:batchh}
{\vecw}^{(i+1)} &  = & {\vecw}^{(i)} -{\eta}  {\nabla}_{{\vecw}^{(i)}} \myL(\vecw)                
\end{eqnarray}

Where in above equation\eq{batchh}, $\eta$ denotes the learning rate, ${\vecw}^{(i)}$ denotes the weight vector that contains the weights in $i$th iteration and ${\vecw}^{(i+1)}$ denotes the weight vector in $i+1$~th iteration. As we need to calculate the gradients for the whole dataset to perform just one update, batch gradient descent can be very slow and is intractable for large training datasets that do not fit in memory. We then update our parameters in the direction of the gradients with the learning rate determining how big of an update we perform. Batch gradient descent is guaranteed to converge to the global minimum for convex error surfaces and to a local minimum for non-convex surfaces \cite{Sebastian}.

\paragraph{Stochastic gradient descent (SGD)}
When we have a very large training dataset, running batch gradient descent can be computationally costly, because we need to reevaluate the whole training dataset each time we take one step towards global minimum. A popular alternative to the batch gradient descent algorithm is ''stochastic gradient descent'', sometimes SGD is called online or iterative gradient descent. Instead of updating the weights based on the sum of the accumulated loss over all the sample batch.

The SGD updates the weights in\eq{batchh} incrementally after each training sample. Thus, it reaches convergence much faster because of the more frequent weight updates. Since each gradient is calculated based on a single training example, SGD is computationally efficient as it processes one example at a time, but its parameter updates can be noisy and exhibit high variance, which can also have the advantage that SGD can escape shallow local minimum more rapidly. To obtain accurate results via SGD it is important to present it with the data in random order, which is why the training data should be shuffled for each epoch. Another advantage of SGD is that we can use it for online learning, means our model is trained on-the-fly as new training data arrives. This is especially useful if we are accumulating large amounts of data.

The noise in SGD arises from a few factors: 
\begin{itemize}
\item Sample Variability: The gradients computed from individual examples may vary significantly due to the inherent noise or randomness in the data. This variation in gradients can lead to inconsistent updates to the model's parameters. 

\item Learning Rate Sensitivity: SGD typically uses a fixed learning rate for parameter updates. This can cause larger fluctuations in the optimisation process, especially when the learning rate is not carefully tuned. The learning rate determines the step size in each parameter update, and if set too high, it can cause the model to overshoot the optimal solution, resulting in instability. 
\item Sequential Dependency: Since SGD processes examples one at a time, the order of the examples can impact the optimisation process. The sequence in which examples are presented affects the parameter updates and can introduce bias or oscillations in the convergence process. 
\item Local Minima Escaping: The noisy updates in SGD can sometimes help the algorithm escape shallow local minima and find better solutions. This is because the randomness in the updates allows the algorithm to explore different areas of the optimisation landscape, potentially finding better regions that could have been missed by a more deterministic algorithm like batch gradient descent.
\end{itemize}

\paragraph{Mini-batch gradient descent}

Mini-batch gradient descent finally takes the best of both worlds and performs an update in\eq{batchh} for every mini-batch training example. Mini-batch reduces the variance of the parameter updates, which can lead to more stable convergence. In addition, it can make use of highly optimised matrix optimisation, common to state-of-the-art deep learning software libraries that make computing the gradient of a mini-batch very efficient. Common mini-batch sizes range between 50 and 256, that can vary for different applications. Mini-batch gradient GD is typically the algorithm of choice when training a neural network, and usually ''SGD'' is employed when mini-batches are used \cite{Sebastian}.

\begin{figure}
\centering\includegraphics[scale=1.5]{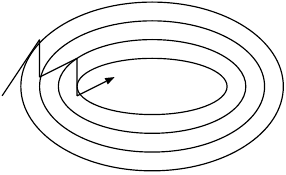}
\caption{The SGD with momentum}
\label{fig:moment2}
\end{figure}

\paragraph{Momentum}
There are some techniques that are widely used by the ANN and deep learning community to deal with the aforementioned challenges. Some of the techniques are momentum and Nesterov accelerated gradient. SGD has trouble navigating ravines, such as areas where the surface curves much more steeply in one dimension than in another, which are common around local optima. In these scenarios, SGD oscillates across the slopes of the ravine while only making hesitant progress along the bottom towards the local optimum, as in\fig{moment2}. 

A higher momentum value increases the impact of past gradients on the parameter updates. This can help overcome small, localised fluctuations in the gradient and provide smoother convergence. However, a very high momentum value may cause the updates to overshoot the optimal solution, leading to instability or oscillations. The momentum value should be chosen in conjunction with the learning rate. If a higher momentum value is used, a smaller learning rate might be appropriate to ensure stability. It is important to note that the choice of the momentum value often involves empirical experimentation and tuning. It depends on the characteristics of the specific problem, the dataset, and the behavior of the optimisation process. 

\begin{figure}
\centering\includegraphics[scale=1.5]{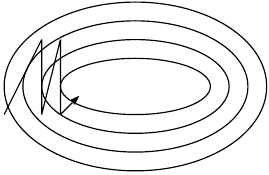}
\caption{The SGD without momentum}
\label{fig:moment1}
\end{figure}

Momentum \cite{Sebastian} is a method that helps accelerate SGD in the relevant direction and dampens oscillations as can be seen in\fig{moment1}. It does this by adding a fraction $\gamma$ of the update vector of the past time step to the current update vector as in    

\begin{eqnarray}
\label{eq:momentumon}
\Delta {\vecw} ^{(i+1)} = - \myeta \nabla_{{\vecw} ^{(i)}} {\myL} + \mu \Delta {\vecw} ^{(i)}\nonumber \\
{\vecw} ^{(i+1)} = {\vecw} ^{(i)} + \Delta {\vecw} ^{(i)}
\end{eqnarray}
where $\mu$ is the momentum parameter,~$\Delta {\vecw} ^{(i)}$  is, the modification of the weight vector at the
current time step depends on both the current gradient and the weight change of the previous step. The momentum term is usually set to $0.9$ or a similar value, as it has shown to work well in many cases. This value provides a reasonable balance between incorporating past gradients and maintaining stability. 

However, it is important to note that the optimal value may vary depending on the specific problem and dataset. Essentially, when using momentum, we push a ball down a hill. The ball accumulates momentum as it rolls downhill, becoming faster and faster on the way (until it reaches its terminal velocity, if there is air resistance as $\mu$ $ < 1$). The same thing happens to our parameter updates: The momentum term increases for dimensions whose gradients point in the same directions and reduces updates for dimensions whose gradients change directions. As a result, we gain faster convergence and reduced oscillation.

\subsubsection{Back propagation}
BP or ''backward propagation of errors'', is a standard method of training ANN. This method helps to calculate the gradient of a loss function with respects to all weights in the network. The BP algorithm was originally introduced in the $1970$s, but its importance wasn't fully appreciated until a famous $1986$ paper by David Rumelhart, Geoffrey Hinton, and Ronald Williams \cite{ANNone}. BP is an expression for the partial derivative of the loss function $\myL$ with respect to any weight $\vecw$ in the network. BP provides detailed insights into how changing the weights changes the overall behaviour of the network \cite{ANNone}.

Although BP was rediscovered and popularised almost $30$ years ago, it still remains one of the most widely used algorithms to train an ANN. BP is a very popular neural network learning algorithm because
it is conceptually simple, computationally efficient and it often works accurately. Designing and training a network using BP requires
making many seemingly arbitrary choices such as the number and types of nodes,
layers, learning rates, training and test sets. Proper tuning of the weights allows you to reduce error rates and to make the model reliable by increasing its generalisation\cite{backprop}.
\begin{figure}
\centering\includegraphics[height=5cm, width=13cm]{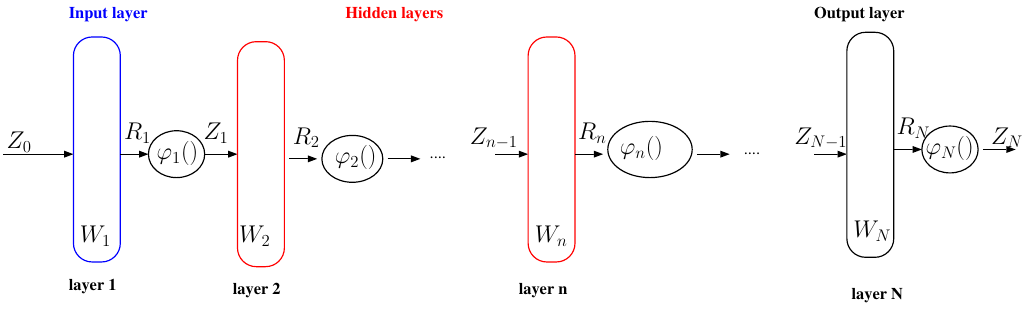}
\caption{A multi-layer feed forward neural network}
\label{fig:backpropnet}
\end{figure}
Lets assume we have a multi-layer feed-forward neural network which consist of $N$ layers of neurons(\fig{backpropnet}). 

\begin{eqnarray}
\label{eq:bkproptwo}
{{\Rv}_{n}} &=& {{\Wv}_{n}} {{\Zv}_{n-1}} \nonumber \\
{{\Zv}_{n}} &=& {{\Qvec}_{n}}  {{\Rv}_{n}}.
\end{eqnarray}

For each layer of the multi-layer neural network, where in\eq{bkproptwo} ${{\Wv}_{n}}$ is a weight parameter matrix of the $n$th layer whose number of columns is the dimension of ${\Zv}_{n-1}$, and number of rows is the dimension of ${\Zv}_{n}$. $\Qvec$ is a vector function that applies an activation function to each component of its input. Each layer implement the functions as in \eq{bkproptwo} , where ${\Rv}_{n}$ is a vector representing the $n$-th layer's input to activation function and ${\Zv}_{n-1}$ is the output vector of the $n-1$th layer as well as the input vector of $n$-th layer. ${\Zv}_{0}$ is the input vector and  ${\Zv}_{N}$ is the output. ${\myL}^{(m)}$ is the loss function for the $mth$ sample. In the BP algorithm, we calculate the $\frac{\partial {\myL}^{(m)}}{\partial {\Wv}_{n}}$ for each training sample $m$, then calculate the $\frac{\partial \myL}{\partial {\Wv}_{n}}$ by averaging over the training samples.

\begin{eqnarray}
\label{eq:backponee}
\frac{\partial {\myL}}{\partial {\Rv}_n} & = & \frac{\partial \myL} {\partial {\Zv}_n} { \frac {\partial{\Zv}_n}{\partial {\Rv}_n}}\nonumber \\ 
& = & \frac{\partial {\myL}}{\partial {\Zv}_{n}} {\frac {\partial {\Qvec}_{n} ({\Wv}_{n} {\Zv}_{n-1})}{\partial {\Rv}_{n}}} \nonumber \\
\frac{\partial \myL}{\partial {\Wv}_{n}} & = & \frac{\partial \myL}{\partial {\Zv}_{n}} {\frac {\partial {\Zv}_{n}}{\partial {\Wv}_{n}}} \nonumber \\
& = & \frac{\partial \myL}{\partial {\Zv}_{n}}  {\frac {\partial {\Qvec}_{n} ({\Wv}_{n} {\Zv}_{n-1})}{\partial {\Wv}_{n}}}\nonumber \\
\frac{\partial \myL}{\partial {\Zv}_{n-1}} & = & \frac{\partial \myL}{\partial {\Zv}_{n}}\frac  {\partial {\Zv}_{n}}{\partial {\Zv}_{n-1}}\nonumber  \\ 
& =& \frac {\partial \myL}{\partial {\Zv}_{n}}\frac {\partial {\Qvec}_{n} ({\Wv}_{n} {\Zv}_{n-1}) }{\partial {\Zv}_{n-1}}
\end{eqnarray}
If the partial derivative of ${\myL}$ with respect to ${\Zv}_{n}$ is known then the partial derivatives of ${\myL}$ with respect to ${\Wv}_{n}$  and ${\Zv}_{n-1}$ can be computed using the backward recurrence as in\eq{backponee}.

\begin{eqnarray}
\label{eq:varph}
\frac {\partial {\Qvec}_{n} ({\Wv}_{n} {\Zv}_{n-1})}{\partial {\Rv}_{n}} = {\Qvec}^{\prime}_{n} ({\Rv}_{n})
\end{eqnarray}
Let  ${\Qvec}^{\prime}_{n} ({\Rv}_{n})$ be the derivative of ${\Qvec}$ with respect to ${\Rv}_{n}$ as in\eq{varph} and\eq{bkproptwo}. Applying the chain rule to BP equations, the classical BP
equations in matrix form are obtained as: 

\begin{eqnarray}
\label{eq:backpropthree}
\frac {\partial \myL}{\partial {\Rv}_{n}} & = & \frac{\partial \myL}{\partial {\Zv}_{n}} {\Qvec}^{\prime}_{n} ({\Rv}_{n}) \nonumber \\
\frac {\partial \myL}{\partial {\Wv}_{n}} & = & \frac{\partial \myL}{\partial {\Zv}_{n}}  {\Zv}_{n-1} \nonumber \\
\frac {\partial \myL}{\partial {\Zv}_{n-1}} & = & \frac{\partial \myL}{\partial {\Zv}_{n}}  {{\Wv}_{n}}^T.
\end{eqnarray}

When the BP equations are applied to the layers in reverse order, from layer N to layer 1, all the partial derivatives of the loss function with respect to all the weights parameters can be computed. The way of computing gradients is known as BP. Let ${{\myS}_{n}}$ be the sensitivity in layer $n$.

\begin{eqnarray}
\label{eq:backpropfive}
{\myS}_{n} & =& \frac {\partial \myL}{\partial {\Rv}_{n}}
\end{eqnarray}
We define the sensitivity as in equation\eq{backpropfive}, where the ${{\myS}_{n}}$ is a vector of sensitivity of the neurons in layer $n$. The BP provides an algorithm to compute the $ {{\myS}_{n}}$ for every layer and relating the sensitivities to the $\frac{\partial \myL}{\partial {\Wv}}$:

\begin{eqnarray}
\label{eq:backpropfour}
\frac {\partial \myL}{\partial {\Wv}_{n}} & = & \frac {\partial \myL}{\partial {\Rv}_{n}}  {\frac {\partial {\Rv}_{n}}{\partial {\Wv}_{n}}}\nonumber \\
& =& {\myS}_{n} \frac {\partial {\Rv}_{n}}{\partial {\Wv}_{n}}\nonumber \\
&=& {\myS}_{n} {\Zv}_{n-1}.
\end{eqnarray}
Now we start backward and compute the $ \frac {\partial \myL}{\partial {\Wv}_{n}}$ from the output layer towards input layer as in equation\eq{backpropfour}. We can compute the sensitivity of the last layer $ {{\myS}_{N}}$ from\eq{backpropthree} as in:
\begin{eqnarray}
\label{eq:sensyek}
 {{\myS}_{N}} & =& \frac {\partial \myL}{\partial {\Rv}_{N}}.
\end{eqnarray}

If we apply\eq{backpropthree} to compute the sensitivity of the last layer $N$ we have
\begin{eqnarray}
\label{eq:bpyek}
{\myS}_{N}= \frac {\partial \myL}{\partial {\Zv}_{N}} {{\Qvec}^{\prime}_{N}({\Rv}_{N})}
\end{eqnarray}


Every term in\eq{bpyek} is easily computed. In particular, we compute ${\Zv}_{N}$ while computing the feed-forward network,so that we can compute $\Rv_{n}$ so that we can compute ${\Qvec}^{\prime}_{N}({\Rv}_{N})$ as the activation function and its derivation is known. The exact form of $\frac {\partial \myL}{\partial {\Zv}_{N}}$ will depend on the form of the loss function.  For example, if we are using the MSE loss function then ${\myL}=\frac{1}{2}({\vecy}_{N}-{\Zv}_{N})^{2}$ and so in the case of MSE loss function the $\frac {\partial \myL}{\partial {\Zv}_{N}}$ will be $({\Zv}_{N} - {\vecy}_{N})$. So for the case of MSE loss function we have

\begin{eqnarray}
\label{eq:sendoo}
S_{N} =({\Zv}_{N} - {\vecy}_{N})  \odot {\Qvec}^{\prime}_{N}({\Rv}_{N})
\end{eqnarray}


In the general case, we define $\nabla_{\Zv} \myL = \frac {\partial \myL}{\partial {\Zv}_{N}}$, so applying \eq{bpyek} we have
\begin{eqnarray}
\label{eq:bpdoo}
\myS_{N} = \nabla_{{\Zv}_N} \myL \odot {{\Qvec}^{\prime}_{N}} ({\Rv}_{N})
\end{eqnarray}

Whereas $\nabla_{{\Zv}_{N}} \myL$ is defined to be a vector whose components are the partial derivatives $\frac {\partial \myL}{\partial {\Zv}_{N}}$. The term $\nabla_{{\Zv}_{N}} \myL$ expresses the rate of change of $\myL$ with respect to the output activation. We use $\odot$ to denote the elementwise product of the two vectors.
This kind of elementwise multiplication is sometimes called the Hadamard product or Schur product. Next, we will provide a way to compute the sensitivity of the other layer ~\eq{backpropfive}.

\begin{eqnarray}
\label{eq:sense}
{\myS}_{n} & = & \frac{\partial \myL}{\partial {\Rv}_{n}}\nonumber \\
  & = & \frac {\partial \myL}{\partial {\Rv}_{n+1}}  {\frac {\partial {\Rv}_{n+1}}{\partial {\Rv}_{n}}} \nonumber \\ 
  & = & \frac{\partial{\Rv}_{n+1}}{\partial {\Rv}_{n}} {\myS}_{n+1}
\end{eqnarray}

Thus, as in \eq{sense} in order to compute the sensitivity of the $n$th layer we need the sensitivity of the next layer in addition to the term $ \frac{\partial{\Rv}_{n+1}}{\partial {\Rv}_{n}}$. Applying forward propagation rules, we have 
\begin{eqnarray}
{\Rv}_{n+1} =  {\Wv}_{n+1} {\Zv}_{n} =  {\Wv}_{n+1} {\Qvec}_{n}({\Rv}_{n})
\end{eqnarray}
We defined ${\Qvec}^{\prime}$ as \eq{varph}, so we have 
\begin{eqnarray}
\frac {\partial {\Rv}_{n+1}}{\partial{{\Rv}_{n}}}
& =& {\Wv}_{n+1} {\Qvec}^{\prime}_{n}( {\Rv}_{n})
\end{eqnarray}

\begin{eqnarray}
\label{eq:bpse}
\myS_{n} = {\Wv}_{n+1}{}^T {\myS}_{n+1} \odot {{\Qvec}^{\prime}({\Rv}_{n})}  
\end{eqnarray}

So we have the vectorized format of ${\myS}_{n}$ as~\eq{bpse}. The simplest learning (minimisation) procedure in such a setting is the gradient
descent algorithm where ${\Wv}$ is iteratively adjusted as
\begin{eqnarray}
{\Wv}_{n}^{(i+1)} & = & {\Wv}_{n}^{(i)} - \myeta \frac{\partial \myL}{\partial {\Wv}_{n}}\nonumber \\
{\Wv}_{n}^{(i+1)} & = & {\Wv}_{n}^{(i)} - \myeta { {S}_{n}} {\Zv}_{n-1}
\end{eqnarray}
where ${\Wv}^{(i)}$ is the $i$-th weight parameter matrix and $\myeta$ is the learning rate. In the simplest case, $\myeta$ is a scalar constant. The sensitivity of the neuron in layer $n$ depends on the sensitivity of the neuron in layer $n-1$, it is recursion relation for the sensitives of the different layers of the network. Since the sensitivity of the last layer $N$ is known, To calculate the sensitivity of other layers, we need to start from the last layer and use the recursion relation and go backward, that is why the training algorithm is called back propagation.

The BP can be very slow particularly for multilayered networks where
the loss surface is typically non-quadratic, non-convex, and high dimensional
with many local minima and/or flat regions. There is no formula to guarantee
that the network will converge to a good solution, convergence is swift or convergence even occurs at all. However there are some techniques such as SGD that can improve the minimising procedure. 

\subsubsection{Activation function}

Activation functions are mathematical equations that determine the output of a neural network. The function is attached to each neuron in the network, and determines whether it should be activated (“fired”) or not, based on whether each neuron’s input is relevant for the model’s prediction. One aspect of activation functions is that they must be computationally efficient because they are calculated across thousands or even millions of neurons for each data sample. Modern neural networks use BP technique to train the model, which places an increased computational strain on the activation function, and its derivative function.

\begin{figure}
\centering\includegraphics[height=5cm, width=11cm]{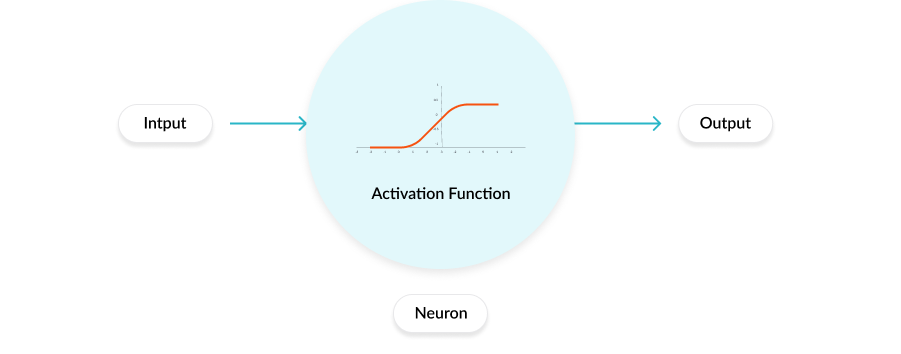}
\caption{Activation function role in ANN \cite{activationfunction} .}
\label{fig:actfunone}
\end{figure}

In a neural network, numeric data points, called inputs, are fed into the neurons in the input layer. Each neuron has a weight, and multiplying the input number with the weight gives the output of the neuron, which is transferred to the next layer.
The activation function is a mathematical ''gate'' in between the input feeding the current neuron and its output going to the next layer(\fig{actfunone}). It can be as simple as a step function that turns the neuron output on and off or it can be a transformation that maps the input signals into output signals that are needed for the neural network to function \cite{activationfunction}.

Increasingly, neural networks use non-linear activation functions, which can help the network learn complex data, compute and learn almost any function representing a question, and provide accurate predictions. Non-linear functions address the problems of a linear activation function. They allow backpropagation because they have a derivative function which is related to the inputs. They allow “stacking” of multiple layers of neurons to create a deep neural network. Multiple hidden layers of neurons are needed to learn complex data sets with high levels of accuracy\cite{activationfunction}.

\subsubsection{Sigmoid}
\begin{figure}
\centering\includegraphics[height=5cm, width=10cm]{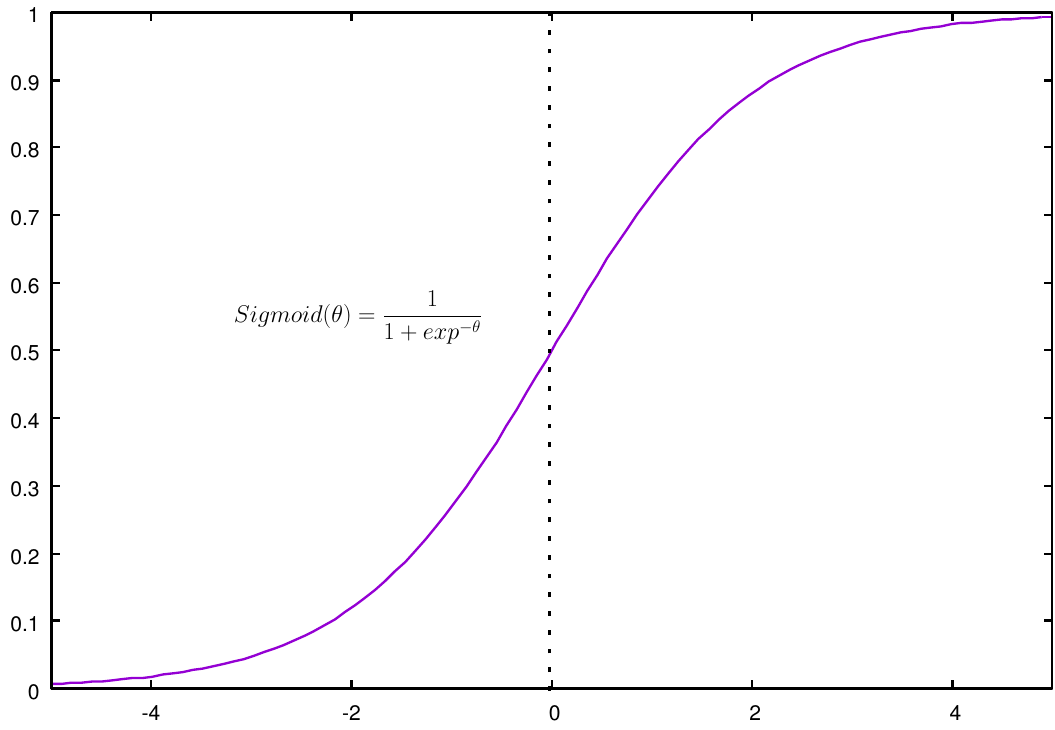}
\caption{Sigmoid activation function.}
\label{fig:sigmoidact}
\end{figure}

The Sigmoid activation function $Sigmoid(\mathrm \theta)=\frac{1}{1+{\exp}^{- \mathrm {\theta}}}$ as shown in~\fig{sigmoidact}, is one of most widely used non-linear activation. The smooth gradient of sigmoid activation function prevents “jumps” in the output values. Moreover, the output values bound between $0$ and $1$, normalising the output of each neuron makes this function very suitable for models that require probabilistic interpretations or binary classification tasks. It can also be used as an activation function in the output layer for multi-label classification.

However for very high or very low values of activation function input there is almost no change to the prediction, causing a vanishing gradient problem. This can result in the network refusing to learn further, or being too slow to reach an accurate prediction. In addition, the outputs are not zero centered and sigmoid calculation is computationally expensive.

\subsubsection{Tanh (Hyperbolic tangent)}

\begin{figure}
\centering\includegraphics[height=5cm, width=10cm]{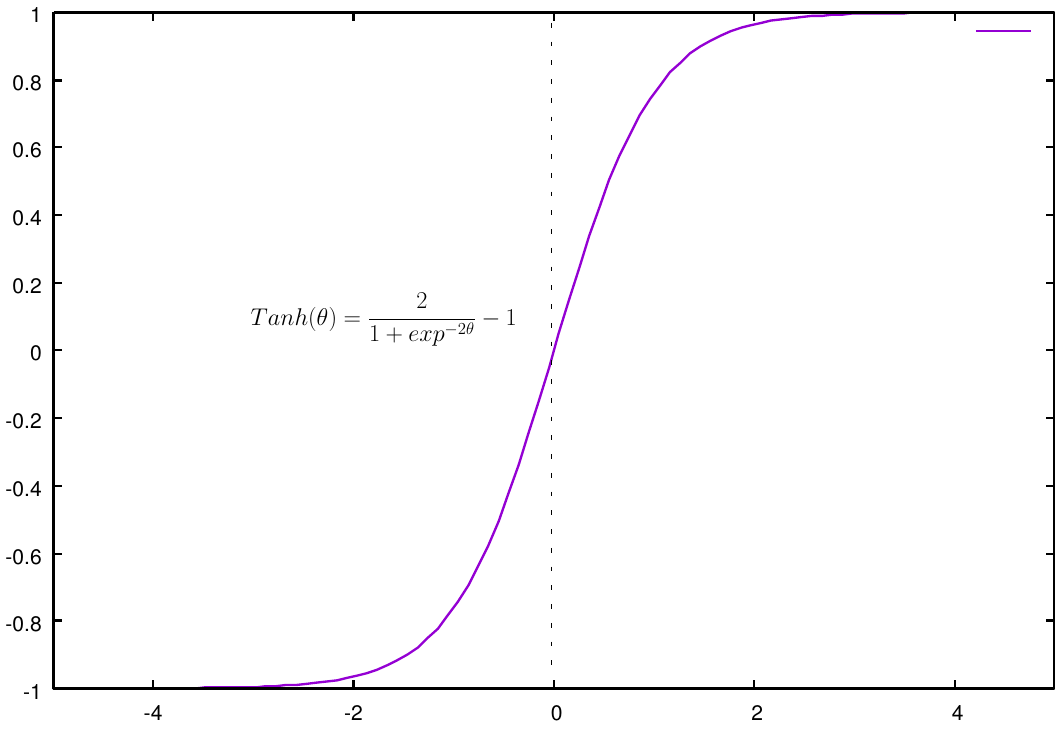}
\caption{Tanh activation function Tanh.}
\label{fig:tanh}
\end{figure}

The Tanh activation function 
Tanh$(\mathrm{\theta})=\frac {2}{1+{\exp}^{-2 \mathrm{\theta}}} - 1$ as shown in\fig{tanh} is a zero-centred function which makes it easier to model inputs that have strongly negative, neutral and strongly positive values. The characteristics of Tanh function is similar to sigmoid function, however, the gradient is stronger for Tanh than sigmoid ( derivatives are steeper). Deciding between the sigmoid or Tanh will depend on the requirement of gradient strength. Like sigmoid, Tanh also has the vanishing gradient problem.
Tanh is also a very popular and widely used activation function.

\subsubsection{ReLU ( rectified linear unit)}

\begin{figure}
\centering\includegraphics[height=5cm, width=10cm]{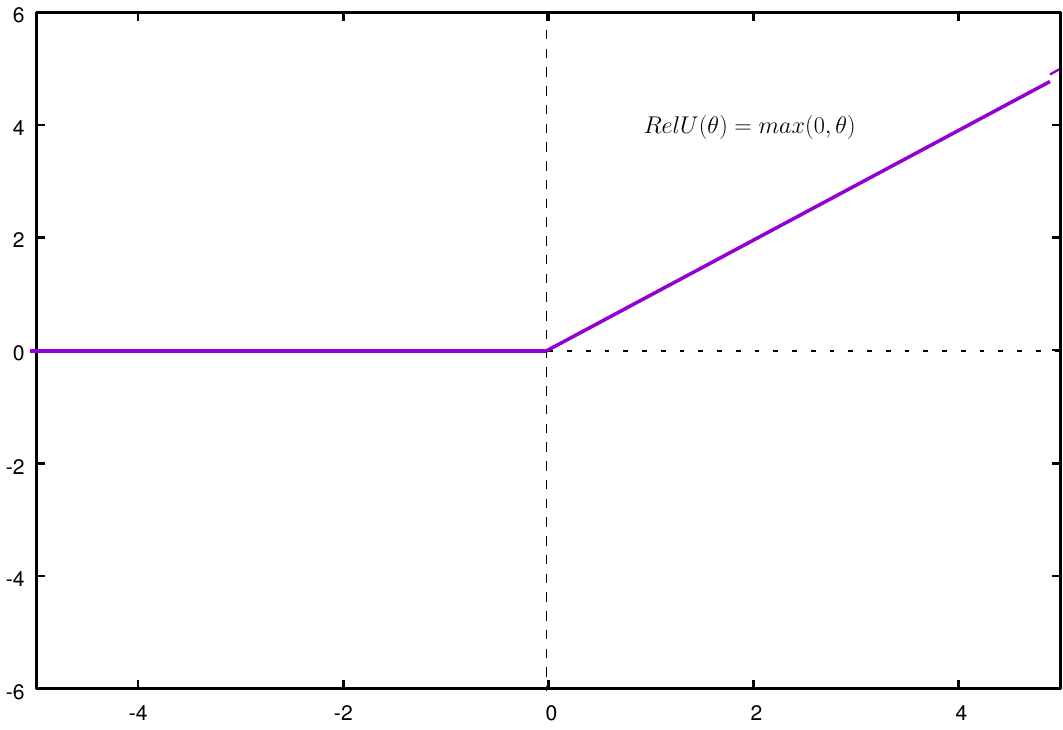}
\caption{ReLU activation function.  }
\label{fig:relu}
\end{figure}

ReLU function ReLU$(\mathrm{\theta})= max(0,\mathrm{\theta})$ as shown ins\fig{relu}. ReLU calculation is computationally efficient (only comparison, addition and multiplication) which allows the network to converge very quickly. ReLU activation function is non-linear, although it looks like a linear function, ReLU has a derivative function so it can be utilised for BP.

In addition, for larger Neural Networks, the speed of building models based on ReLU is very fast because of sparse activation, which means in a randomly initialised network, only about half of hidden units are activated (having a non-zero output). Moreover, ReLU has better gradient propagation characteristics, so fewer vanishing gradient problems compared to sigmoidal activation functions will accrue. However, when inputs approach zero or are negative, the gradient of the function becomes zero, the network cannot perform backpropagation and cannot learn. Another disadvantage of ReLU is the "dying ReLU" problem, where some neurons can become inactive and stop learning if they consistently receive negative inputs.

\subsection{Validation methods}

If all the data is used for training the model and the error rate is evaluated based on model's outcome compare to  actual value from the same training data set, this error is called the resubstitution error. This technique is called the resubstitution validation technique. Cross validation is a model evaluation technique that is more accurate than resubstitution. The problem with resubstitution evaluations is that they do not give an indication of how well the model will do when it is asked to make new predictions for data it has not already seen. One way to overcome this problem is to not use the entire data set when training a model. Some of the data is removed before training begins. Then when training is done, the data that was removed can be used to test the performance of the learned model on new data. This is the basic idea for a whole class of model evaluation cross validation technique.

In ML, we usually divide the dataset into Training dataset, Validation dataset, and Test dataset. The allocation of training, validation, and test data percentages depends on several factors, such as: the size of the dataset, the complexity of the problem, and the availability of data. However, some commonly used splits are as follows: The training data is used to train the machine learning model. It is the largest portion of the dataset and typically accounts for 60\% to 80\% of the data. A larger training set can allow the model to learn more effectively, especially for complex tasks. The validation data is used to tune the hyperparameters of the model and evaluate its performance during training. It helps in preventing over-fitting and selecting the best model configuration. The validation set is usually around 10\% to 20\% of the dataset. The test data is used to assess the final performance of the trained model. It provides an unbiased estimate of the model's generalisation capability on unseen data. The test set is typically around 10\% to 20\% of the dataset, similar to the validation set. Additionally, it is recommended to use techniques like cross-validation or stratified sampling when the dataset is limited or imbalanced.

\subsubsection{The Holdout Method}
The holdout method is the simplest kind of cross validation. The dataset is separated into two sets, called the training set and the testing set. The model is trained using the training set only. Then the model is asked to predict the output values for the data in the testing set (it has never seen these output values before). The errors it makes are accumulated as before to give the mean absolute test set error, which is used to evaluate the model. The advantage of this method is that it is usually preferable to the residual method and takes no longer to compute. However, its evaluation can have a high variance. The evaluation may depend heavily on which data points end up in the training set and which end up in the test set, and thus the evaluation may be significantly different depending on how the division is made.
\subsubsection{K-fold Cross Validation}
\begin{figure}
\centering\includegraphics[height=8cm, width=14cm]{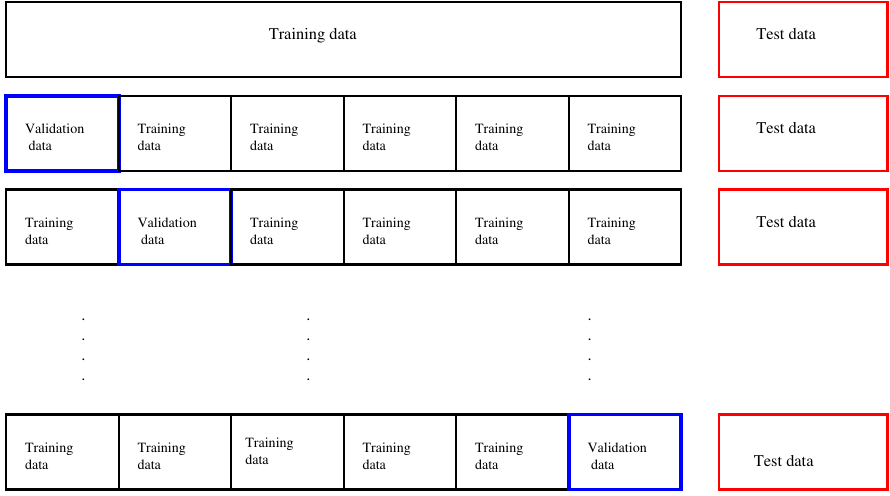}
\caption{K-fold cross validation }
\label{fig:kfold}
\end{figure}

K-fold cross validation is one way to improve over the holdout method. As \fig{kfold}illustrates, the data set is randomly divided into k subsets of approximately equal size, and the holdout method is repeated k times. Each time, one of the k subsets is used as the test set and the other k-1 subsets are put together to form a training set. Then the average error across all k trials is computed. The advantage of this method is that it matters less how the data gets divided, therefore, it provides a more robust estimate of the model's performance compared to a single train-test split. Every data point gets to be in a test set exactly once, and gets to be in a training set k-1 times. The variance of the resulting estimate is reduced as k is increased. K-fold cross-validation allows the model to be trained on a larger portion of the dataset, as each sample gets an opportunity to be part of the training and validation sets. K-fold cross-validation can be used to compare the performance of different models or different hyperparameter settings. By evaluating each model or configuration on multiple validation sets, it provides a fair comparison and helps in selecting the best-performing model. The disadvantage of this method is that the training algorithm has to be rerun from scratch k times, which means it takes k times as much computation to make an evaluation. A variant of this method is to randomly divide the data into a test and training set k different times. The advantage of doing this is that you can independently choose how large each test set is and how many trials you average over.

\subsubsection{Leave-one-out Cross Validation}
Leave-one-out cross validation (LOOCV) is K-fold cross validation taken to its logical extreme, with K equal to the number of the instance in the set $\mym$. That means that $\mym$ separate times, the model is trained on all the data except for one instance and a prediction is made for that instance. The average error is computed and used to evaluate the model. The evaluation given by LOOCV error is good, but it is expensive to compute. Fortunately, locally weighted learners can make LOO predictions just as easily as they make regular predictions. That means computing the LOOCV error takes no more time than computing the residual error and it is a much better way to evaluate models. Leave-one-out cross-validation is a special case of  cross-validation where the number of folds equals the number of  instances in the  data set. Thus, the learning algorithm is applied once for each instance, using all other instances as a  training set and using the selected instance as a single-item  test set. Thus, LOOCV is far less bias as we have used the entire dataset for training compared to the validation set approach where we use only a subset (60\% to 80\%) of the data for training. In addition, there is no randomness in the training or test data as performing LOOCV multiple times will yield the same results. However, MSE will vary as test data uses a single observation.

\section{ Regularisation}
There are three main ways to improve the performance of a model, to increase the training data, to increase the complexity of the network and to regularise the network \cite{regimpact}. All these three ways are related to each other and can improve the performance in combination to each other. Regularisation is a technique that is used to avoid over-fitting and improve the generalisation performance of a model \cite{optcom}. There are different types of regularisation methods utilised in literature (dropout, batch normalisation kernel regularisation and early stopping). The dropout method is the most commonly used regularisation technique for deep neural network, it can be implemented easily in CNN and is computationally cheap \cite{regimpact}. Moreover, batch normalisation has been emerged as another effective and strong regularisation method and has been utilised in many computer vision tasks. Kernel regularisation (L1 and L2) have been effectively applied in optimising the deep neural networks in the literature.

\subsection{Dropout}
Dropout handles the over-fitting issue by randomly dropping units from the neural network with their connections during training, which enables every neuron to work independently. The unit with all incoming and outgoing connections is removed temporarily from the network is called a dropout. The dropout technique is not applied during testing, it is only applied to input or hidden layer nodes and not output nodes \cite{regimpact} \cite{regsriv}.
\subsection{Batch Normalisation}
In deep neural networks, during training, the input of each layer changes due to parameters update of the previous layer, thus training slows down \cite{regimpact}. This phenomenon is called internal covariate shift (ICS), which is solved by normalising the input of the layer (batch normalising method). During training, each batch is normalised using much higher learning rate. Batch normalising 
not only reduces the over-fitting, but also improves the training by allowing higher learning rates and reducing the sensitivity to the initial starting weights. For convolutional layers, normalisation should follow the convolution property as well, means that different elements of the same feature map at various locations are normalised in a same way. Thus, all nodes activation in the mini batch are jointly normalised, over all the locations and parameters are learned per feature map not per node activation. Normalisation avoid the gradient vanish for high value learning rate, in addition \cite{regsantur} shows that batch normalisation adds smoothness to the internal optimisation problems of the network.

\subsection{Kernel Regularisation}
${\myL}_{1}$ regularisation method penalises the absolute value of the weights and tends to drive some weights exactly to
zero \cite{regimpact}. ${\myL}_{2}$ penalises the square value of the weights and tends to drive all weights to smaller
values. ${\myL}_{1}$ and ${\myL}_{2}$ regularisation can be combined and this combination is called Elastic Net
Regularisation. ${\myL}_{1}$ regularisation uses most important inputs and behaves invariantly to the
noisy ones. ${\myL}_{2}$ regularisation is preferable over ${\myL}_{1}$, because ${\myL}_{2}$ gives final weight vectors in
small numbers. ${\myL}_{2}$ regularisation is utilised more commonly in literature. Kernel regularisation has
produced excellent results in terms of accuracy when applied to the convolutional neural
networks for visual recognition tasks including hand written digits recognition, gender classification,
ethnic origin recognition and, object recognition \cite{reggong}. Kernel regularisation smooths the parameter distribution and reduces the
magnitude of parameters, hence resulting in less prone to over-fitting and effective solution. The idea of regularisation is to add an extra term to the loss function, the additional term is called the regularisation term \cite{NN1}.The difference between the ${\myL}_{1}$ and ${\myL}_{2}$ is just that the regularisation term for ${\myL}_{2}$ is the sum of the square of the weights, while for ${\myL}_{1}$ is just the sum of the weights. The original un-regularised MSE loss function is
\begin{eqnarray}
{\myL} &=& \frac{1}{2 \cal{M}} \sum_{m=1}^{\cal{M}} | {\vecy}^{(m)} -\hat{\vecy}^{(m)} |^{2}.
\end{eqnarray}
\\
Here's the ${\myL}_{1}$ regularised MSE loss function.

\begin{eqnarray}
{\myL}_{1} &=& \myL + \frac{\lambda}{2 \cal{M}} \sum_n |{\cal{W}}_{n}|,
\end{eqnarray}
\\
and the ${\myL}_{2}$ regularised MSE loss function
\begin{eqnarray}
{\myL}_{2} &=& \myL + \frac{\lambda}{2 \cal{M}} \sum_n {\cal{W}}_{n}^{2}
\end{eqnarray}
\\
$\lambda>0$ is known as the regularisation parameter and $\frac{\lambda}{2 \cal{M}} \sum_{n} {\cal{W}}_{n}^{2}$ is the ${\myL}_{2}$ regularisation term. Regularisation is a way of compromising between finding small weights and minimising the original loss function. Small $\lambda $ means we prefer to minimise the original loss function when large $\lambda $ means we prefer small weights. In order to find out how to apply the SGD learning algorithm in a regularised neural network we take the partial derivatives of ${\myL}_{2}$ with respect to $\myvec{\cal{W}}$, so we have:

\begin{eqnarray}
\frac{\partial {\myL}_{2}}{\partial \myvec{\cal{W}}} & = & \frac{\partial {\myL}}{\partial \myvec{\cal{W}}} + \frac{\lambda}{\cal{M}} \myvec{\cal{W}},\nonumber \\
\frac{\partial {\myL}_{2}}{\partial {\cal{\vecb}}} & = & \frac{\partial {\myL}}{\partial {\cal{\vecb}}}
\end{eqnarray}
\\
where $\myvec{\cal{W}}$ is the vector contains of all network's weights parameters and ${\cal{\vecb}}$ is the network's bias vector that contains of all network's biases. The $\frac{\partial {\myL}}{\partial \myvec{\cal{W}}}$ and $\frac{\partial {\myL}}{\partial {\cal{\vecb}}}$ terms can be computed using back propagation, so the gradient descent learning rule for the biases update doesn't change, but the learning rule for the weights becomes as:

\begin{eqnarray}
\myvec{\cal{W}}[t+1] &=& \myvec{\cal{W}} [t] -\eta \frac{\partial {\myL}}{\partial \myvec{\cal{W}}}-\frac{\eta \lambda}{\cal {M}} \myvec{\cal{W}} \nonumber \\
  & =&  \left(1-\frac{\eta \lambda}{\cal{M}}\right) \myvec{\cal{W}} -\eta \frac{\partial {\myL}}{\partial \myvec{\cal{W}}}.
\end{eqnarray}
\\
The $(1-\frac{\eta \lambda}{\cal{M}})$ is the re-scaling term which is referred to as the weight decay because it makes the weights smaller.

\subsection{Early Stopping}
 Early stopping is not a regularisation technique in the traditional sense, but it can be considered as a form of regularisation. It involves monitoring the model's performance on a validation set during training and stopping the training process when the performance starts to deteriorate. Early stopping prevents the model from over-fitting by finding the point at which it achieves the best trade-off between training and validation performance.

\section{Radar-based Hand Gesture Classification}

Radar-based hand gesture recognition has gained attention in recent years as a non-contact and robust method for human-computer interaction (HCI). A representative example is a technology that replaces a switch or remote control that requires existing physical contact with only a gesture [1]. However, while the importance of hand gesture recognition technology increases, the accuracy of hand gesture recognition technology is still insufficient \cite{gescnndo}. Several academic papers have explored this topic, presenting various techniques and approaches. Academic research in radar-based hand gesture recognition spans multiple disciplines, including signal processing, machine learning, computer vision, and human-computer interaction. Here is a summary of the key research areas related to radar-based hand gesture recognition in recent years:
\begin{enumerate}
\item Sensing Modality: Radar-based hand gesture recognition utilises radar systems to capture and analyse the reflected signals from hand movements. It offers advantages over other sensing modalities like vision-based systems, as radar can work in different lighting conditions and is not affected by occlusions \cite{Beltran-Hernandez2017} \cite{Anil2016}.
\item Doppler Signature Analysis: Radar systems can capture the Doppler signatures caused by the hand's motion, which carry valuable information about the gesture. Researchers focus on extracting and analysing these signatures to recognise specific gestures. Signal processing techniques, such as time-frequency analysis, Fourier analysis, or wavelet transforms, are commonly employed to analyse the radar signals \cite{8835661}.
\item Feature Extraction: Extracting discriminative features from radar signals is crucial for accurate gesture recognition. Various features have been explored, including statistical features (such as mean, variance), time-domain features (such as peak amplitude, duration), frequency-domain features (such as spectral energy, frequency components), and joint time-frequency features (such as time-frequency representation, spectrogram) \cite{FeatureExtraction}.
\item ML and Classification: ML algorithms play a significant role in radar-based gesture recognition. Researchers have employed various classification techniques, including traditional methods such as Support Vector Machines (SVM), k-Nearest Neighbors (k-NN), Random Forests, and more advanced techniques like deep learning-based approaches, including CNNs and Recurrent Neural Networks (RNNs) \cite{gescnnyek} \cite{gescnnsurv}.
\item Dataset Development: Building annotated datasets specifically designed for radar-based hand gesture recognition is an important aspect of academic research in this area. These datasets capture a wide range of hand gestures performed by different individuals in various scenarios. Researchers use these datasets to train and evaluate their gesture recognition models.
\item Real-Time Implementation: Real-time gesture recognition is crucial for practical applications. Therefore, researchers have focused on developing efficient algorithms and system architectures that can achieve real-time performance on resource-constrained platforms \cite{realtime}.
\end{enumerate}

\subsection{ML-based Methods for Hand Gesture Classification}
In radar-based hand gesture recognition using ML, there are several challenges that researchers face. First challenge is to build a well-annotated dataset specifically designed for radar-based hand gesture recognition can be challenging. Collecting a diverse range of hand gestures performed by multiple individuals in various scenarios requires careful planning and coordination. 

Second challenge is that the radar signals can be affected by various noise sources and interference, such as background noise, multipath reflections, and clutter which can effect the quality of the radar data. Third challenge is that radar signals captured during hand movements can have high dimensionality and variability. Different individuals may perform the same gesture with variations in speed, amplitude, or hand orientation. 

Forth challenge is the fact that compared to other modalities like vision-based systems, radar-based hand gesture recognition often has smaller datasets available for training and evaluation. Limited dataset sizes can impact the generalisation and performance of ML models. Finally it is worth mentioning that, in hand gesture recognition, there may be imbalanced distribution among different gesture classes. Some gestures may occur more frequently than others, leading to bias in the recognition system. 

 Recently researchers focus on addressing these challenges to improve the accuracy, robustness, and real-world applicability of ML-based radar hand gesture recognition systems. However, the literature about this topic is still very limited. Based on the type of sensor being used for data acquisition,
gesture recognition systems can largely be classified into two
classes: 1) wearable sensor based and 2) wireless sensor
based. Wearable sensor requires the user to attach the sensor to
their body \cite{kumar2020radar}. the FMCW radar has
widely been explored previously for several applications, such
as vital sign monitoring \cite{hashtom}, human gait analysis \cite{nohom} and
specifically, hand gesture recognition \cite{dahom}.

 Radar has recently shown its footprints for multiple target gesture recognition as well \cite{yazdahom}. Nowadays, devices such as Google Pixel 4
smartphone contains in-built radar sensor \cite{davazdahom} dedicated solely
for gesture recognition-based applications. \cite{kumar2020radar} Propose the opted multistream convolutional neural
network (MS-CNN) for in-air digit recognition using the FMCW radar sensor. A multistream CNN model capable of extracting information from
the range-time (RTM), Doppler-time (DTM), and angle-time (ATM) patterns was
proposed. The MS-CNN model combines different features
from multiple input streams simultaneously and concatenates
the features at the later stage that results in an overall better
performance in comparison to the tradition CNN approaches.

\subsection{CNN-based Methods for Hand Gesture Classification}
A common approach in radar hand-gesture recognition is to use CNN, which does not require predefined features, but rather, the network self-learns the features from input signals during the training process The majority of CNN-based hand-gesture recognition methods extract the signature from either the changes in Doppler over time, or from a snapshot of the overall range-Doppler fingerprint. Both of these signal types are represented in the form of a 2D matrix (monochromatic image) that is further processed by the CNN \cite{gescnnsurv}.  Recently, several neural network technologies have been studied, and results have been derived that CNN is easy to learn image data. Therefore, CNN was judged to be useful for classifying image data output by radar, so it was used for hand gesture identification \cite{gescnndo}.

 \cite{gescnnyek} Proposed a radar-based hand gesture recognition technique, which applies a CNN-based machine learning algorithm to time-domain I–Q plot trajectory images. The measurement data were analysed to evaluate the accuracy in recognizing six different hand gestures for the ten participants. Results indicate that the proposed technique can recognize hand gestures with average accuracy exceeding 90\%.

\cite{gescnndo} learning was conducted using proposes two CNN models. The first proposes a two-stage serial CNN model that learns by connecting two CNN terminals, and the second proposes a double parallel CNN model that connects two CNN terminals in parallel. However, VGG-19 and ALEXNET have higher accuracy than serial models, but lower accuracy than parallel CNN models. Moreover, a model with a shorter learning time is judged to be a more competitive model, and the model proposes a parallel model with less time execution and higher accuracy.

\cite{gescnnsurv} Compare 4 different classification architectures to predict the gesture class, namely: 1)fully connected neural network (FCNN), 2)k-Nearest Neighbours (k-NN), 3)support vector machine (SVM), 4)long short term memory (LSTM) network. The shape of the range-Doppler-frame tensor and the parameters of the classifiers are optimised in order to maximise the classification accuracy. The classification results of the proposed architectures show a high level of accuracy above 96 \% and a very low confusion probability even between similar gestures.

Nevertheless, \cite{gescnnse} propose TS-CNN, which includes the following three steps: 1)Design a CNN network to extract features from each Range-Doppler map. 2)Parallelly fuse features extracted from multiple Range-Doppler maps in the time series. 3)Add fully connected layer and softmax layer for feature classification. Experimental results show that the average gesture recognition accuracy of the TS-CNN method proposed in this paper is improved by 5\% compared to the accuracy of traditional machine learning methods, reaching 93\%

\chapter{ CV Datasets}

In this chapter, first we describe the three main types of existing radar systems, their specifications, challenges and their applications. Then we review the DopNet radar dataset, which is the hand gesture radar dataset utilised for the purpose of this report. The DopNet radar system specifications, gesture types, each sample's characteristics are discussed. Then we explain how each of our 2 binary CV datasets are created. Finally for each binary dataset, we present their number of samples and each samples specifications for each hand gesture.
\section{Radar Systems} 
Radar (Radio Detection and Ranging) systems are widely used in various applications to detect, track and locate objects in the surrounding environment. They work on the principle of sending out radio waves, which bounce off objects and return to the radar system. By analysing the characteristics of these returned signals, radar systems can provide valuable information about the location, speed, direction, and size of the detected objects.

We summarise the basic operations of a radar system as: transmitter, antenna, receiver and signal processor. Radar systems emit radio frequency(RF) signals, typically in the microwave or millimetre wave bands. A radar antenna directs the transmitted RF signal towards the target area and receives the reflected signals. The radar receiver amplifies and processes the received signals. The signal processor analyses the received signals to extract information about the target objects.

There are different radar types. Primary Radar, also known as active radar, sends out its own signals and detects the reflections from objects.
Whereas, secondary Radar(IFF), which is used in aviation for identifying and tracking aircraft, relies on targets transponding signals sent by the radar system.
On the other hand, doppler Radar measures the velocity of moving objects based on the Doppler effect. However, Synthetic Aperture Radar (SAR), which is used in remote sensing, produces high-resolution images of the Earth's surface by processing radar reflections.

There are some different types of radar signals. First type is pulse radar, which transmits short bursts or pulses of RF energy and measures the time it takes for the signal to return.
Second type, continuous wave (CW) radar, transmits a continuous wave and detects changes in frequency caused by the Doppler effect.
The third radar signal type is frequency-modulated continuous wave (FMCW) Radar, it uses a continuously varying frequency to measure range and velocity simultaneously.
\subsection{Pulse Radar System}

Pulse radar, also known as pulsed radar, is a type of radar system used for detecting and tracking objects by emitting short bursts of radio frequency(RF) energy, called pulses, and then analysing the returning echoes from those pulses.
Here are the key components and principles of pulse radar:
\begin{enumerate}
\item Transmitter: The radar transmitter generates short-duration RF pulses at a specific frequency. These pulses are typically high-powered to maximise their range and penetration capabilities.
\item Antenna: The radar antenna directs the pulses of RF energy into a specific direction. The antenna also collects the returning echoes (reflected signals) from the target objects.
\item Pulse Duration (Pulse Width): Pulse radar systems emit very short pulses, typically on the order of microseconds ($\mu$ s) to milliseconds (ms). The pulse duration determines the radar's ability to resolve targets at different distances.\item Pulse Repetition Frequency (PRF): PRF is the rate at which pulses are emitted from the radar transmitter. It is measured in Hertz (Hz) and affects the radar's ability to distinguish between multiple targets at different ranges. High PRF allows for better target discrimination but may limit the radar's maximum range.
\item Receiver: The radar receiver amplifies and processes the returning echoes, filtering out unwanted signals and noise. It measures the time delay between the transmitted pulse and the received echo to calculate the target's range.
\item Display and Data Processing: The processed radar data is displayed on a screen or used for further analysis. Modern radar systems often incorporate advanced signal processing techniques to improve target detection, track moving objects, and reduce interference.

\end{enumerate}
Pulse radar is commonly used in Air Traffic Control(ATC) systems to detect and track aircraft. Also pulse radar is essential for weather monitoring. They can detect precipitation, measure its intensity, and track the movement of weather systems. In addition pulse radar is used for various military applications, including target detection, tracking, and missile guidance. Pulse radar aids in maritime navigation by detecting other vessels, landmasses, and obstacles. Furthermore, it is employed for ground surveillance applications, such as border control and perimeter security.

\subsection{FMCW Radar System}
FMCW radar system is a special type of radar system that measures both velocity and distance of a moving object. FMCW, employs a continuous wave with a linearly increasing or decreasing frequency (known as a chirp). This is achieved by continuously transmitting a Chirp, which is a signal that increases (up-chirp) or decreases (down-chirp) its frequency linearly with time, This is then mixed with the received signal in order to obtain the range (Doppler) of a target.

Micro-Doppler is the additional signatures imparted onto the reflected signal back to the radar that a target generates. This movement creates a signature which was coined as Micro-Doppler by researcher V.Chen \cite{vchen}. This signatures are in addition to the bulk velocity and are created by vibration, rotation and other subtle movements. For example a person may walk at 3 m/s but as they move at this speed their arms and legs oscillate back and forth. there has been a recent growth of research evaluating the use of Doppler data to monitor human vital signs without the need for continuous contact between the senor and the subject \cite{dopnet}.  

FMCW radar system measures the frequency difference between the transmitted and received echo signal for calculating the distance, and it also measures the Doppler frequency (due to the Doppler effect) for calculating the speed of the object.  The key characteristics of FMCW radar include:

\begin{enumerate}

\item Frequency Modulation: FMCW radar transmits a modulated signal with a frequency sweep, usually a linear ramp. This modulation allows for range and velocity information extraction from the received signal.

\item Range and Velocity Measurement: FMCW radar measures range by comparing the frequency difference between the transmitted and received signals. It also utilises the Doppler effect to determine the target's velocity.
\item Range Resolution: FMCW radar offers excellent range resolution since it measures the frequency difference of the received signal over time. This enables the detection and differentiation of multiple targets at different ranges.

\item Complex Signal Processing: FMCW radar requires more sophisticated signal processing techniques, such as Fast Fourier Transform (FFT) and matched filtering, to extract range and velocity information accurately.

\end{enumerate}

 Applications: FMCW radar is commonly used in automotive radar systems, altimeters, and distance measurement devices due to its ability to provide accurate range and velocity measurements.

FMCW is widely used in the automotive industry for applications like adaptive cruise control, collision avoidance systems, and blind-spot monitoring. It can measure both range and relative velocity with high precision. Furthermore, FMCW radar is used in industrial settings for level measurement in tanks and process control. It can measure the distance to materials or objects accurately. Recently, FMCW radar finds use in aerospace applications for altimeter and collision avoidance systems in unmanned aerial vehicles (UAVs).
Nevertheless, FMCW radar can be employed in security systems for intrusion detection and surveillance, especially in scenarios where accurate range and velocity information are needed.

\subsection{CW Radar System}

CW (continuous wave) radar operates by transmitting a continuous wave of radio frequency energy without any modulation. It emits a steady signal continuously while listening for the reflected echoes. The key characteristics of CW radar include:
\begin{enumerate}
\item Simplicity: CW radar systems are relatively simple, as they involve continuous transmission and reception without any need for modulation or frequency sweeping.

\item Range Measurement: CW radar can measure the range to a target by measuring the time delay between the transmitted signal and the received echo. However, it does not provide any information about the target's velocity or range rate.

\item Doppler Effect: CW radar is particularly useful for measuring the Doppler frequency shift caused by the relative motion between the radar and the target. This allows for velocity measurement and target detection.

\item Limited Range Resolution: CW radar has limited range resolution capabilities due to its continuous wave nature. It cannot accurately resolve multiple targets at different ranges, and it lacks the ability to distinguish between closely spaced objects.

\end{enumerate}

In summary, CW radar is simpler and useful for measuring Doppler shifts, while FMCW radar provides better range resolution and allows for simultaneous range and velocity measurements. The choice between these two techniques depends on the specific application requirements and the level of complexity and accuracy needed.
Radar sensors have previously been successfully used to classify different actions such as walking, carrying an item, discriminating between people and animals gaits or drones and bird targets \cite{fiorone}~\cite{fiortwo}~\cite{fiorthree}~\cite{tah}. 

\section{DopNet Radar Datasets}

Radar sensors have a new growing application area of dynamic hand gesture recognition. Traditionally radar systems are considered to be very large, complex and focused on detecting targets at long ranges. With modern electronics and signal processing it is now possible to create small compact RF sensors that can sense subtle movements over short ranges. For such applications, access to comprehensive databases of signatures is critical to enable the effective training of classification algorithms and to provide a common baseline for benchmarking purposes. 

DopNet is a large Radar database organised in a hierarchy, each node in DopNet represents the data of one person which is divided into different gestures recorded from that person. The data is measured with FMCW and CW Radars. DopNet‘s structure makes it a useful tool for ML gesture recognition, software and Image Processing for the spectrograms. The shared data was generated by Dr. Matthew Ritchie (University College London (UCL)) and Richard Capraru (Nanyang Technological University (NTU) and Singapore Agency for Science, Technology and Research (A*STAR), Singapore) within the UCL Radar Research Group in collaboration with Dr. Francesco Fioranelli (Delft University of Technology (TU Delft), Delft, Netherlands). Furthermore, it started as a Laidlaw Scholarship project \cite{dopnet}.

\begin{figure}
\centering\includegraphics[scale=0.9]{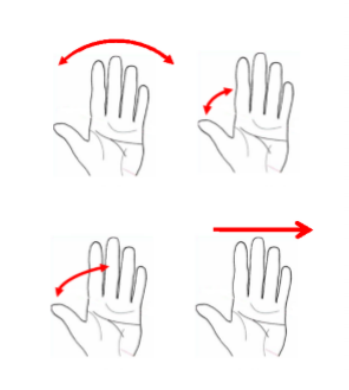}
\caption{ Wave, Pinch, Click, Swip gestures \cite{dopnet}}
\label{fig:gestu}
\end{figure}

A database of gestures has been created which includes signals from $4$ different types Wave, Pinch, Click, Swipe for person A, B, C, D, E and F\fig{gestu}. The data itself has been pre-processed so that the signatures have been cut into individual actions from a long data stream, filtered to enhance the desired components and processed to produce the Doppler vs. time-domain data. The data is then stored in this format in order for it to be read in, features to be extracted and the classification process to be performed \cite{dopnet}.

\begin{table}
\centering\scalebox{0.9}{  
 \begin{tabular}{|ccccccc|ccccccc|r}
  \hline
  \hline
  &&&Parameter &&&&& &Value&&&& \\
  \hline
  \hline
  &&&Frequency &&&&&& $24$GHz&&&&  \\
  \hline
  &&&Bandwidth &&&&&& $750$MHz&&&& \\
  \hline
  &&&ADC bits &&&&&& $12$&&&&\\
  \hline
  &&&TX power& &&&&& +$13$dBm&&&&\\
  \hline
  \hline
 \end{tabular}}
\caption{The DopNet radar system specifications}  
\label{tbl:gest}  
\end{table}
The data generated for this classification challenge was created using an Ancortek $24$ GHz FMCW radar a $750$ MHz bandwidth, more details about the radar system can be seen in the \tbl{gest}.

The comparative FMCW radar system that has been used for this work is the Ancortek SDR-KIT 2400AD2. This is a $24$ GHz device that has up to $2$ GHz bandwidth (although this was set to 750MHz for the data used within this report) and has been set to $1$ ms chirp period. It has $+13$ dBm power and used $14$dBm horn antennas. The sensor has a standalone GUI to control and capture data or can be commanded within a Matlab interface to capture signals. The system has one transmit and two receive antennas (only one was used for the purposes of this dataset). It was set up on a lab bench at the same height as the gesture action. It was then initiated to capture $30$ seconds of data and the candidate repeated the actions numerous times within this window. Afterwards, the raw data was then cut into individual gestures that occurred over the whole period.

For data gathering repeated hand gestures were made approximately $30-40$ cm away from the radar over a long continuous period. The single data file generated was then cut into individual gesture actions for feature extraction and classification processing.
 These individual gesture actions have varying matrix sizes hence a cell data format was used to create a ragged data cube. The data was created by the following flow of processing:
\begin{itemize}
\item De-interleave Channel $1$ $-2$ and I$/$Q samples.
\item Break vector of samples into a $2$D matrix of Chirp vs. Time.
\item FFT samples to covert to range domain. Resulting in a Range vs. Time matrix (RTI)
\item Filter signal such that static targets are suppressed and moving targets are highlighted. This is called MTI filtering in radar signal processing.
\item Extract rows within the RTI that contain the gesture movement and coherently sum these.
\item Generate a Doppler vs. Time $2$D matrix by using a Short Time Fourier Transform on the vector of selected samples.
\item Store the complex samples of the Doppler vs. Time matrix within a larger cell array which is a data cube of the N repeats of the $4$ gestures from each person.
\end{itemize}

\begin{figure}
\centering\includegraphics[scale=0.6]{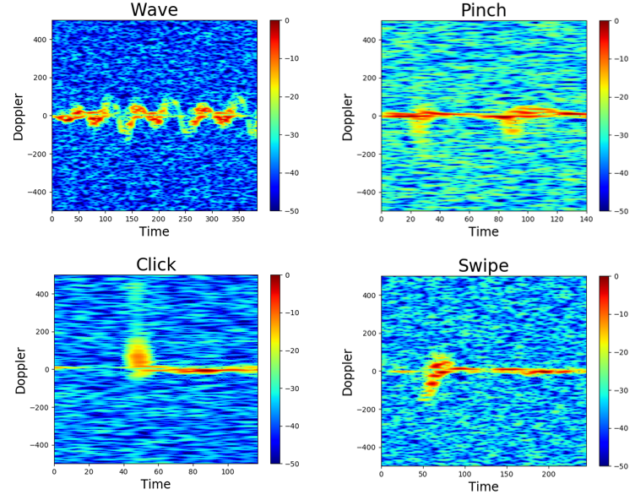}
\caption{Doppler vs. Time matrix for each of four gestures \cite{dopnet} }
\label{fig:fourgest}
\end{figure}
Example of a Doppler vs. Time matrix for each gesture can be seen in\fig{fourgest}. The waving gesture which has the oscillatory shape and longer duration, whereas the click gesture happens over the shortest time frame (as a click is only a short sharp action). Then the pinch and swipe actions do show some level of similarity which could make them challenging for a classifier \cite{dopnet}. \tbl{fourgesture} illustrates the number of samples per person. Each sample contains a $2$D matrix of complex-valued numbers, the first dimension (Doppler) of all samples'matrices size are the same and equal to $800$, and the second dimension of smples'matices (Time) size is not the same for all the samples, the range of the Time dimensions for each gesture are shown in \tbl{gestdim}.

\begin{table}[htbp]
\centering
 \begin{tabular}{|c|c|c|c|c|c|c|r|}
  \hline
  \hline
  Gesture & A & B & C & D & E  & F & Total \\
  \hline
  Wave & 56 & 112 & 85 & 70 & 56 & 87 &  466\\
  \hline
  Pinch & 98 & 116 & 132 & 112 & 98 & 140 & 696\\
  \hline
  Swipe & 64 & 72 & 80 & 71 & 91 & 101 & 479\\
  \hline
  Click & 105 & 105 & 137 & 93 & 144 & 208 & 792\\
  \hline
  \hline
 \end{tabular}
\caption{The number of each gesture's samples per person }  
\label{tbl:fourgesture}  
\end{table}

\begin{table}
\centering\scalebox{0.9}{  
 \begin{tabular}{|cccccccc|ccccc|c|}
  \hline
  &&&Gesture &&&&&&& Dimension range &&\\
  \hline
  \hline
  &&&Wave &&&&&&& 116 - 540&&\\
  \hline
  &&&Pinch &&&&&&& 60 - 231&&\\
  \hline
  &&&Swipe &&&&&&& 118 - 480&&\\
  \hline
  &&&Click &&&&&&& 43 - 211&&\\
  \hline
  \hline
 \end{tabular}}
\caption{The range of each gesture's Time dimension  }  
\label{tbl:gestdim}  
\end{table}

\section{Our proposed CV Datasets}
We have created two CV datasets from the Dopnet hand gesture samples. The number of samples for each gesture was different, as \tbl{fourgesture} and \tbl{gestdim} show, wave and swipe gestures samples have similar dimension range ($540$ and $480$) and number of samples ($466$ and $479$), the remaining $2$ gestures (pinch and click) have similar dimension range ($231$ and $211$) and number of samples ($692$ and $792$). Therefore, we chose wave and swipe gesture samples to create complex-$1$ dataset and chose pinch and click gesture samples to create complex-$2$ dataset.

For complex-$1$ dataset, we uniformed all the samples dimensions to $800$ $\times$ $540$ by adding zeros wherever the time dimension was less than $540$. Therefore, we have got $945$ CV samples in our database. The second CV dataset (complex-$2$) consists of $1488$ samples, the dimension of the samples in this dataset is $800$ $\times$ $231$, zeros are added wherever the time dimension was less than $231$. 

We used $80$ percent of the samples in each of the datasets as training and $20$ percent as test samples. So we have:

\begin{itemize}
\item complex-1 dataset: \\
$945$ samples, \\
sample size is $800$ $\times$ $540$ \\
\item complex-2 dataset: \\
$1488$ samples, \\
sample size is $800$ $\times$ $231$ 

\end{itemize}

Wheres, sample size $800$ $\times$ $540$ for complex-$1$ dataset, describes that each hand gesture sample in this database consists of $800$ $\times$ $540$ CV pixels, which each pixel has a real and imaginary components. Similarly, each sample in complex-$2$ dataset, has $800$ $\times$ $231$ CV pixels.

\chapter{Complex-Valued CNN (CV-CNN)}
\label{cha:ytwocomplexcnn}

CNNs have become one of the most accurate techniques in image recognition, segmentation and detection tasks \cite{survey11}. CNNs can improve the learning and extraction of invariant representation by utilising multiple mapping functions. The multiple functions also enable the CNN to recognise hundreds of categories. The hierarchical learning, automatic feature extraction, weight sharing and multi-tasking are the key advantages of the CNN \cite{survey}. 

However most researchers develop and utilise the RV CNN blocks, as there is still a gap in the literature for the fully CV-CNN building blocks. Many researchers attempt to develop CV blocks that is able to compute a CV input through the training or BP. However, the most adopted technique is to separate the real and imaginary components of each input data sample, then to feed them into two parallel RV-CNNs or concatenate the real and imaginary components. This results in doubling the size of each input sample and therefore it will be computationally more expensive.

On the other hand, some researchers\cite{complexnn} take a different approach, the approach is to separate the real and imaginary parts and apply the simulated CV convolutional arithmetic function bu using RV arithmetic. The simulation can be implemented by utilising the existing convolutional CV-forward python libraries. Although the real part and imaginary components are separated and the all the computations only deals with RVs, the convolutional operation is simulates the CV convolutional operation. However using the popular programming languages' ML libraries such as Python ignores the effect of the CV numbers and their derivatives on BP of each layer of CNN.  

With the knowledge of the author today, there is a gap in the literature for a fully CV-CNN building blocks and precise mathematical operations which includes each layer's CV operation, each layer's CV-BP operations and their derivatives. This chapter explains the architecture of our proposed CV-CNN including detailed specifications, dimensions and operation of each layer forward and backward. We implement the architecture of the suggested CV-CNN in this chapter in Python language from scratch and without the use of any machine learning libraries. Our contribution is the detail mathematical explanations of a fully CV-CNN with CV input data, CV operations in each step of every layer forward and backward and implementation of the python code from scratch.

This chapter first defines each layer of the proposed CV-CNN building block and its interaction with the connecting layers. Then denotes the mathematical operation of each block, then explains each step of the BP derivatives with all utilised vectors and matrices dimensions of each operation. We also explain the selected CV loss function and activation function and validation technique.  At the end of this chapter, we display the training and test results of implementing our fully CV-CNN on two CV radar images datasets (complex-$1$ and complex-$2$). We compare our proposed CV-CNN's results with the equivalent RV-CNN model with same architecture, settings and CV datasets as a baseline.

\section{Proposed CV-CNN }

\begin{figure}
\centering\includegraphics[width=\textwidth, height= 12cm]{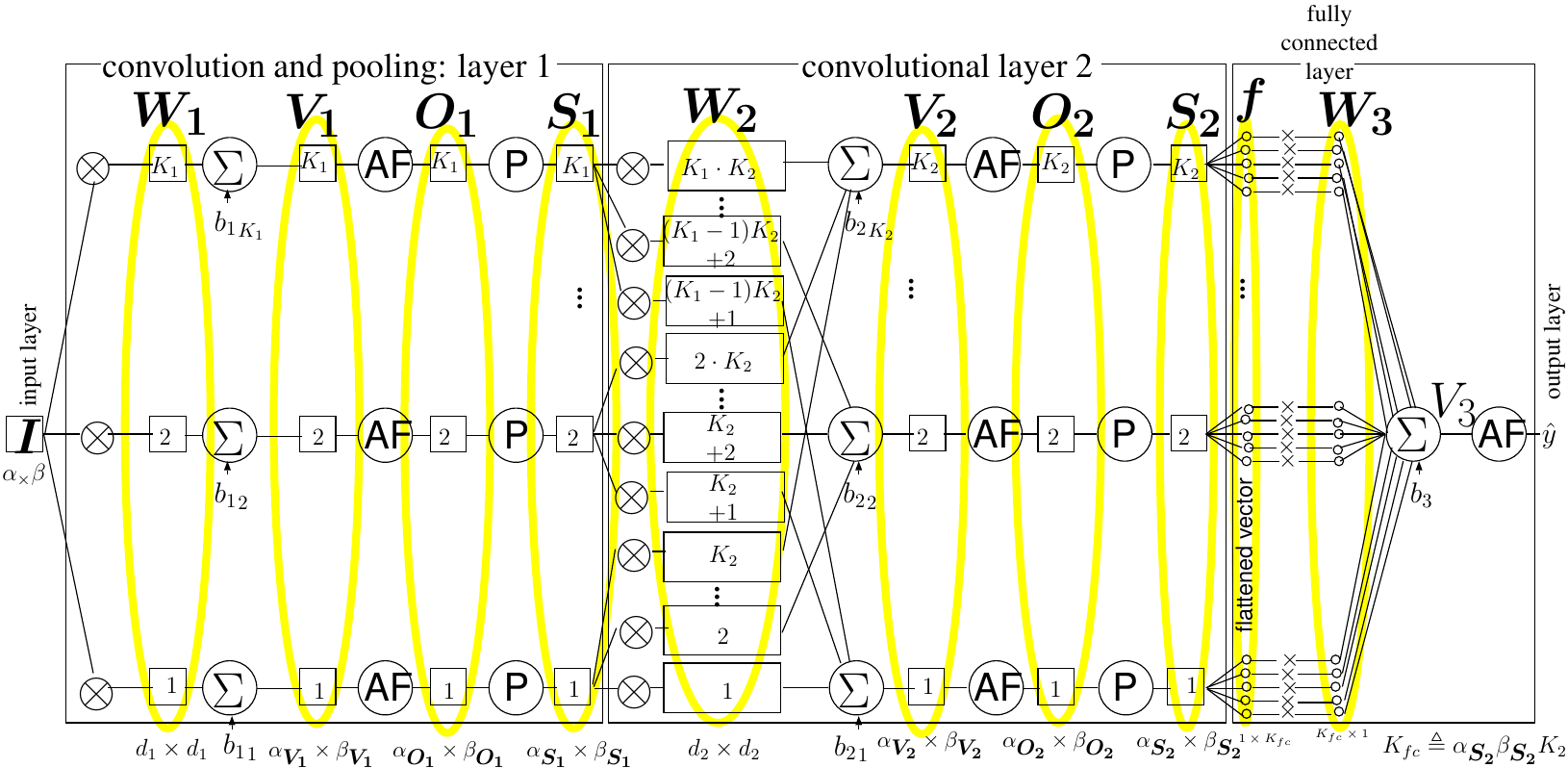}
\caption{The two layer CV-CNN architecture.}
\label{fig:cnnone}
\end{figure}

Generally, in order to train a CNN we use $\mym$ sample pairs of $(\Xvar^{(m)} , {\vecy}^{(m)})$, where $1\leq m \leq \mym$ and ${\Xvar}^{(m)}, {\vecy}^{(m)}$ denote the $m$-th sample input matrix and its vector label respectively. The architecture of the proposed CV-CNN is illustrated in\fig{cnnone}, where we have two convolutional layers and one fully connected layer. In this research, we assume that $\Ivec$ is a sample of input matrix $(\Ivec = {\Xvar}^{(m)})$, which can be any 2D image (RV or CV), we use 2D CV radar images, where $\Ivec \in \mathbb{C} ^{\alpha \times \beta}$ and $y^{(m)}$ is the $m$-th label which is the corresponding hand gesture class, so it is a scalar value. 

${\vecW}_{1}$ is the first convolutional layer (Conv1)'s weight tensor, where ${\vecW}_{1} \in \mathbb{C} ^{\dl{1} \times \dl{1} \times \kl{1}}$, the weight tensor consists of $\kl{1}$ CV kernel map of $\vecwonekone$ where $\vecwonekone \in \mathbb{C} ^{\dl{1} \times \dl{1}}$, where $1 \leq \kpl{1} \leq \kl{1}$. In addition to this, Conv1 layer has its bias vector ${\myvec{b}}_{1}$, which is a CV vector and ${\myvec{b}}_{1} \in \mathbb{C} ^{\kl{1} \times 1 }$ which consists of $\kl{1}$ CV bias scalar values of $\bonekone$. Second convolutional layer (Conv2)'s weight tensor is ${\vecW}_{2}$, where ${\vecW}_{2} \in \mathbb{C} ^{\dl{2} \times \dl{2} \times (\kl{1} \times \kl{2})}$ consists of $\kl{1} \times \kl{2}$ CV kernel map matrices of $\vecwtwoktwo \in \mathbb{C} ^{\dl{2} \times \dl{2}}$ where $1 \leq \kpl{2} \leq \kl{2}$ and $\wtwoktwo$ is the $( (\kpl{1}-1) \cdot \kl{2} +\kpl{2} )$-th $\myvec{W}_{2}$'s plane. In addition, conv2 layer's bias vector ${\myvec{b}}_{2}$, is a CV vector where ${\myvec{b}}_{2} \in \mathbb{C} ^{\kl{2} \times 1 }$ consist of $\kl{2}$ CV bias scalar values of $\btwoktwo$. Whereas fully connected layer's CV weight matrix $\vecwthree \in \mathbb{C} ^{1 \times \kl{fc}}$  (\sect{fc}) and a CV scalar value bias $b$.  The output $\haty$ is the predicted hand gesture class and it is a CV scalar.

\subsection{Initialisation of the Parameters}
We initialise the weight tensors and matrices of ${\vecW}_{1}$, ${\vecW}_{2}$ and $\vecwthree$ with normalised random numbers (in the range of $[0,1]$) and the biases ${\myvec{b}}_{1}$, ${\myvec{b}}_{2}$  and $b$ with zero.

\subsection{First Convolutional Layer (Conv1)}
$\vecvonekone$ is the Conv1's $\kpl{1}$-th
 feature map, which is computed as a convolutional operation between each input image ($\Ivec$) matrix and the $\kpl{1}$-th first layer's kernel map $\vecwonekone$. The $\kappa$-th CV output matrix of the first convolutional layer $\vecoonekone$ is the result of applying the activation function on the $\kpl{1}$-th weighted matrix $\vecvonekone$. The Conv1 output matrix ${\myvec{O}}_{1}$ consists of $\kl{1}$ CV matrices of $\vecoonekone$ where $1 \leq \kpl{1} \leq \kl{1}$ , as in:

\begin{eqnarray}
\label{eq:apyekcon}
\vecoonekone   & = &  \sigma (\vecvonekone) \nonumber \\
&=& \sigma (\Ivec \otimes {\vecwonekone} + \bonekone)
\end{eqnarray}
\\
where $\otimes$ represents the convolutional operation and $\sigma$ is the activation function.

\begin{figure}
\centering\includegraphics[width=\textwidth, height=5cm]{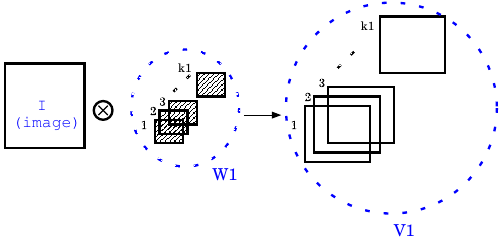}
\caption{First forward convolution- layer one.}
\label{fig:fwconv}
\end{figure}

In the first convolutional layer, we have $\kl{1}$ kernel maps, thus we get $\kl{1}$ feature maps $(\myvec{V}_{1})$. The image matrix dimensions are $\alpha \times \beta$, the kernel maps ${\vecW}_{1} \in \mathbb{c} ^{\dl{1} \times \dl{1} \times \kl{1}}$, and ${\myvec{V}}_{1} \in \mathbb{C} ^{\alphl{V^{1}} \times \betl{V^{1}} \times \kl{1}}$ where $\alphl{V_{1}} = (\alpha - \dl{1} +1)$ and $\betl{V_{1}} =(\beta - \dl{1} +1)$ because we set the convolutional padding and stride parameters to zero and one respectively. $\vecoonekone$ and the $\vecvonekone$ are 2D matrices and have the same dimensions, so we have $\alphl{V_{1}} = \alphl{O_{1}}$ and $\betl{V_{1}} = \betl{O_{1}}$  .\eq{elmoone} computes the first convolutional operation element-wise as in:

\begin{eqnarray}
\label{eq:elmoone}
\oonekone (i,j)  & = &  \sigma \Bigl (\sum _{u=0}^{\dl{1}-1}  \sum _{v=0}^{\dl{1}-1} ( {I}(i-u,j-v) \cdot {\wonekone}(u,v) + \bonekone) \Bigr ) \nonumber \\
& = &  \sigma \Bigl (\sum _{u=0}^{\dl{1}-1}  \sum _{v=0}^{\dl{1}-1} ( {I}(i+u,j+v) \cdot {\wonekone}(-u,-v) + \bonekone) \Bigr )\nonumber \\
& = &  \sigma \Bigl( \sum _{u=0}^{\dl{1}-1}  \sum _{v=0}^{\dl{1}-1} ( {I}(i+u,j+v) \cdot \rotl{\wonekone}(u,v) + \bonekone) \Bigr)\nonumber \\
& = & \sigma \Bigl( \sum _{u=0}^{\dl{1}-1}  \sum _{v=0}^{\dl{1}-1} \Bigl( \Re (I(i+u,j+v)) \cdot \Re(\rotl{\wonekone}(u,v))\nonumber \\
&-& \Im (I(i+u,j+v)) \cdot \Im(\rotl{\wonekone}(u,v))    + \Re(\bonekone) \Bigr ) \nonumber \\
& +&  \jmath \Bigl(\Re (I(i+u,j+v)) \cdot \Im(\rotl{\wonekone}(u,v)) \nonumber\\
& + & \Im (I(i+u,j+v)) \cdot \Re(\rotl{\wonekone}(u,v))    + \Im(\bonekone) \Bigr)\Bigr)
\end{eqnarray}
\\
\fig{fwconv} illustrates the first forward convolution's equation.

\subsection{First Pooling Layer \texorpdfstring{($\vecs_{1}$)}{(s1)}}

In this research we assume the dimensions of the pooling window is $g \times g$ and we set the pooling stride to $g$, therefore in this stage we replace each $g \times g$ window of the Conv1 output matrix with the scalar value of the average of the window as in:

\begin{eqnarray}
{S_{1}}_{\kpl{1}} (i,j) &=& \frac{1}{g^{2}}  \sum_{u=0}^{g}  \sum_{v=0}^{g} \oonekone (i  \times u +i,j \times v+j)
\end{eqnarray}
\\
where ${\vecs}_{1} \in \mathbb{C}^{\alphl{S_{1}} \times \betl{S_{1}} \times \kl{1}}$, so we have $i = 1,2,...\frac{1}{g}(\alphl{O_{1}})$ and $j = 1,2,...\frac{1}{g} (\betl{O_{1}})$.

\subsection{Second Convolutional Layer (Conv2)}
The $\vecvtwoktwo$ is the Conv2's $\kpl{2}$-th feature map. The $\kpl{2}$-th feature map matrix of the second convolutional layer ($\vecvtwoktwo$) is computed as a convolution between $\myvec{S_{1}}$ and $\kpl{2}$-th second layer's kernel maps ($\vecwtwoktwo$). Whereas the output of the Conv2 layer ($\vecotwoktwo$) is the result of applying the activation function on the feature maps as in

\begin{eqnarray}
\label{eq:apdocon}
\vecotwoktwo   & = &  \sigma (\vecvtwoktwo) \nonumber \\
&=& \sigma (\sum ^{\kl{1}}_{\kpl{1}=1} {{\vecs}_{1}}_{\kpl{1}} \otimes {\vecwtwoktwo} + \btwoktwo)
\end{eqnarray}
\\


\begin{figure}
\centering\includegraphics[width=\textwidth, height=7cm]{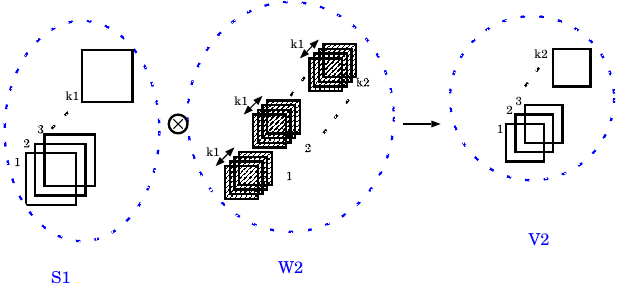}
\caption{Second forward convolution.}
\label{fig:fwconvtwo}
\end{figure}
\fig{fwconvtwo} illustrates the second forward convolutional operation as in\eq{apdocon}. We compute the element-wise second convolutional operation as in:
\begin{eqnarray}
\label{eq:convone}  
\otwoktwo (i,j)  & = &  \sigma \Bigl( \sum ^{\kl{1}}_{\kpl{1}=1} \sum _{u=0}^{\dl{2}-1}  \sum _{v=0}^{\dl{2}-1} {S_{1}}_{\kpl{1}} (i-u,j-v) \cdot {\wtwoktwo}(u,v) + \btwoktwo \Bigr) \nonumber \\
  & = &  \sigma \Bigl( \sum ^{\kl{1}}_{\kpl{1}=1} \sum _{u=0}^{\dl{2}-1}  \sum _{v=0}^{\dl{2}-1} {S_{1}}_{\kpl{1}} (i+u,j+v) \cdot {\wtwoktwo}(-u,-v) + \btwoktwo \Bigr) \nonumber \\
  & = &  \sigma \Bigl( \sum ^{\kl{1}}_{\kpl{1}=1} \sum _{u=0}^{\dl{2}-1}  \sum _{v=0}^{\dl{2}-1} {S_{1}}_{\kpl{1}} (i+u,j+v) \cdot \rotl{\wtwoktwo}(u,v) + \btwoktwo \Bigr) \nonumber \\
  & = &  \sigma \Bigl(\sum ^{\kl{1}}_{\kpl{1}=1} \sum _{u=0}^{\dl{2}-1}  \sum _{v=0}^{\dl{2}-1} \Bigl( \mathfrak{R}({S_{1}}_{\kpl{1}} (i+u,j+v)) \cdot  \mathfrak{R}(\rotl{\wtwoktwo}(u,v)) \nonumber\\
&-& \Im({S_{1}}_{\kpl{1}} (i+u,j+v)) \cdot  \Im(\rotl{\wtwoktwo}(u,v)) + \Re( \btwoktwo) \Big) \nonumber \\
    &+& \jmath \Bigl( \Re({S_{1}}_{\kpl{1}} (i+u,j+v)) \cdot  \Im(\rotl{\wtwoktwo}(u,v)) \nonumber \\
    &+& \Im({S_{1}}_{\kpl{1}} (i+u,j+v)) \cdot  \Re(\rotl{\wtwoktwo}(u,v)) + \Im( \btwoktwo) \Big)\Bigr)
\end{eqnarray}
\\
Where $\kpl{2}= 1,2,...\kl{2}$. The dimensions of $\vecotwoktwo$ matrix ($\alphl{O_{2}} \times \betl{O_{2}}$) can be computed from ${{\vecs}_{1}}_{\kpl{1}}$ dimensions considering the padding and stride parameters are set to zero and one respectively. The Conv2 kernel map's($\wtwoktwo$) dimensions are  $\dl{2} \times \dl{2}$. Thus we have  $\alpha_{O_{2}} = (\alphl{S_{1}} - \dl{2} +1)$ and $\betl{O_{2}} =(\betl{S_{1}} - \dl{2} +1)$. $\vecotwoktwo$ and $\vecvtwoktwo$ matrices have the same dimensions so we have $\alphl{O_{2}}=\alphl{V_{2}}$ and $\betl{O_{2}}=\betl{V_{2}}$.

\subsection{\texorpdfstring{First Pooling Layer (${ \vecs_{1} }$)}{First Pooling Layer}}

In this stage we assume the pooling window dimensions and the pooling stride are $g \times g$ and $g$ respectively, so we replace each $g \times g$ window of the Conv2's output matrix $\vecotwoktwo$ with the scalar value of the average of the window as in:
\begin{eqnarray}
{S_{2}}_{\kpl{2}}(i,j) &=& \frac{1}{g^{2}}  \sum_{u=0}^{g-1}  \sum_{v=0}^{g-1} \otwoktwo (i \times u +i,j \times v + j)
\end{eqnarray}
\\
we have ${\vecs}_{2} \in \mathbb{C}^ {\alphl{S_{2}} \times \betl{S_{2}} \times \kl{2}}$ and each ${{\vecs}_{2}}_{\kpl{2}} \in \mathbb{C}^ {\alphl{S_{2}} \times \betl{S_{2}}}$  where $i,j = 1,2, \cdots \frac{1}{g}(\alphl{O_{2}})$ , $\frac{1}{g}(\betl{O_{2}})$.

\subsubsection{Vectorisation and Concatenation}
\label{sec:fc}

First each ${{\vecs}_{2}}_{\kpl{2}}$ is vectorized by column then all of the $\kl{2}$ vectors are concatenated to form a vector $\vecf$ with the size of $\alphl{S_{2}} \times  \betl{S_{2}} \times \kl{2} = \kl{fc}$. We denote the process of the vectorisation and the concatenation of function F, thus we have:

\begin{eqnarray}
\label{eq:vectorisation}
\vecf &=& F (\{ {{\vecs}_{2}}_{\kpl{2}} \} ) \nonumber \\
F^{-1}(\vecf) &=& \{ {{\vecs}_{2}}_{\kpl{2}} \}
\end{eqnarray}
\\
where $1 \leq \kpl{2} \leq \kl{2}$.

\subsubsection{Fully Connected Layer}
The scalar $\vthree$ is the weighted value of the fully connected layer before the activation function.
\begin{eqnarray}
\vthree &=& \vecwthree \times \vecf + \bthree \nonumber \\
&=&  \Re(\vecwthree)\times \Re( \vecf) -  \Im(\vecwthree)\times \Im( \vecf)+ \Re(\bthree) \nonumber \\
&+&  \jmath \Bigl(  \Re(\vecwthree)\times \Im( \vecf) + \Im(\vecwthree)\times \Re( \vecf)+ \Im(\bthree)    \Bigr)
\end{eqnarray}
The scalar value $\hat {y}$, is the predicted value for the hand gesture class, which is the network output, is calculated as:

\begin{eqnarray}
\haty &=& \sigma (\vthree) \nonumber \\
\end{eqnarray}
\\
where $\sigma$ is the activation function, ${\vecwthree} \in \mathbb{C}^{1 \times \kl{fc}}$ and $\vecf \in \mathbb{C}^{\kl{fc} \times 1}$.


\subsection{Back Propagation (BP)}

The network weight and bias parameters are trained using the SGD technique by computing the loss gradient with respect to each weight and bias parameter, we define ${\myL}^{(m)}$ as a CV differentiable loss function for $m$-th training sample as in:

\begin{eqnarray}
\label{eq:zerobp}
{\myL}^{(m)} & = & \frac {1}{2} \mid {y}^{(m)} - {\haty}^{(m)} \mid ^{2}
\end{eqnarray}
\\
where ${y}^{(m)}$ and ${\haty}^{(m)}$ are the $m$-th label and predicted output scalar values respectively. In the field of CV differentiability, in order for a CV function to be differentiable it has to satisfy both two conditions of Couchy-Riemann equations. We compute the average loss for all the sample pairs as in:
\begin{eqnarray}
\label{eq:1bp}
\myL & = & \frac {1}{2 \mym} \sum^{\mym}_{m=1} \mid {y}^{(m)} - {\haty}^{(m)} \mid ^{2}
\end{eqnarray}
\\
\eq{1bp} satisfies two Couchy-Riemann's conditions, Where we can compute $\mid {y} - {\haty} \mid ^{2}$ as:

\begin{eqnarray}
\mid {y} - {\haty} \mid ^{2} &=& \mid {y}^{2} +{\haty}^{2} - 2 y \cdot {\haty} \mid \nonumber \\
&=& \mid (\Re(y))^{2} - (\Im(y))^{2} + \jmath [2 \Re(y) \cdot \Im(y)\nonumber] \\
&+& (\Re(\haty))^{2} - (\Im(\haty))^{2} + \jmath [2 \Re(\haty) \cdot \Im(\haty)]\nonumber \\
&-& 2\bigl[ \Re(y) \cdot \Re(\haty) + \jmath \bigl( \Re(y) \cdot \Im(\haty)+ \Im(\vecy) \cdot \Re(\haty) \bigr )\nonumber \\
&&-  \Im(y) \cdot \Im(\haty)\bigr] \mid
\end{eqnarray}
\\

Thus we have:
\begin{eqnarray}
 \Re(\mid {y} - {\haty}\mid ^{2}) &=& \mid (\Re(y))^{2} - (\Im(y))^{2} +  (\Re(\haty))^{2} \nonumber \\
&-& (\Im(\haty))^{2} - 2 \Re(y) \cdot \Re(\haty) +2  \Im(\vecy) \cdot \Im(\haty)\mid
\end{eqnarray}
\\and:
\begin{eqnarray}
 \Im(\mid {y} - {\haty}\mid ^{2}) &=& \mid 2 \Re(y) \cdot \Im(y)+ 2 \Re(\haty) \cdot \Im(\haty)\nonumber \\
&-& 2 \Re (y) \cdot \Im(\haty)- 2 \Im(y) \cdot \Re(\haty)\mid
\end{eqnarray}
\\
In BP we compute $\myL$'s partial derivatives with respect to each layer's parameter backwards, which means from the output layer towards input layer, then based on the computed partial derivatives we update the weights and biases parameters. As such we compute partial derivative of $\myL$ with respect to weights and biases as $\nabll{\vecwthree}$, $\nabll{\bthree}$, $\nabll{{\vecW_{2}}_{\kpl{1},\kpl{2}}}$, $\nabll{{{\vecb}_{2}}_{\kpl{2}}}$, $\nabll{{{\vecW}_{1}}_{\kpl{1}}}$ then $\nabll{{\myvec{b}_{1}}_{\kpl{1}}}$. Let $\zeta$ be a function of vector $\myvec{\theta}= [\theta_{1},\theta_{2}, \cdots \theta_{n}]$ we define $\nabla_{\myvec{\theta}}^{\zeta}=\frac {\partial \zeta}{\partial \myvec{\theta}}$, so $\nabla_{\myvec{\theta}}^{\zeta}$ has a partial derivative of $\frac {\partial {\zeta}}{\partial \theta_{i}}$ with respect to each variable $\theta_{i}$ when $1 \leq i \leq n$. The partial derivatives define the gradient vector of $\zeta$ with respect to $\myvec{\theta}$ as in:

\begin{eqnarray}
{\nabla}_{\myvec{\theta}}^{\zeta}(a) &=&  \Bigl[ \frac{\partial \zeta} {\partial {\theta}_{1}(a)},\ldots \frac{\partial \zeta}{\partial {\theta}_{n} (a)}  \Bigr]
\end{eqnarray}
\\
first we compute the derivation of the loss function in respect with the network output $\haty$ as in:

\begin{eqnarray}
\nabll{\haty}&=& \frac {\partial \myL}{\partial \haty}  =  \frac{\partial \Re({\myL})}{\partial \Re(\haty)}+ \jmath  \frac{\partial \Im({\myL})}{\partial \Re(\haty)}
\end{eqnarray}
\\
so we have:

\begin{eqnarray}
\label{eq:lyhat}
\nabll{\haty} &=&\frac {\partial \myL}{\partial \haty}= \mid \Re (y) - \Re( {\haty}) \mid + \jmath \mid \Im (y) - \Im( {\haty}) \mid
\end{eqnarray}

\subsubsection{\texorpdfstring{Loss Gradient with Respect to $\vecwthree$ ($\nabll{\vecwthree} $)}{Loss Gradient with Respect to}}
The dimensions of the  $\nabll{\vecwthree}$ is same as $\vecwthree$, which is $1 \times \kl{fc}$, applying the chain rule and Couchy-Reimann equations we have

\begin{eqnarray}
\label{eq:lwyek}
\nabll{\wthree} (1,i)  &=& \frac {\partial \myL}{\partial \haty} \cdot \frac {\partial \haty}{\partial  \vthree } \cdot \frac {\partial \vthree} {\partial \wthree (1,i)}
\end{eqnarray}

\begin{eqnarray}
\label{eq:lwse}
\frac  {\partial \haty} {\partial  \vthree}&=&\frac{\partial \sigma(\vthree)}{\partial \vthree}
\end{eqnarray}
\\
\begin{eqnarray}
\label{eq:lwchar}
\frac {\partial \vthree}{\partial \wthree (1,i)}&=& \frac{\partial \Re(\vthree)}{\partial \Re(\wthree (1,i))} +\jmath \frac {\partial \Im \vthree}{\partial \Re(\wthree (1,i))}\nonumber \\
&=& \Re f(i) + \jmath \Im f(i) \nonumber \\
&=& f(i)
\end{eqnarray}
\\

We replace the derivations in\eq{lwyek} with\eq{lwse} and\eq{lwchar}, so we have:
\begin{eqnarray}
\label{eq:lwpanj}
\frac {\partial \haty} {\partial  \wthree (1,i)}&=&  \frac {\partial {\sigma} (\vthree)}{\partial \vthree} \cdot f(i)
\end{eqnarray}
\\
\eq{lwpanj} and\eq{lwyek} we have:

\begin{eqnarray}
\nabll{{\myvec{W}}_{3}}&=& \nabll{\haty} \cdot \frac {\partial {\sigma}(\vthree)}{\partial \vthree}\cdot {\vecf}^{T}
\end{eqnarray}
\\

The dimensions of ${\vecf}^{T}$ and $\nabll{\vecwthree}$ are also $1 \times \kl{fc}$.

\subsubsection{\texorpdfstring{Loss Gradient with Respect to $\bthree$ ($\nabll{\bthree}$)}{Loss Gradient with Respect to}}
$\nabll{b}$ is a CV scalar value, so we have:

\begin{eqnarray}
\nabll{\bthree} &=& \frac {\partial {\myL}}{\partial \bthree}  = \frac{\partial {\myL}}{\partial \haty} \cdot \frac {\partial \haty} {\partial \vthree} \cdot \frac{\partial \vthree}{\partial  \bthree}   \nonumber \\
&=& \nabll{\haty} \cdot \frac {\partial \sigma(\vthree)}{\partial \vthree} \cdot \frac{\partial \vthree}{\partial \bthree} \nonumber \\
&=&  \nabll{\haty} \cdot \frac {\partial \sigma(\vthree)}{\partial \vthree} \cdot ( \frac{\partial \Re(V)}{\partial \Re(b)}+ \jmath \frac{\partial \Im(V)}{\partial \Re(b)}) \nonumber \\
&=&  \nabll{\haty} \cdot \frac {\partial \sigma(\vthree)}{\partial \vthree}.   
\end{eqnarray}
\\

\subsubsection{\texorpdfstring{Loss Gradient with Respect to ${\vecW}_{2}$ ($\nabll{{\vecW}_{2}}$) }{Loss Gradient with Respect to}}
$\nabll{\vecwtwoktwo}$ is the $((\kpl{1} -1) \cdot \kl{2} +\kpl{2})$ th plane of $\nabll{{\vecW}_{2}}$, where $1 \leq \kpl{2} \leq \kl{2}$. Dimensions of each  $\nabll{ \vecwtwoktwo}$ plane is $\dl{2} \times \dl{2}$ which is same as $\vecwtwoktwo$. First we compute $\nabll{\vecf}$ and $\nabll{{{\vecs}_{2}}_{\kl{2}}}$, then we compute the $\nabll{\vecotwoktwo}$ and finally the $\nabll{\vecwtwoktwo}$ accordingly. So we have:

\begin{eqnarray}
\nabll{f}(i) &=& \frac {\partial \myL}{\partial f(i)} \nonumber \\
&=& \frac {\partial \myL}{\partial \haty} \cdot \frac {\partial \haty}{\partial \vthree} \cdot \frac{\partial \vthree}{\partial f(i)} \nonumber \\
&=& \nabll{\haty} \cdot \frac{\partial \sigma (\vthree)}{\partial \vthree} \cdot (\frac {\partial \Re(\vthree)}{\partial \Re(f(i))} + \jmath \frac {\partial \Im(\vthree)}{\partial \Re(f(i))}) \nonumber \\
&=& \nabll{\haty} \cdot \frac {\partial \sigma(\vthree)}{\partial \vthree}  \cdot \Bigl (\Re(\wthree(1,i)) + \jmath \Im(\wthree(1,i)) \Bigr) \nonumber \\
&=& \nabll{\haty} \cdot \frac {\partial \sigma(\vthree)}{\partial \vthree}  \cdot \wthree (1,i) 
\end{eqnarray}
\\
So we have:
\begin{eqnarray}
\nabll{\vecf}  &=& {\vecwthree}^{T} \cdot \nabll{ \haty}\cdot \frac {\partial \sigma(\vthree)}{\partial \vthree} 
\end{eqnarray}
\\
where ${\vecW}^{T}$ dimensions are $\kl{fc} \times 1$, so $\nabll{\vecf}$'s dimensions are also $\kl{fc} \times 1$ which is the same as $\vecf$. We reshape the 1D vector $\nabll{\vecf}$  by:

\begin{eqnarray}
\vecf &=& F ({{\vecs}_{2}}_{\kpl{2}}) \nonumber \\
F^{-1}(\nabll{ \vecf}) &=& \{ \nabll{{{\vecs}_{2}}_{\kpl{2}}} \}
\end{eqnarray}
\\
where $\kpl{2}=1,2,..{\kl{2}}$, thus we get $\kl{2}$ planes on ${\myvec{S}_{2}}$ layer with the $\alphl{S_{2}} \times \betl{S_{2}}$ dimensions. As there are no parameters in the ${\myvec{S}}_{2}$ layer, we do not need to compute the derivation of the ${S_{2}}_{\kpl{2}}$. In order to obtain the ${\nabll{ \vecotwoktwo}}$ on the second convolutional layer, we perform up-sampling on $\nabll{\myvec{S}_{2}}$ planes, so we have:

\begin{eqnarray}
\nabll{ \otwoktwo} (i,j) &=& \frac {1}{g^{2}} \nabll{ {{\vecs}_{2}}_{\kpl{2}}} (\lceil \frac {i}{g} \rceil , \lceil \frac {j}{g} \rceil)
\end{eqnarray}
\\
where $i=1,2, \cdots \alphl{\otwoktwo}$, $j=1,2, \cdots \betl{\otwoktwo}$ and $\nabll{\vecotwoktwo}$'s dimensions are ($\alphl{O_{2}} \times \betl{O_{2}}  $. The $\lceil. \rceil$ denotes the ceiling function and $ \nabll{ {{\vecs}_{2}}_{\kpl{2}}}$'s dimensions are $\frac {1}{g^{2}}$ of  $\nabll{ \vecotwoktwo}$. In this stage, as we have already computed $\nabll{ \vecotwoktwo}$, we can finally calculate the $\nabll{\vecwtwoktwo}$ as in

\begin{eqnarray}
\label{eq:nabwdo}
\nabll{ \wtwoktwo}(u,v) &=&  \frac {\partial \myL}{\partial \wtwoktwo(u,v)} \nonumber \\
&=& \sum_{i=1}^{\alphl{O_{2}}} \sum_{j=1}^{\betl{O_{2}}} \frac {\partial \myL}{\partial \otwoktwo(i,j)} \cdot \frac {\partial \otwoktwo(i,j)}{\partial \vtwoktwo (i,j)}\cdot \frac {\partial \vtwoktwo(i,j)}{\partial \wtwoktwo(u,v)} \nonumber \\
&=& \sum_{i=1}^{\alphl{O_{2}}} \sum_{j=1}^{\betl{O_{2}}} \nabll{ \otwoktwo}(i,j) \cdot \frac{\partial \sigma(\vtwoktwo (i,j))}{\partial \vtwoktwo(i,j)} \nonumber \\
&&\cdot \Bigl( \frac {\partial \Re(\vtwoktwo(i,j))}{\partial \Re(\wtwoktwo(u,v))}+ \jmath \frac {\partial \Im(\vtwoktwo(i,j))}{\partial \Re(\wtwoktwo(u,v))} \Bigr) \nonumber \\
&=& \sum_{i=1}^{\alphl{O_{2}}} \sum_{j=1}^{\betl{O_{2}}}  \nabll{ \otwoktwo}(i,j) \cdot \frac {\partial \sigma(\vtwoktwo(i,j))}{\partial \vtwoktwo(i,j)} \nonumber \\
&&\cdot \Bigl( \Re( {S_{1}}_{\kpl{1}} (i-u,j-v)) + \jmath \Im ( {S_{1}}_{\kpl{1}} (i-u,j-v)) \Bigr) \nonumber \\
&=&  \sum_{i=1}^{\alphl{O_{2}}} \sum_{j=1}^{\betl{O_{2}}}  \nabll{ \otwoktwo}(i,j) \cdot  \frac {\partial \sigma(\vtwoktwo(i,j))}{\partial \vtwoktwo(i,j)} \cdot {S_{1}}_{\kpl{1}} (i-u,j-v)
\end{eqnarray}
\\
In order to simplify the\eq{nabwdo}, we use:
\begin{eqnarray}
\nabll{\vtwoktwo}(i,j) &=& \nabll{ \otwoktwo}(i,j) \cdot  \frac {\partial \sigma(\vtwoktwo(i,j))}{\partial \vtwoktwo(i,j)} 
\end{eqnarray}
\\
Which we can display as:
\begin{eqnarray}
\nabll{\myvec{\vtwoktwo}} &=& \nabll{\myvec{\otwoktwo}} \odot  \frac {\partial \sigma(\myvec{\vtwoktwo})}{\partial \myvec{\vtwoktwo}} 
\end{eqnarray}

\noindent where $\odot$ is the element-wise multiplying operation and we know that \\ $ \rotl{{S_{1}}_{\kpl{1}}} (u-i,v-j)= {S_{1}}_{\kpl{1}} (i-u,j-v)$, so:

\begin{eqnarray}
\nabll{ \wtwoktwo}(u,v) &=& \sum_{i=1}^{\alphl{O_{2}}} \sum_{j=1}^{\betl{O_{2}}}  \rotl{{S_{1}}_{\kpl{1}}} (u-i,v-j) \cdot \nabll{ \vtwoktwo} (i,j)
\end{eqnarray}
\\
therefore, we have:

\begin{eqnarray}
\label{eq:deltaw2} 
\nabll{ \vecwtwoktwo} &=&  \rotl{{{\vecs}_{1}}_{\kpl{1}}} \otimes  \nabll{ \vecvtwoktwo}
\end{eqnarray}
\\
\begin{figure}
\centering\includegraphics[width=\textwidth, height=15cm]{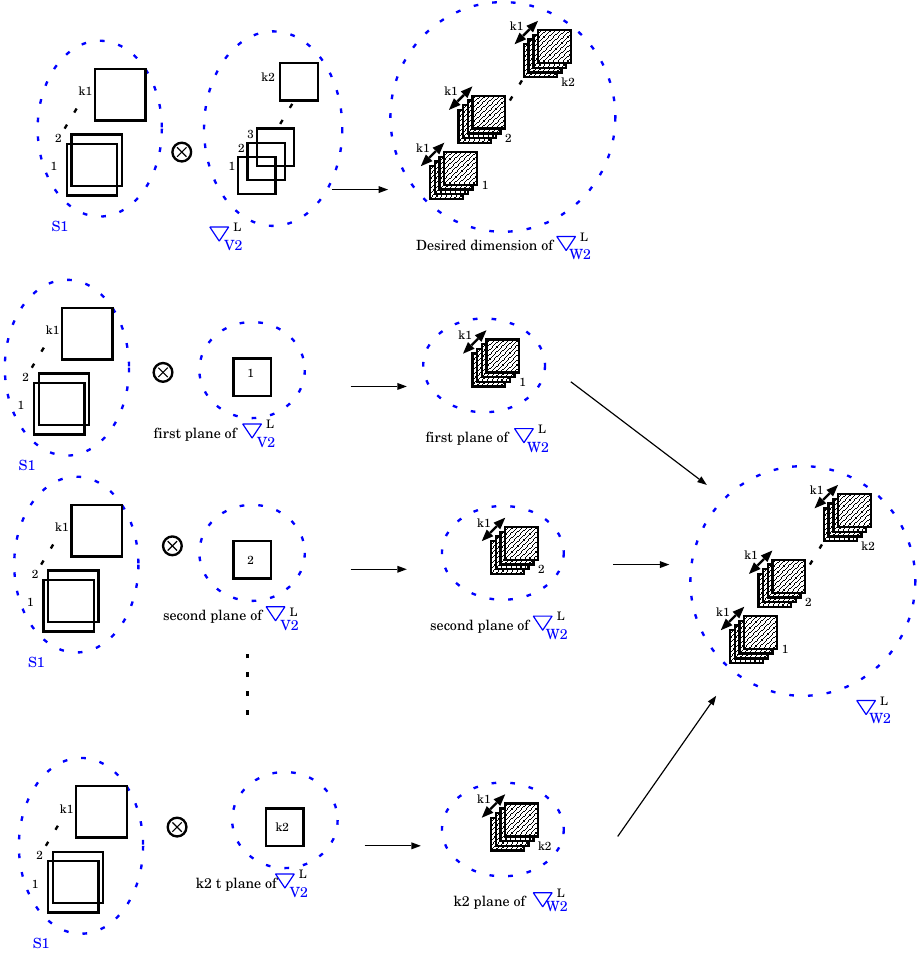}
\caption{First back propagation convolution details.}
\label{fig:bwconv}
\end{figure}
\\
\fig{bwconv} illustrates the first back propagation convolutional operation as in\eq{deltaw2} with details.

\subsubsection{\texorpdfstring{Loss Gradient with Respect to $\myvec{b}_{2}$ ($\nabll{ \myvec{b}_{2}} $)}{Loss Gradient with Respect to}}
The dimensions of $ \nabll{{\vecb}_{2}} $ is $\kl{2} \times 1$ consist of $\kl{2}$ scalar values of $\nabll{ \btwoktwo}$ where $1 \leq \kpl{2} \leq \kl{2}$ we have:

\begin{eqnarray}
\nabll{ \btwoktwo} &=&  \frac {\partial \myL}{\partial \btwoktwo} \nonumber \\
&=& \sum_{i=1}^{\alphl{O_2}} \sum_{j=1}^{\betl{O_2}} \frac {\partial \myL}{\partial \otwoktwo(i,j)} \cdot \frac {\partial \otwoktwo(i,j)}{\partial \vtwoktwo (i,j)} \cdot \frac {\partial \vtwoktwo (i,j)}{\partial \btwoktwo} \nonumber \\
&=& \sum_{i=1}^{\alphl{O_2}} \sum_{j=1}^{\betl{O_2}} \nabll{\otwoktwo}(i,j) \cdot \frac {\partial \sigma (\vtwoktwo(i,j))}{\partial \vtwoktwo (i,j)} \cdot \frac {\partial \vtwoktwo (i,j)}{\partial \btwoktwo} \nonumber \\
&=& \sum_{i=1}^{\alphl{O_{2}}} \sum_{j=1}^{\betl{O_{2}}} \nabll{ \otwoktwo}(i,j) \cdot  \frac {\partial \sigma(\vtwoktwo(i,j))}{\partial \vtwoktwo(i,j)} \nonumber \\
&=& \sum_{i=1}^{\alphl{O_{2}}} \sum_{j=1}^{\betl{O_{2}}} \nabll{ \vtwoktwo}(i,j)
\end{eqnarray}
\\

\subsubsection{\texorpdfstring{Loss Gradient with Respect to ${\vecW}_{1}$ ($\nabll{ {\vecW}_{1}} $)}{Loss Gradient with Respect to}}
The dimensions of $\nabll{{\vecW}_{1}} $ are $\dl{1} \times \dl{1} \times \kl{1}$, where $\vecwonekone$ is the $\kpl{1}$-th plane of $\nabll{ {\vecW}_{1}}$  and $1\leq \kpl{1} \leq \kl{1}$.  In order to compute the $\nabll{ \vecwonekone} $, first we need to obtain $\nabll {{{\vecs}_{1}}_{\kpl{1}}}$ and $ \nabll{ \vecoonekone}$. Therefore, we have:

\begin{eqnarray}
\nabll{{S_{1}}_{\kpl{1}}(i,j)} &=&  \frac {\partial \myL}{\partial S ^{1}_{\kpl{1}}(i,j) } \nonumber \\
&=& \sum_{\kpl{2}=1}^{\kl{2}}  \sum_{u=0}^{\dl{1}-1} \sum_{v=0}^{\dl{1}-1}  \frac {\partial \myL}{\partial \vtwoktwo (i+u, j+v)} \cdot \frac {\partial \vtwoktwo (i+u, j+v)}{\partial {S_{1}}_{\kpl{1}}(i,j) } \nonumber \\
&=& \sum_{\kpl{2}=1}^{\kl{2}}  \sum_{u=0}^{\dl{1}-1} \sum_{v=0}^{\dl{1-1}} \nabll{ \vtwoktwo} (i+u, j+v) \cdot \frac {\partial}{\partial {S_{1}}_{\kpl{1}}(i,j)} \nonumber \\
& & \bigl( \sum_{\kpl{1}=1}^{\kl{1}}  \sum_{u=0}^{\dl{1}-1} \sum_{v=0}^{\dl{1}-1}  {S_{1}}_{\kpl{1}}(i,j) \cdot \wtwoktwo (u,v) + \btwoktwo \bigr) \nonumber \\
&=& \sum_{\kpl{2}=1}^{\kl{2}}  \sum_{u=0}^{\dl{1}-1} \sum_{v=0}^{\dl{1}-1} \nabll{ \vtwoktwo} (i+u, j+v) \cdot \wtwoktwo (u,v) 
\end{eqnarray}
\\

We know that $ \rotl{\wtwoktwo} (-u,-v) = \wtwoktwo (u,v)$, therefore we have:

\begin{eqnarray}
\nabll{{S_{1}}_{\kpl{1}}}(i,j) &=&  \sum_{\kpl{2}=1}^{\kl{2}}  \sum_{u=0}^{\dl{1}} \sum_{v=0}^{\dl{1}} \nabll{\vtwoktwo} (i-(-u),j-(-v)) \cdot \nonumber \\
&& \rotl{\wtwoktwo} (-u,-v)
\end{eqnarray}
\\
so we have:

\begin{eqnarray}
\label{eq:apdobpcon}
\nabll{{\vecs_{1}}_{\kpl{1}}} &=&  \sum_{\kpl{2}=1}^{\kl{2}} \nabll{\vecvtwoktwo} \otimes \rotl{\vecwtwoktwo}
\end{eqnarray}
\\
\begin{figure}
\centering\includegraphics[width=\textwidth, height=16cm]{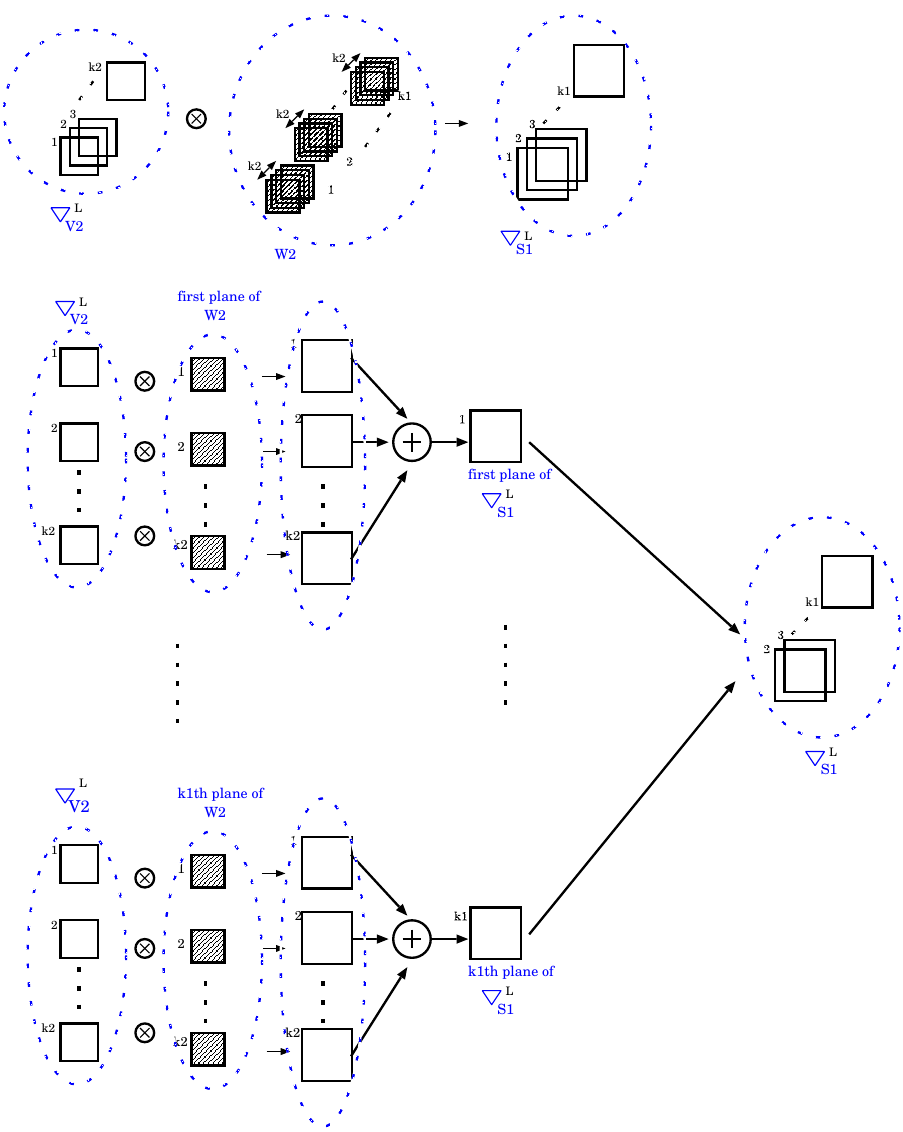}
\caption{Second back propagation convolution.}
\label{fig:bwconvtwo}
\end{figure}
\\
\fig{bwconvtwo} displays the second back propagation convolution as in\eq{apdobpcon} with details. In order to compute the loss derivative with respect to $\oonekone$ we need to up-sample the pooling layer's error maps, as follows:

\begin{eqnarray}
\nabll{\oonekone} (i,j) &=& \frac {1}{g^{2}} \nabll{ {{\vecs}_{1}}_{\kpl{1}}} (\lceil \frac {i}{g} \rceil , \lceil \frac {j}{g} \rceil)
\end{eqnarray}
\\
where $i=1,2,\cdots \alphl{O_{1}}$ and $j= 1,2, \cdots \betl{O_{1}}$. Now we can compute the $\nabll{\vecwonekone} $, therefore 

\begin{eqnarray}
\nabll{ \wonekone}(u,v) &=&  \frac {\partial \myL}{\partial \wonekone(u,v)} \nonumber \\
&=& \sum_{i=1}^{\alphl{O_{1}}} \sum_{j=1}^{\betl{O_{1}}} \frac {\partial \myL}{\partial \oonekone(i,j)} \cdot \frac {\partial \oonekone(i,j)}{\partial \vonekone (i,j)} \cdot \frac {\partial \vonekone (i,j)}{\partial \wonekone(u,v)} \nonumber \\
&=& \sum_{i=1}^{\alphl{O_{1}}} \sum_{j=1}^{\betl{O_{1}}} \nabll{\oonekone}(i,j) \cdot \frac {\partial \sigma(\vonekone(i,j))}{\partial \vonekone (i,j)} \cdot \frac {\partial \vonekone (i,j)}{\partial \wonekone(u,v)} \nonumber \\
&= &\sum_{i=1}^{\alphl{O_{1}}} \sum_{j=1}^{\betl{O_{1}}} \nabll{ \oonekone}(i,j) \cdot \frac {\partial \sigma(\vonekone(i,j))}{\partial \vonekone(i,j)}  \cdot \nonumber \\
&& \Bigl( \frac{\partial \Re(\vonekone(i,j))}{\partial \Re(\wonekone(u,v))}+\jmath \frac{\partial \Im(\vonekone(i,j))}{\partial \Re(\wonekone(u,v))} \bigr) \nonumber \\ 
&=& \sum_{i=1}^{\alphl{O_{1}}} \sum_{j=1}^{\betl{O_{1}}} \nabll{ \oonekone}(i,j) \cdot  \frac {\partial \sigma(\vonekone(i,j))}{\partial \vonekone(i,j)}  \cdot \nonumber \\
&&\Bigl( \Re (I(i-u,j-v))+ \Im (I(i-u,j-v)) \Bigr) \nonumber \\
&=& \sum_{i=1}^{\alphl{O_{1}}} \sum_{j=1}^{\betl{O_{1}}} \nabll{ \oonekone}(i,j) \cdot \frac {\partial \sigma(\vonekone(i,j))}{\partial \vonekone(i,j)}  \cdot I (i-u,j-v)
\end{eqnarray}
\\
We rotate $I$, 180 degrees and we have:

\begin{eqnarray}
\nabll{ \vonekone} (i,j) &=& \nabll{ \oonekone}(i,j) \cdot \frac {\partial \sigma(\vonekone(i,j))}{\partial \vonekone(i,j)} 
\end{eqnarray}
\\
which means:

\begin{eqnarray}
\nabll{\myvec{\vonekone}} &=& \nabla{\myvec{\oonekone}} \odot {\sigma}^{\prime} {(\myvec{\vonekone})}
\end{eqnarray}
\\
thus:

\begin{eqnarray}
\nabll{ \wonekone}  (u,v) &=&   \sum_{i=1}^{\alphl{O_{1}}} \sum_{j=1}^{\betl{O_{1}}} {I}_{\mathrm{rot180}} (u-i,v-j) \cdot \nabll{ \vonekone} (i,j).
\end{eqnarray}
\\
So we have:

\begin{eqnarray}
\label{eq:apsebpcon}
\nabll{\myvec{\wonekone}} &=& {\Ivec}_{\mathrm{rot180}} \otimes \nabll {\myvec{\vonekone}}
\end{eqnarray}
\\

\begin{figure}
\centering\includegraphics[width=\textwidth, height= 9cm]{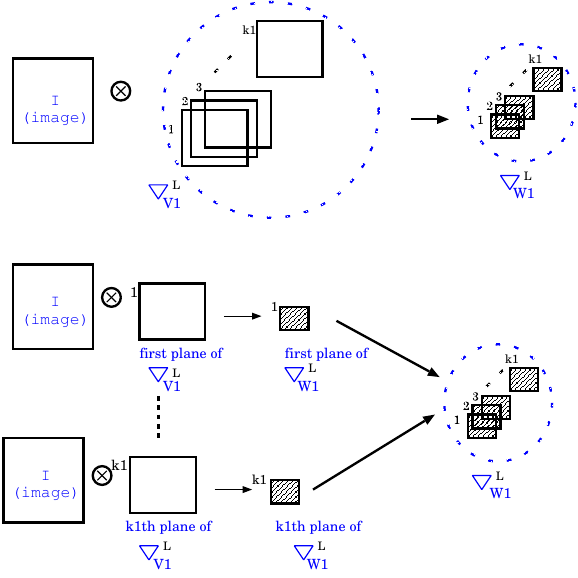}
\caption{Third back propagation convolution.}
\label{fig:bwconfour}
\end{figure}

\fig{bwconfour} displays the third back propagation convolution as in\eq{apsebpcon} with details.
\subsubsection{\texorpdfstring{Loss Gradient with Respect to $\vecbonekone$ ($\nabll{\vecbonekone} $)}{Loss Gradient with Respect to}}
The dimensions of $\nabll {\myvec{b}_{1}} $ is $ \kl{1} \times 1$.  For each $\nabll{ \bonekone}$ where $1 \leq \kpl{1} \leq \kl{1}$ we have:

\begin{eqnarray}
\nabll{\bonekone} &=&  \frac {\partial \myL}{\partial \bonekone} \nonumber \\
&=& \sum_{i=1}^{\alphl{O_{1}}} \sum_{j=1}^{\betl{O_{1}}} \frac {\partial \myL} {\partial \oonekone(i,j)} \cdot \frac {\partial \oonekone(i,j)}{\partial \vonekone (i,j)} \cdot \frac {\partial \vonekone (i,j)}{\partial \bonekone} \nonumber \\
&=& \sum_{i=1}^{\alphl{O_{1}}} \sum_{j=1}^{\betl{O_{1}}} \nabll {\oonekone} \cdot \frac {\partial \sigma(\vonekone(i,j))}{\partial \vonekone (i,j)} \cdot \frac {\partial \vonekone (i,j)}{\partial \bonekone} \nonumber \\
&=& \sum_{i=1}^{\alphl{O_{1}}} \sum_{j=1}^{\betl{O_{1}}} \nabll{ \oonekone}(i,j) \cdot \frac {\partial \sigma(\vonekone(i,j))}{\partial \vonekone(i,j)}  \nonumber \\
&=& \sum_{i=1}^{\alphl{O_{1}}} \sum_{j=1}^{\betl{O_{1}}} \nabll{ \vonekone}(i,j)
\end{eqnarray}
\\
thus, we have:
\begin{eqnarray}
\nabll{ \vecbonekone}&=&  \sum_{i=1}^{\alphl{O^{1}}} \sum_{j=1}^{\betl{O^{1}}} \nabll{ \vecvonekone}
\end{eqnarray}

\subsection{Parameters Update}

Following computing the partial derivative of loss function with respect to weights and bias in each layer, we can update the parameters after each iteration, If the number of samples are very high, usually the entire dataset is not passed into the network at once, the training dataset is divided into mini-batches.  Batch size is the total number of training samples present in a single min-batch.  An iteration is a single gradient update of the network's weights and biases during training.  The number of iterations is equivalent to the number of batches needed to complete one epoch. We need to set the value of the learning rate ($\lr$) so we can update parameters accordingly as in:

\begin{eqnarray}
\vecW [t+1] & = & \vecwthree [t] + \Delta \vecwthree [t] \nonumber \\
\vecb [t+1] & = & b_{3} [t] + \Delta b_{3} [t] \nonumber \\
\vecwonekone [t+1] & = &  \vecwonekone[t] + \Delta \vecwonekone[t] \nonumber \\
\vecbonekone [t+1] & = & \vecbonekone[t] + \Delta \vecbonekone[t] \nonumber \\
\vecwtwoktwo [t+1] & = & \vecwtwoktwo [t]+ \Delta \vecwtwoktwo [t] \nonumber \\
\vecbtwoktwo [t+1] & = & \vecbtwoktwo [t]+ \Delta \vecbtwoktwo [t]
\end{eqnarray}
\\
where $t$ denote the iteration number.

\begin{eqnarray}
 \Delta \vecwthree [t] &=& - \lr \nabll{ \vecwthree} [t] \nonumber \\
 \Delta {b}_{3} [t] & =& - \lr \nabll{ b_{3}} [t] \nonumber \\
 \Delta \vecwonekone[t] &=& - \lr \nabll{ \vecwonekone}[t] \nonumber \\
 \Delta \vecbonekone[t] &=& - \lr \nabll{\vecbonekone}[t] \nonumber \\
 \Delta \vecwtwoktwo [t]& =& - \lr \nabll{ \vecwtwoktwo} [t] \nonumber \\
 \Delta \vecbtwoktwo [t]& =& - \lr \nabll{ \vecbtwoktwo} [t]
\end{eqnarray}

\begin{figure}
\centering\includegraphics[scale=1.02]{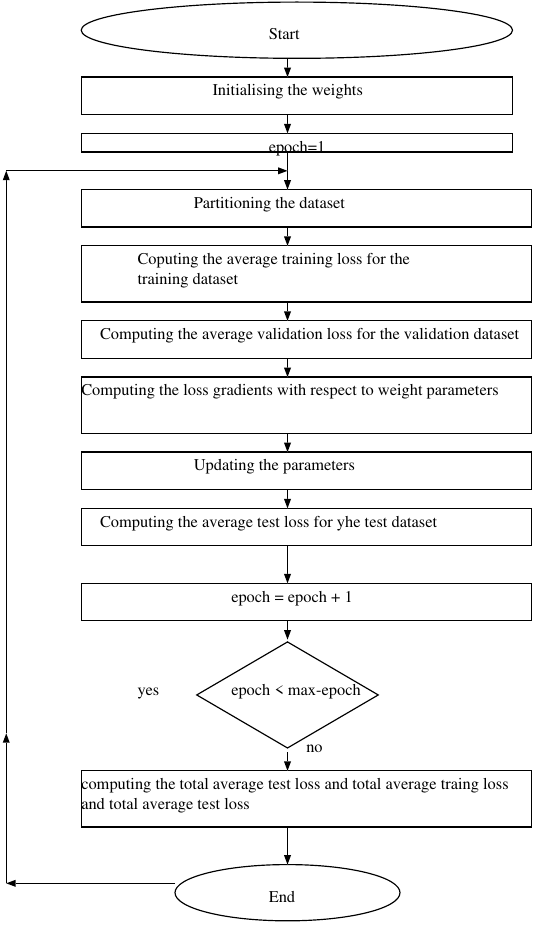}
\caption{The applied flowchart.}
\label{fig:flowchart}
\end{figure}

\begin{figure}
\centering\includegraphics[width=\textwidth, height= 12cm]{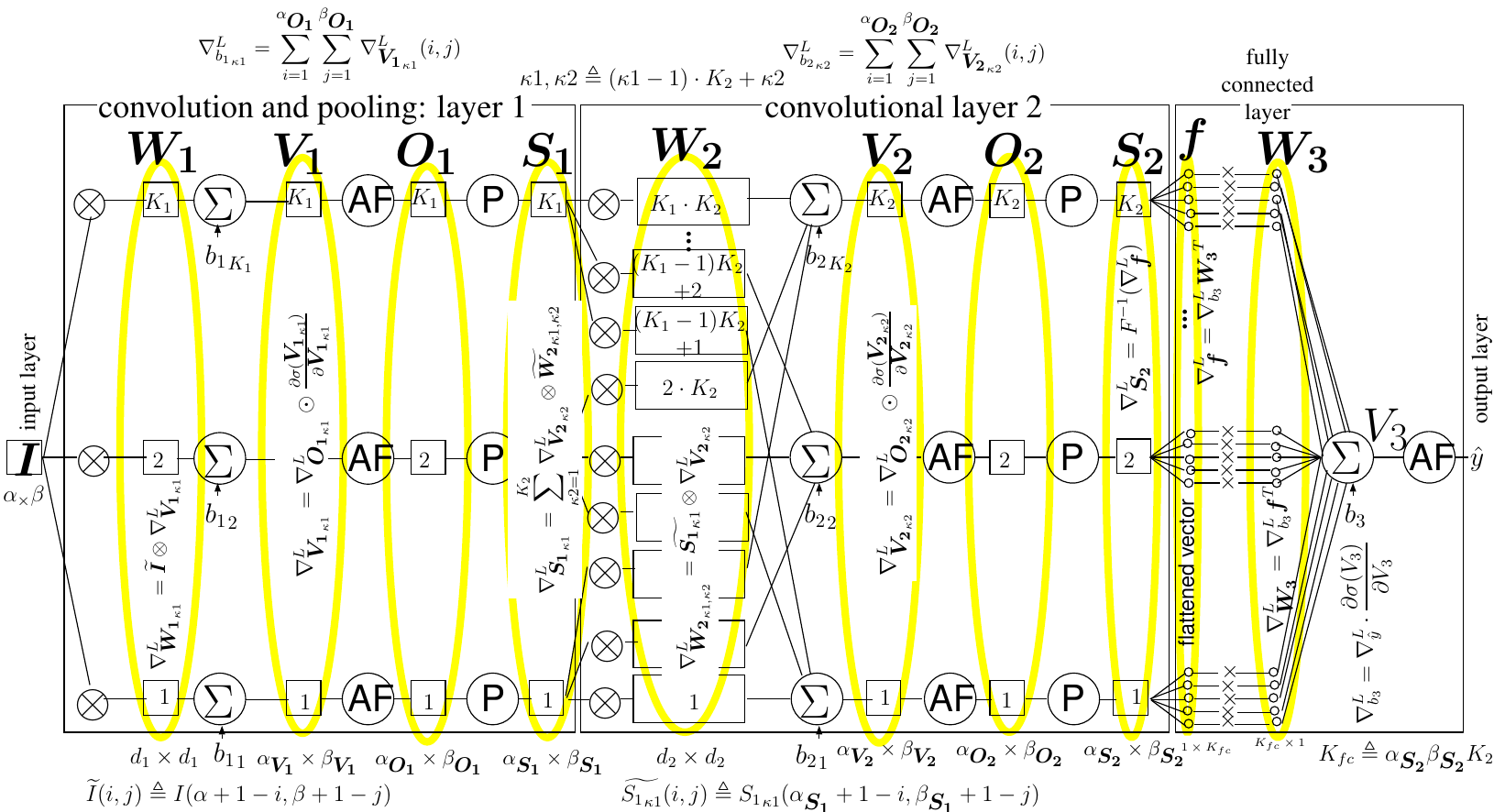}
\caption{The full two layer CV-CNN forward and backward architecture. }
\label{fig:cnnall}
\end{figure}

\fig{cnnall} illustrates the two layer forward network operations' details and backward propagation with all the vectors and matrices dimensions and used equations.

\section{Validation Method}
\fig{flowchart} displays the deployed algorithm, as the figure shows, in each epoch the dataset is partitioned into training, validation and test datasets. The network is trained for each training sample from the training dataset, then the parameters are updated and training loss for that sample is computed. At the end of the epoch, when the
network is trained for all the training samples, we compute the average of training loss for the whole training dataset. Moreover, at the end of each epoch we compute the test loss for the test dataset of that epoch and calculate the average validation loss which is the average of loss over the validation dataset.
As the\fig{flowchart} shows, after the training is run for as many times as the number set for ''max-epoch'', we can compute the total training, validation and test loss.

\subsection{Cross Validation}
The ''cross validation'' method is a model validation technique used to evaluate the accuracy of a model. We split the dataset into three parts of the training data, the validation data and the test data. The training data is used to train the network and we utilise the validation data to monitor the trained model's performance while in training process or to determine the termination of the CNN training iterations if we reach the desired accuracy. The accuracy of the estimation of the system output via the trained CNN network, to unseen data is evaluated using the test data. We partitioned the dataset sample pairs into ${\mym}_{tr}$  for training, ${\mym}_{val}$ for validation and ${\mym}_{tst}$ for test datasets. The CNN is trained using the training dataset. At each iteration CNN updates its weights.
${\myL}_{tr}$ the average training loss which is computed as\eq{trainl}. ${\vecy}_{tr}^{(m)}$ is the predicted output of the $m$-th input sample and $\hat{\ymm}$ is the $m$-th desired output.

\begin{eqnarray}
\label{eq:trainl}
{\myL}_{tr} &  = & \frac{1}{{\mym}_{tr}} \sum_{m=1}^{{\mym}_{tr}} | {y}_{tr}^{(m)} - \hat{\ymm}| ^{2}
\end{eqnarray}
\\
The validation loss is computed as in\eq{vall}, ${\myL}_{val}$ is the average validation loss. ${y}_{val}^{(m)}$ is the predicted output of the $m$-th input sample and $\hat{\ymm}$ is the $m$-th desired output.
\begin{eqnarray}
\label{eq:vall}
{\myL}_{val} &  = & \frac{1}{{\mym}_{val}} \sum_{m=1}^{{\mym}_{val}} | {y}_{val}^{(m)} - \hat{\ymm}| ^{2}
\end{eqnarray}
\\

\begin{eqnarray}
{\myL}_{tst} &=&\frac{1}{{\mym}_{tst}} \sum_{m=1}^{{\mym}_{tst}}  |{y}_{tst}^{(m)} - \hat{\ymm}|^{2}
\end{eqnarray}
\\
After the $\mym$ LOO iterations, the LOO loss is the average of test loss for all test sample in the loop.

\section{ReLU Activation Function's Advantages}

The activation function contributes and effects the gradient decent algorithm and optimising the network. Activation functions can be divided into two main categories of non-saturated and saturated \cite{activechar}. The non-saturated activation functions such as the ReLU family are used more in the literature than saturated activation functions such as Sigmoid, Tanh, etc \cite{activepanj} \cite {activeshish}. In deep neural network, while using the non-saturated activation function, there is no vanishing gradient problem and the learning process can be implemented rapidly. 

Computation of BP in neural networks with ReLU function is cheaper than sigmoid and hyperbolic tangent activation functions because there are no need for computing the exponential functions \cite{activeyek} \cite{activese} \cite{activedo}. In addition, the neural networks with ReLU activation functions converge much faster than those with saturating activation functions in terms of training time with gradient descent. Moreover, the ReLU function allows a network to easily obtain sparse representation. More specifically, when the output is $0$ for the negative value input, ReLU provides the sparsity in the activation of neuron units and improves the efficiency of data learning. However, when the input is zero or positive value, the features of the data can be retained largely.

The derivatives of ReLU function keep as the constant $1$, which can avoid trapping into the local optimisation and resolve the vanishing gradient effect occurred in Sigmoid and Tanh activation functions. Furthermore, deep neural networks with ReLU activation functions can reach their best performance with large labelled datasets. 

The invention of ReLU is one of the key factors leading to the recent revival of CNNs, however, the main drawback of ReLU is its zero derivative for negative inputs, which blocks the loss gradient from the layer before so may prevent the network from reactivating dead neurons \cite{recactive}. In order to overcome this problem, leaky rectified linear unit (LReLU) assigns a non-zero slope to the negative part of ReLU \cite{rect8}. Unlike ReLU, it allows a small portion of the back-propagated signal to pass to the layer before. By using a small value, the network can still output sparse activations and preserve its ability to reactivate the dead units.

\begin{eqnarray}
\label{eq:lrelu}
\forall z \in \mathbb{R}: \mathrm{LReLU}(z) =\left \{\begin{array}{lcl}
\alpha z & \mbox{for} & z < 0 \\
z & \mbox{for} & z \ge 0 
\end{array} \right.
\end{eqnarray}
\\
\eq{lrelu} solves the problem of deactivated neurons by assigning a non-zero fixed slope value ($\alpha$) to the negative part. However, LReLU is sensitive to the slope value \cite{recactive}\cite{rect8}. In order to avoid specifying the slope value, Parametric Rectified Linear Unit (PReLU) \cite{rect9}\cite{recactive} adaptively learns its value as in:

\begin{eqnarray}
\label{eq:prelu}
\forall {z}_{i} \in \mathbb{R}: \mathrm{PReLU}({z}_{i}) =\left \{\begin{array}{lcl}
\alphl{i} z_{i} & \mbox{for} & {z}_{i} < 0 \\
{z}_{i} & \mbox{for} & {z}_{i} \ge 0 
\end{array} \right.
\end{eqnarray}
\\
In order to avoid setting the slope value to a fix value, PReLU sets the slop value to be trainable similar to the network's weights and bias parameters. The $\alphl{i}$ indicates that PReLU's output varies on different inputs. Researchers have shown that learning the slope parameter gives better performance than manually setting it to a constant value \cite{rect9}. Moreover, Exponential Linear Unit (ELU) \cite{rect10} uses the exponential function for negative value inputs, ELU, not only speeds up training but also improves the network's performance\cite{recactive}.

\begin{eqnarray}
\label{eq:elu}
\forall z \in \mathbb{R}: \qquad \mathrm{ELU}(z) =\left \{\begin{array}{lcl}
\alpha({e}^{z} -1)  z & \mbox{for} & z < 0 \\
z & \mbox{for} & z \ge 0 
\end{array} \right.
\end{eqnarray}

\subsection{CV ReLU Activation Functions} 

The most commonly used activation function in CNN is ReLU (rectified linear unit).

\begin{eqnarray}
\forall z \in \mathbb{R}: {\relu}_{\Re} (z) =\left \{\begin{array}{lcl}
0 & \mbox{for} & z < 0 \\
z & \mbox{for} & z \ge 0 
\end{array} \right.
\end{eqnarray}
\\
${\relu}_{\Re}$ is the real form of $\relu$. We construct the CV ReLU ( $\mathbb{C}$ ReLU ) similar to the RV form which applies separate ReLUs on both real and imaginary parts of neurons as in:
\begin{eqnarray}
\label{eq:complexrelu}
\forall z \in \mathbb{C}:\mathbb{C} \relu(z) = \relu (\Re (z)) + \jmath \relu (\Im (z)),
\end{eqnarray}
\\
where $z \in \mathbb{C}$. However, \cite{oncomplex} explains another form of CV ReLU (zReLU), which satisfies the Couchy-Riemann equations everywhere except for the set of points $ ({\Re (z) > 0 , \Im (z) = 0}) \cup  ({\Re (z) = 0 , \Im (z) > 0})$. 

\begin{eqnarray}
\forall z \in \mathbb{C}:z \relu(z) =\left \{ \begin{array}{lcl}
z & \mbox{for} & arg(z) \in [0, \frac{\pi}{2}] \\
0 & \mbox{for} & \mathrm{otherwise}
\end{array} \right.
\end{eqnarray}
\\
\cite{deepcvNW} have applied $\mathbb{C} \relu$ and $z \relu(z)$ in their research.

\subsection{Proposed CV Activation Function}
In this research we use the $\mathbb{C} \relu$ as in\eeq{complexrelu}. ${\mathbb{C} \relu}$ satisfies the Couchy-Riemann equations when the real and imaginary parts are at the same time either positive or negative, The derivative of $\mathbb{C} \relu$ with respect to $z$ is a CV function which is computed as:
\begin{eqnarray}
\frac {d\mathbb{C} \relu(z)}{d z} =\left \{ \begin{array}{lcl}
1 & \mbox{for} & (\Re(z) > 0 , \Im(z) >0 ) \mathrm{  or  }  (\Re(z) < 0 , \Im(z) <0 ) \\ 
0 & \mbox{for} & \mathrm{otherwise}
\end{array} \right.
\end{eqnarray}

\section{CV-CNN CV Datasets Experiments}

In this section we display the training and test accuracy results of implementing our fully CV-CNN binary classification on two CV radar images datasets: complex-$1$ and complex-$2$. 

For each experiment there are two tables and two graphs attached. The first table demonstrates the architecture of the utilised CV-CNN and the second table, displays the parameter settings. The first graph displays the test and training accuracy that is achieved in the CV-CNN experiment and the second graph is the baseline accuracy graph from the corresponding RV-CNN.

The architecture table shows every layer's name, output dimensions and number of parameters in that layer and total number of trainable parameters of the selected architecture for the corresponding experiment.

The parameter setting table, displays the utilised kernel's(feature maps) dimensions, number of filters in each convolutional layer, learning rate value, pooling window dimensions, batch size and the number of epochs.

Furthermore, we have used the CReLU activation function after each convolutional layer and a CV Sigmoid function after fully connected layer for all the experiments of this chapter.     

\subsection{\texorpdfstring{Complex-$1$} Dataset Experiments}

This section illustrates two experiments and the results of training the complex-$1$ data set with the input sample dimension of $800 \times 540$ for each experiment. There are $945$ sample radar images for the both wave and swipe hand gestures all together, $80 \%$ of sample images are used as the training dataset and $20 \%$ are used as test dataset. 

\begin{table}
\centering\includegraphics[scale=0.9]{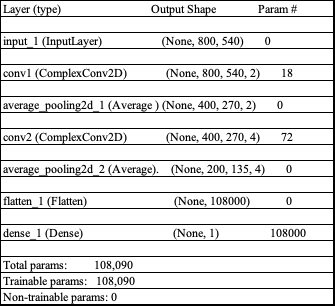}
\caption{Complex-$1$, First experiment, CV-CNN architecture.  }
\label{tbl:cnnforone}
\end{table}

\begin{table}
\centering\includegraphics[scale=0.9]{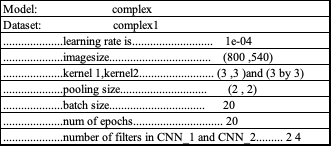}
\caption{Complex-$1$,First experiment, CV-CNN parameter settings. }
\label{tbl:cnnfortwo}
\end{table}

\begin{figure}
\centering\includegraphics[scale=0.76]{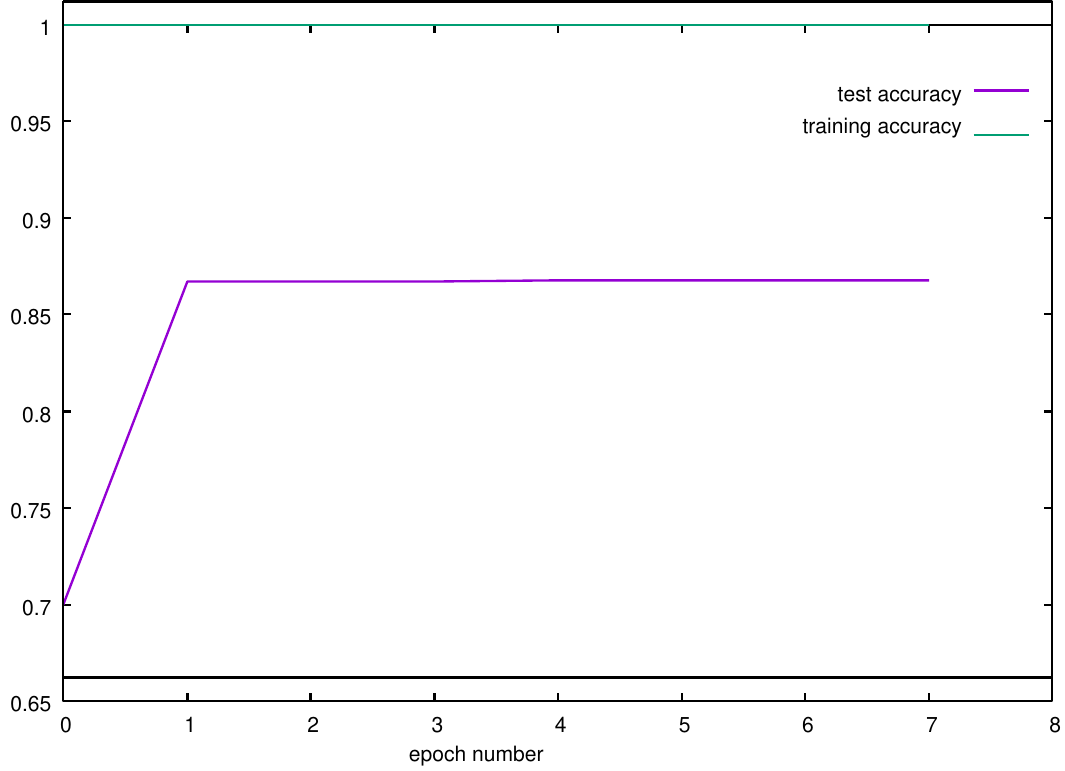}
\caption{Complex-$1$,First experiment, CV-CNN accuracy (setting:\figg{cnnforone},\figg{cnnfortwo}).  }
\label{fig:cnnforthree}
\end{figure}
\subsection{First Pooling Layer \texorpdfstring{($\vecs_{1}$)}{(s1)}}
\subsubsection{\texorpdfstring{Complex-$1$}{, First Experiment}}

Tables \figg{cnnforone} and\figg{cnnfortwo} illustrate that we used the complex-$1$ dataset in batches of $20$ with two feature maps in first convolutional layer and $4$ feature maps at the second convolutional layer. The feature map dimensions are $3 \times 3$ and the average pooling window dimensions are $2\times2$.\fig{cnnforthree} shows that even with only a couple of epochs we achieve the excellent result of $100 \%$ test accuracy and $83 \%$ of training accuracy.     

\begin{table}
\centering\includegraphics[scale=0.9]{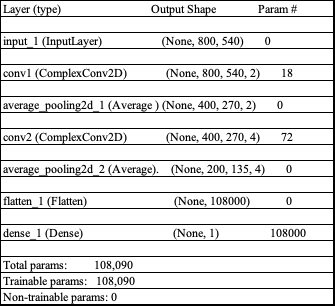}
\caption{Complex-$1$, Second experiment, CV-CNN architecture. }
\label{tbl:cnnforfour}
\end{table}

\begin{table}
\centering\includegraphics[scale=0.9]{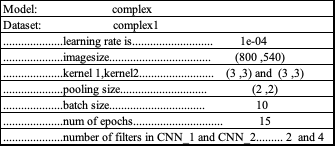}
\caption{Complex-$1$, Second Experiment, CV-CNN parameter settings. }
\label{tbl:cnnforfive}
\end{table}

\begin{figure}
\centering\includegraphics[scale=0.7]{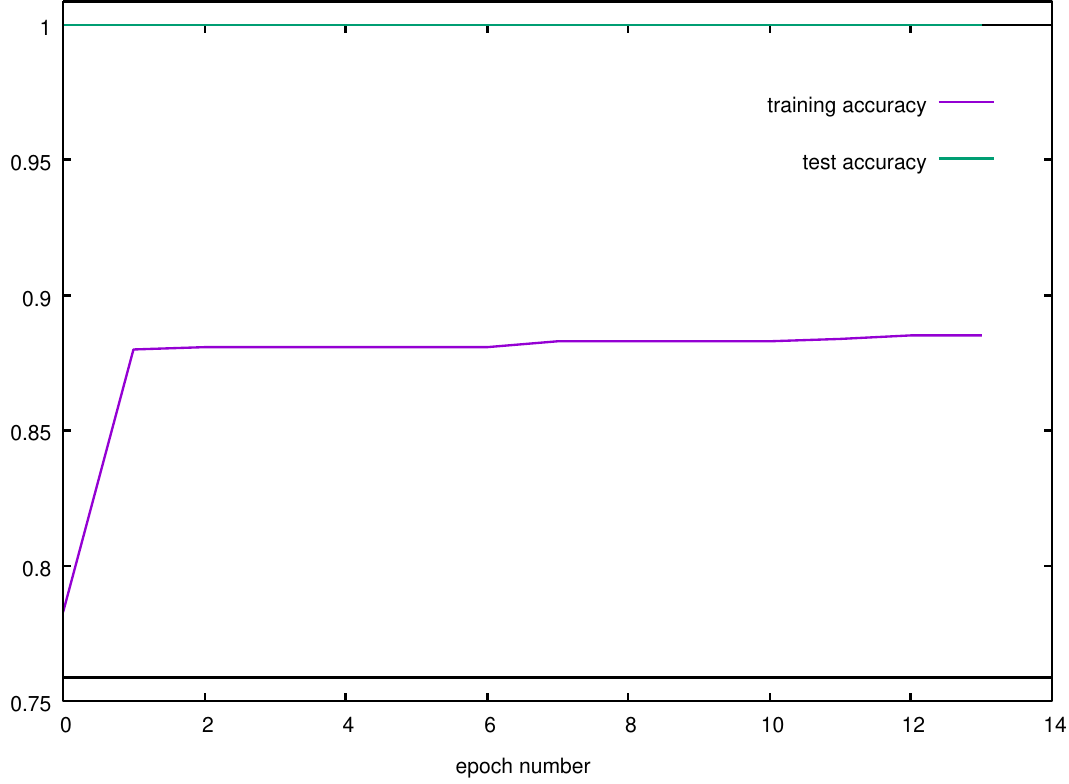}
\caption{Complex-$1$, Second experiment, CV-CNN accuracy (setting:\figg{cnnforfour},\figg{cnnforfive}).  }
\label{fig:cnnforsix}
\end{figure}

\subsubsection{\texorpdfstring{Complex-$1$}{, Second Experiment}}

Tables \figg{cnnforfour} and\figg{cnnforfive} illustrate that we used the complex-$1$ dataset in batches of $10$ with $2$ feature maps in first convolutional layer and $4$ feature maps at the second convolutional layer. The feature map dimensions are $3 \times 3$ and the average pooling window dimensions are $2\times2$.\fig{cnnforsix} shows that even with only a couple of epochs we achieve the
 remarkable result of $100 \%$ test accuracy and $85 \%$ of training accuracy.

\subsubsection{\texorpdfstring{Complex-$1$} {Experiments Conclusion}}

As the tables\figg{cnnforone} and\figg{cnnforfour} display the number of trainable parameters are the same for both experiments with complex-$1$ dataset ($108$k parameters), as they use the same network architecture and the only difference is the batch size, the smaller batch size demonstrates a slight improvement the training accuracy. However, the smaller batch size results in shorter batch training time.  

\begin{figure}
\centering\includegraphics[scale=0.8]{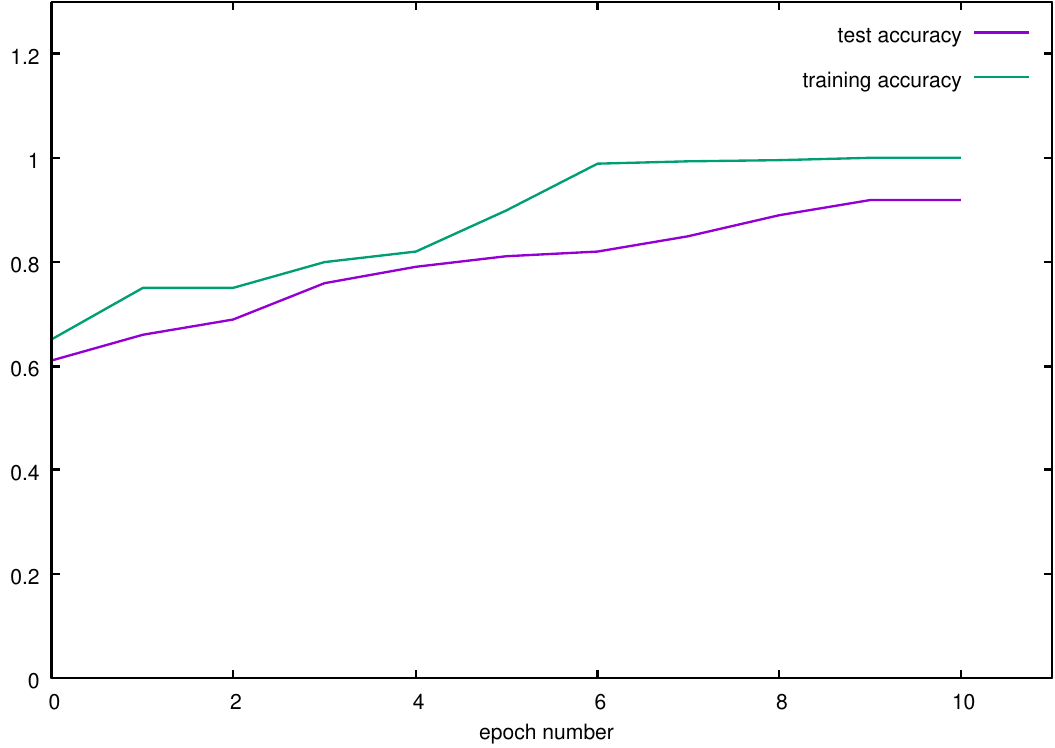}
\caption{Benchmark, Complex-$1$, RV-CNN accuracy. }
\label{fig:cvcnnbmone}
\end{figure}

Both first and second experiments, achieved the perfect result of $100\%$ test accuracy and over $83\%$ training accuracy. In comparison, the equivalent RV-CNN as in \fig{cvcnnbmone}, demonstrates similar accuracy for test and training over a higher number of epoch.

We can conclude, the RV-CNN, converge slower than the corresponding CV-CNN. Furthermore, in the case of CV-CNN, the number of parameters is $108090$ however, in RV-CNN the number of parameters are double, as it utilises two parallel CNNs, this increases the complexity of the network and training time. We should note that the RV-CNN does not take the correlation between the real and imaginary part of the data into account.

\subsection{\texorpdfstring{Complex-$2$}{ Dataset Experiments}}
This section illustrates experiments and the results of training the complex-$2$ data set which has the input sample dimension of $800 \times 231$ for each experiment. There are $1488$ sample radar images for the both pinch and click hand gestures all together that are taken from $6$ different people, $80 \%$ of sample images are used as the training dataset and $20$ percent are used as test dataset. 

\begin{table}
\centering\includegraphics[scale=0.9]{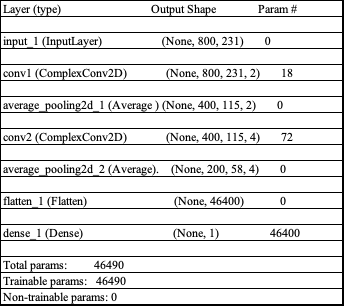}
\caption{Complex-$2$, First experiment, CV-CNN architecture.  }
\label{tbl:cnnforsev}
\end{table}

\begin{table}
\centering\includegraphics[scale=0.9]{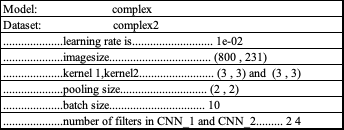}
\caption{Complex-$2$, First experiment, CV-CNN parameter settings. }
\label{tbl:cnnforeight}
\end{table}

\begin{figure}
\centering\includegraphics[scale=0.7]{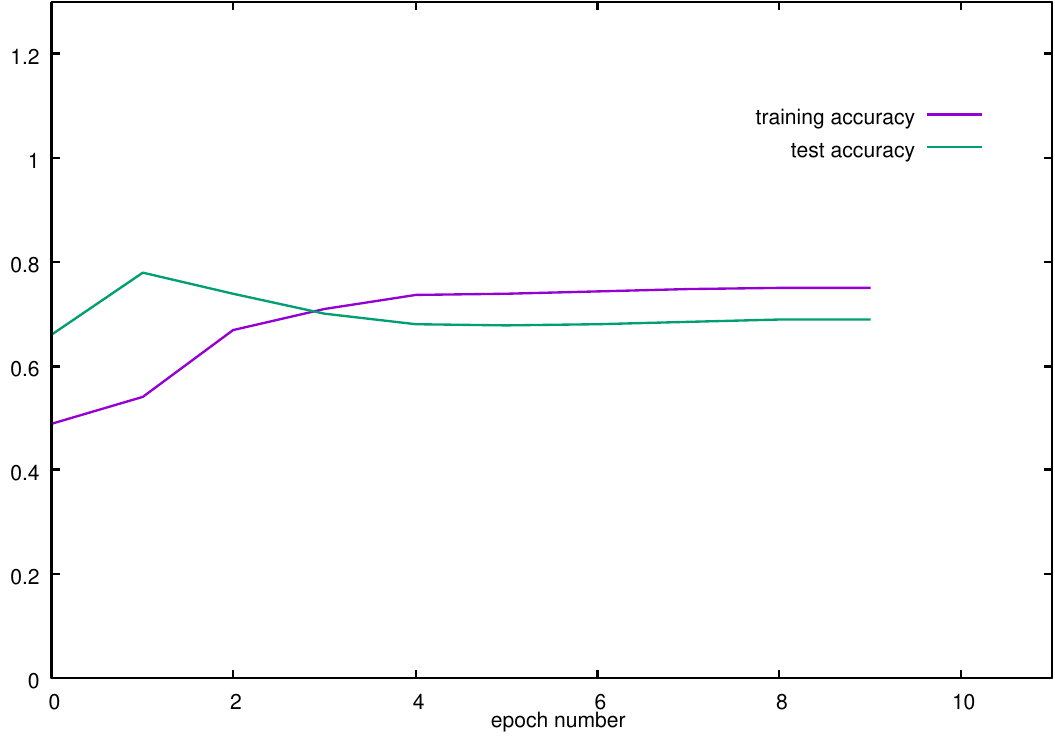}
\caption{Complex-$2$, First experiment, CV-CNN accuracy (setting:\figg{cnnforsev},\figg{cnnforeight}).  }
\label{fig:cnnfornin}
\end{figure}

The tables \figg{cnnforeight} and\figg{cnnforsev} display the CV-CNN architecture for our first experiment with complex-$2$ dataset. We trained the network in batches of $10$ with $2$ feature maps in first convolutional layer and $4$ feature maps at the second convolutional layer. The feature map dimensions are $3 \times 3$ and the average pooling window dimensions are $2\times2$.

\begin{figure}
\centering\includegraphics[scale=0.7]{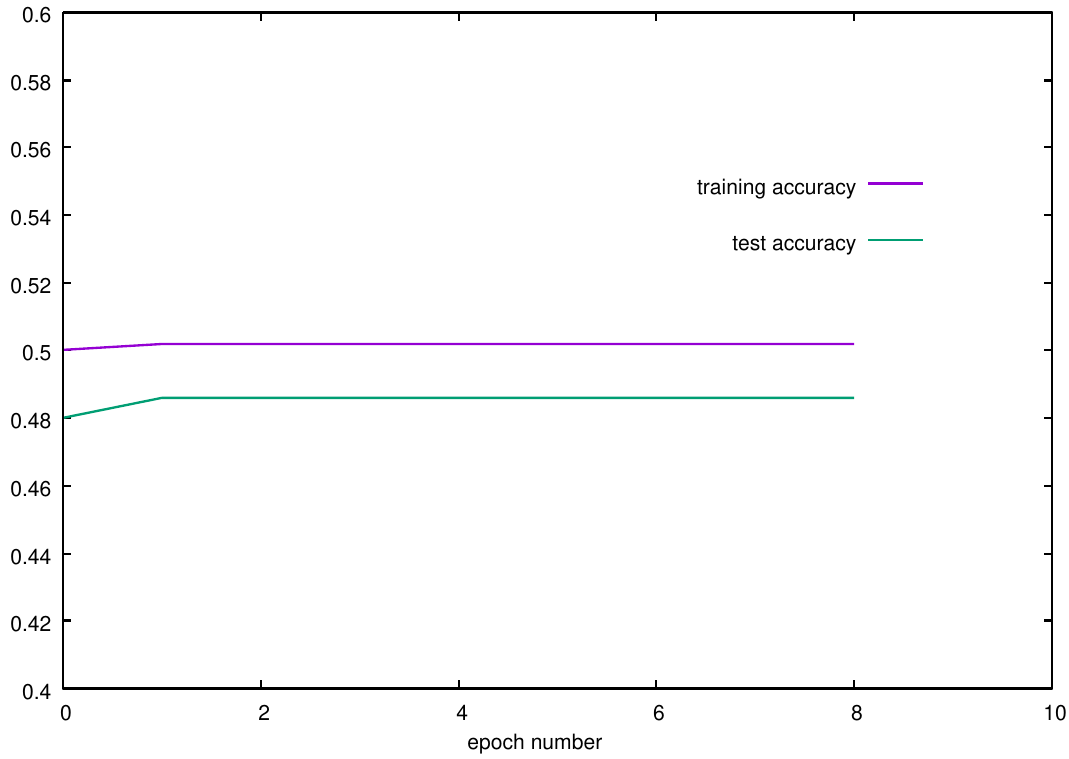}
\caption{Complex-$2$, Second experiment, CV-CNN accuracy (setting:\figg{cnnforten},\figg{cnnforele}).  }
\label{fig:cnnfortwl}
\end{figure}

\begin{figure}
\centering\includegraphics[scale=0.76]{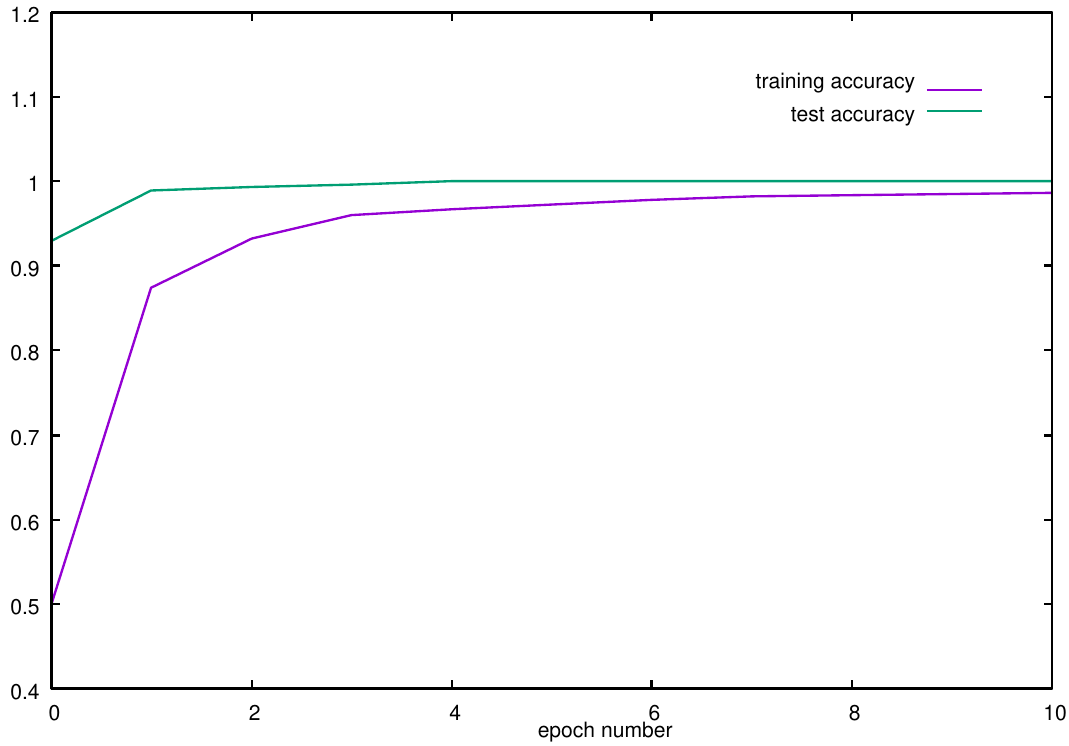}
\caption{Complex-$2$, Third experiment, CV-CNN accuracy (setting:\figg{cnnfortten},\figg{cnnforften}).  }
\label{fig:cnnforfften}
\end{figure}

\fig{cnnfornin},\fig{cnnfortwl} and\fig{cnnforfften} illustrate the number of trainable parameters are the same for all three experiments with complex-$2$ dataset ($46$k parameters), as they use the same network architecture and the only difference is the batch size. The number of trainable parameters is a lot smaller than the experiments with complex-$1$ with $108$k parameters, it is due to smaller sample dimensions in Complex-$2$ dataset, which makes the process of computing the training faster.

\begin{table}
\centering\includegraphics[scale=0.9]{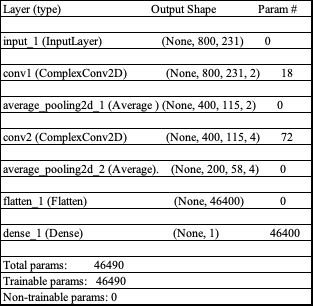}
\caption{Complex-$2$, Second experiment, CV-CNN architecture.  }
\label{tbl:cnnforten}
\end{table}

\begin{table}
\centering\includegraphics[scale=0.9]{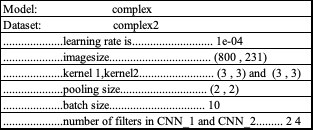}
\caption{Complex-$2$, Second experiment CV-CNN parameter settings. }
\label{tbl:cnnforele}
\end{table}

\begin{table}
\centering\includegraphics[scale=0.9]{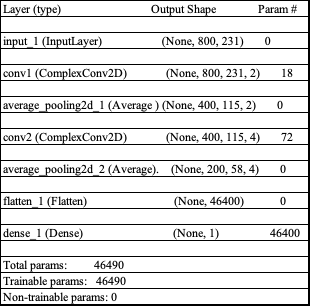}
\caption{Complex-$2$, Third experiment, CV-CNN architecture.  }
\label{tbl:cnnfortten}
\end{table}

\begin{table}
\centering\includegraphics[scale=0.9]{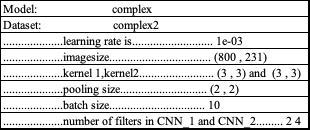}
\caption{Complex-$2$, Third experiment, CV-CNN parameter settings. }
\label{tbl:cnnforften}
\end{table}

\subsubsection{\texorpdfstring{Complex-$2$}{ Experiments Conclusion}}
All three experiments with complex-$2$ dataset have the same architecture and set to the same hyper parameters, the only difference the value of the learning rate, the experiments investigating the effect of bigger and smaller learning rate on the accuracy of the network with the same architecture. As\fig{cnnfornin},\fig{cnnfortwl} and\fig{cnnforfften} display learning rate range of smaller value as $0.0001$ and average value of $0.001$ and a bigger value of $0.01$, the performance of the $0.001$ is remarkable as $100 \%$ accurate classification result, as the very small learning rate can results in very slow or not converging and very big learning rate can cause missing the minimum of the loss function during the process of training the network.    

\begin{figure}
\centering\includegraphics[scale=0.78]{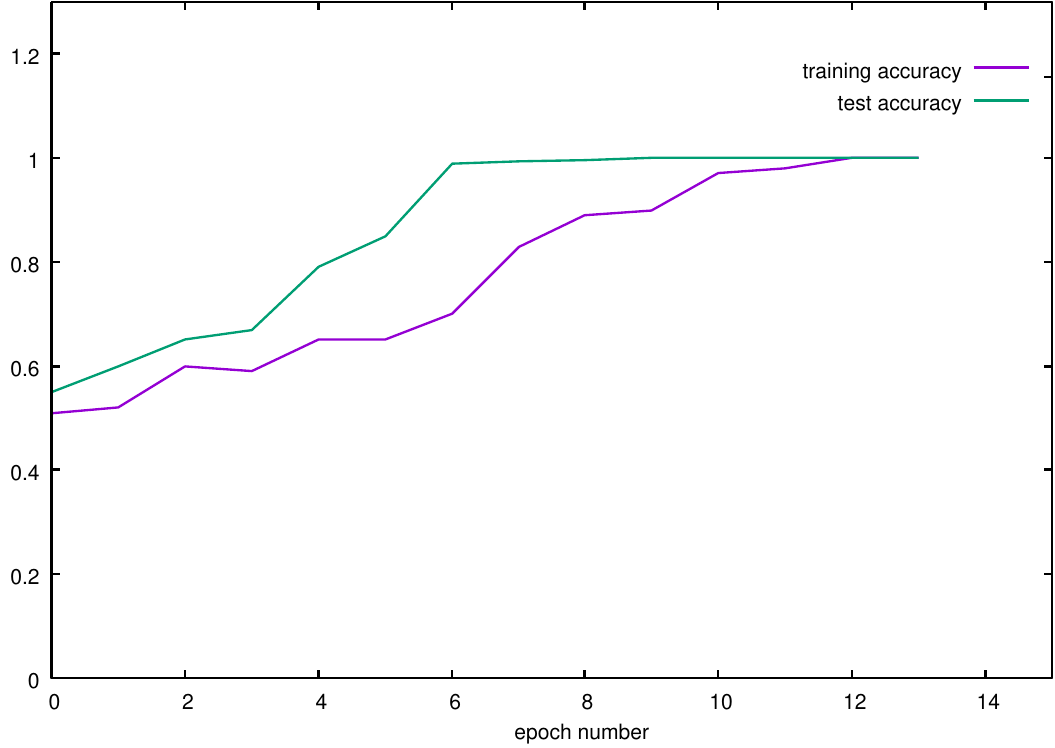}
\caption{Benchmark, Complex-$2$ dataset RV-CNN accuracy. }
\label{fig:cvcnnbmtwo}
\end{figure}

All three CV experiments have total $46490$ trainable parameters. In third experiment we achieved the perfect result of $100\%$ test and training accuracy after 6 epochs. In comparison, the equivalent RV-CNN as in \fig{cvcnnbmtwo}, demonstrates similar accuracy for test and training over 12 epochs.

We can conclude, the RV-CNN, converge slower than the corresponding CV-CNN. Furthermore, in the case of CV-CNN, the number of parameters is $46490$ however, in RV-CNN the number of parameters are double, as it utilises two parallel CNNs, this increases the complexity of the network and training time. We should note that the RV-CNN does not take the correlation between the real and imaginary part of the data into account.




\section{Conclusion}

We run many CV-CNN training experiments on both complex-$1$ and complex-$2$ datasets and explored who different hyper parameters can improve the accuracy or reduce the number of trainable parameters in order to reduce the memory and time required to run the experiment. The results can be summarised as bellow:

\begin{itemize}
\item We achieve the binary classification accuracy of $100\%$ for test and over $83\%$ for training on CV-CNN for complex-$1$.
\item We achieve the binary classification accuracy of $100\%$ for test and $100\%$ for training on CV-CNN for complex-$2$.
 \item Smaller batch size demonstrates a better accuracy result and reduces the time of computation on CV-CNN experiments.
\item Very small learning rate slows the learning down and big learning rate and misses the minimum point of the loss and therefore misses the best accuracy.
  
\end{itemize}
As conclusion, our fully CV-CNN displayed a very accurate and capable learning ability for CV datasets. The equivalent RV-CNN demonstrate lower test and training accuracy in addition to longer training epochs.    

\chapter{ CV-Forward Residual Network}
\label{cha:yfourres}

Residual networks (\cite{hefift} and \cite{hesix}) emerged as one of the most popular and effective strategies for training very deep CNNs. Residual networks facilitate the training of neep networks by providing shortcut paths for easy gradient flow to lower network layers, in order to reduce the effects of vanishing gradients. \cite{hesix} Demonstrates that learning explicit residuals of layers helps in avoiding the vanishing gradient problem
and provides the network with an easier optimisation problem. Batch normalisation (\cite{iof}) demonstrates that standardising the activations of intermediate layers in a network across a mini-batch acts as a powerful regulariser as well as providing faster training and better convergence properties. Furthermore, such techniques that standardised layer outputs become critical in deep architectures due to the vanishing and exploding gradient problems.

 This chapter explains the architecture of our proposed CV-forward residual network including detailed specifications and function of each layer. We implement the architecture of the suggested CV-forward residual in this chapter in Python language. Our contribution is to develop the residual network for CV input data of hand gesture radar images, every layer's function is CV simulated, including convolutional, pooling, weight initialisation, batch normalisation and activation functions. However the back propagation and derivatives of the functions are all in RV domain, thus, we call the network CV-forward not fully-CV.

This chapter first defines each layer of the proposed CV-forward residual network's building block. Then denotes the mathematical operation of each block, then. At the end of this chapter, we display the training and test results of implementing our proposed model on two CV radar images datasets (complex-$1$ and complex-$2$). We compare our proposed model's results with the equivalent RV-residual model with same architecture, settings and CV datasets as a baseline.

\section{CV-forward Residual Building Blocks }

At present, most building blocks, techniques and architectures for deep learning are based on the RV operations and representations. Despite their
attractive properties and potential for opening up entirely new neural architectures, CV deep neural networks have been marginalised due to the absence
of the building blocks required to design such models. \cite{complexnn} Implements CV convolutions and presents an algorithms for CV batch-normalisation and CV weight initialisation strategies for CV neural nets. \cite{complexnn} Results demonstrates that such CV models are competitive with their RV counterparts. They test deep CV residual network models on several computer vision
tasks such as music transcription using the MusicNet dataset and on Speech Spectrum
Prediction, using the TIMIT dataset, however all the datasets that they used are RV datasets.

\subsection{Presentation of the CV Numbers}

\begin{figure}
\centering\includegraphics[scale=0.9]{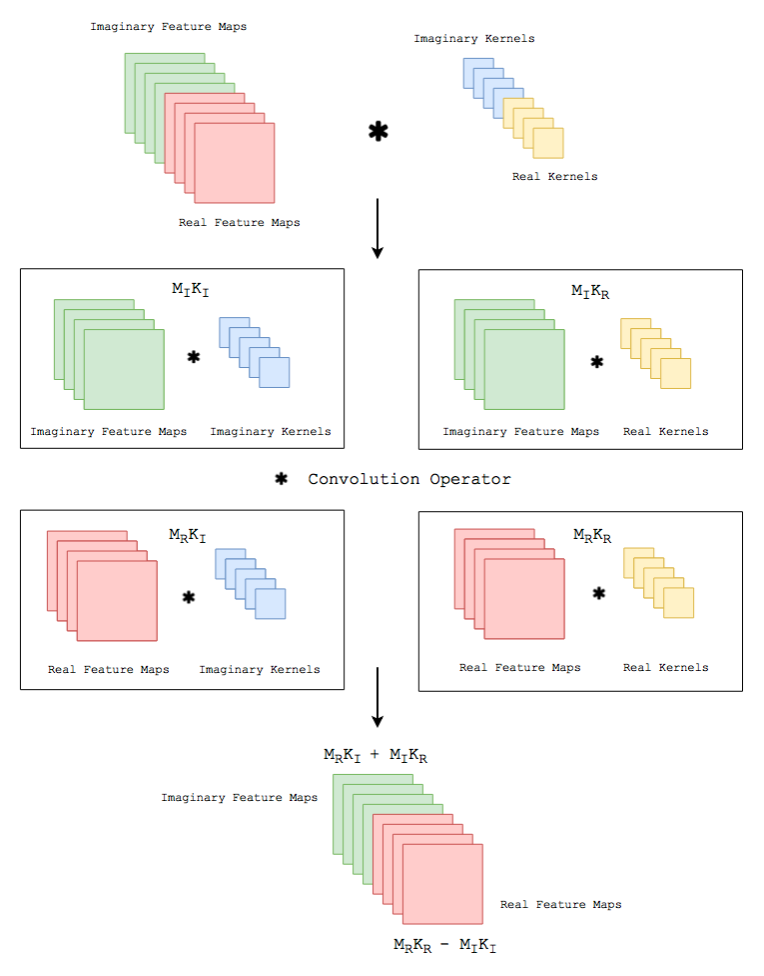}
\caption{A simulated CV convolutional arithmetic using RV arithmetic }
\label{fig:restwo}
\end{figure}

\cite{complexnn} Outlines the way in which CV numbers are represented in the framework. A
CV number $z = a + i b$ has a real component $a$ and an imaginary component $b$. \cite{complexnn} Represents
the real part $a$ and the imaginary part $b$ of a CV number as logically distinct RV entities
and simulate CV arithmetic using RV arithmetic internally as in \fig{restwo}. Considering a typical RV
$2$D convolution layer that has $N$ feature maps such that $N$ is an even number, in order to represent
 CV numbers, \cite{complexnn} method allocates the first $N/2$ feature maps to represent the real components
and the remaining $N/2$ to represent the imaginary ones. Thus, for a four dimensional weight tensor
$\mbox{\boldmath $W$}$ that links $N_{in}$ input feature maps to $N_{out}$ output feature maps and whose kernel size is $m \times m$
we would have a weight tensor of size (${N_{out}}\times {N_{in}} \times\times m$) $/2$ CV weights.

\subsection{Simulated CV Convolutional Operation}

\cite{complexnn} Presents a general formulation for the building components of CV deep neural networks and its application to the context of feed-forward convolutional networks in addition to a formulation of CV batch normalisation and CV weight initialisation utilising Residual network.
\begin{figure}
\centering\includegraphics[scale=0.8]{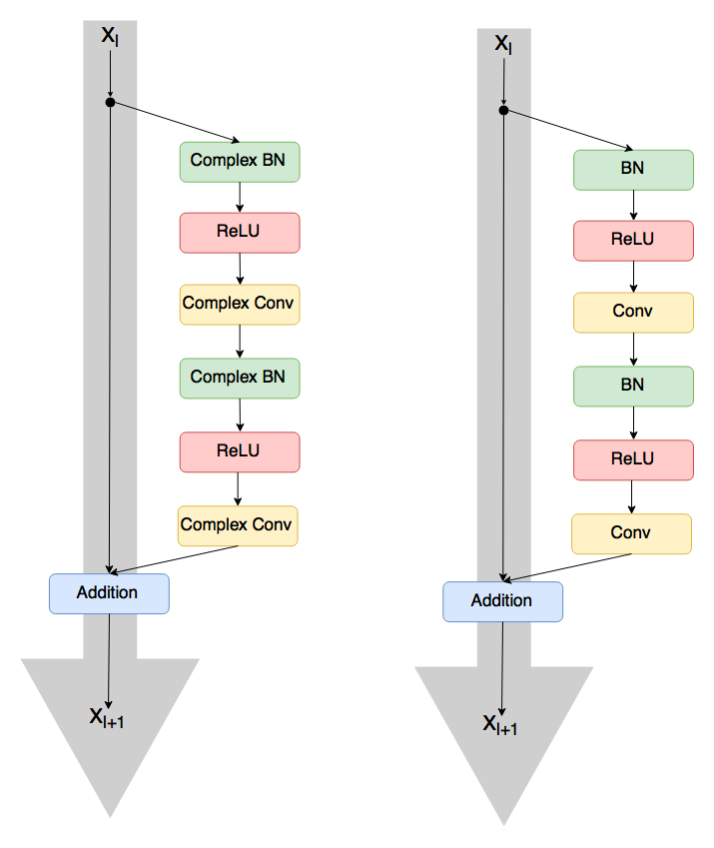}
\caption{A CV convolutional residual network (left) and an equivalent RV residual network (right)  }
\label{fig:resone}
\end{figure}

The CV convolution operation is implemented as a simulation of RV arithmetic, as illustrated in ~\fig{resone}, where $M_I$ , $M_R$ refer to
imaginary and real feature maps and $K_I$ and $K_R$ refer to imaginary and real kernels. $M_I$ $K_I$ refers to result of
a RV convolutional operation between the imaginary kernels $K_I$ and the imaginary feature maps $M_I$. In order to perform the equivalent of a traditional RV  $2$D convolution in the CV domain,
\cite{complexnn} simulate the convolution of a CV filter matrix ${\mbox{\boldmath{W}}} = {\mbox{\boldmath{A}}} + \jmath \mbox{\boldmath{B}}$ by a CV vector ${\mbox{\boldmath{h}}} = {\mbox{\boldmath{x}}} + \jmath \mbox{\boldmath{y}}$. Where $\mbox{\boldmath{A}}$ and
${\mbox{\boldmath{B}}}$ are RV matrices and ${\mbox{\boldmath{x}}}$ and ${\mbox{\boldmath {y}}}$ are RV vectors since we are simulating CV arithmetic using RV entities. As the convolution operator is distributive, convolving the vector ${\mbox{\boldmath{h}}}$ by the filter
$\mbox{\boldmath{W}}$ results in \eq{resyek}.

\begin{eqnarray}
\label{eq:resyek}
{\mbox{\boldmath{W}}}. \mbox{\boldmath{h}} & = &( {\mbox{\boldmath{A}}}. \mbox{\boldmath{x}} - \mbox{\boldmath{B}}. \mbox{\boldmath{y}})+ \jmath ( \mbox{\boldmath{B}}.  \mbox{\boldmath {x}} + \mbox{\boldmath{A}} . \mbox{\boldmath{y}})   
\end{eqnarray}

If we use matrix notation to represent real and imaginary parts of the
convolution operation we have \eq{restdo}

\begin{eqnarray}
\label{eq:restdo}
\begin{bmatrix}
\cal{R}({\mbox{\boldmath{W}}} . {\mbox{\boldmath{h}}})\\
\cal{I}({\mbox{\boldmath{W}}} . {\mbox{\boldmath {h}}})
\end{bmatrix}
=
\begin{bmatrix}
 {\mbox{\boldmath{A}}} & -{\mbox{\boldmath {B}}}\\
 {\mbox{\boldmath{B}}} & {\mbox{\boldmath{A}}})
\end{bmatrix}
.
\begin{bmatrix}
 {\mbox{\boldmath {x}}}\\
 {\mbox{\boldmath {y}}}
\end{bmatrix}
\end{eqnarray}
\\

\subsection{Creating the CV dataset from a RV dataset}

\cite{complexnn} Utilises the deep convolutional residual network which is presented in \cite{hefift} and \cite{hesix},the residual network consists
of $3$ stages within which feature maps maintain the same shape. At the end of a stage, the feature
maps are downsampled by a factor of $2 $and the number of convolution filters are doubled. The sizes
of the convolution kernels are always set to $3 \times 3$. Within a stage, there are several residual blocks
which comprise $2$ convolution layers each. The contents of one such residual block in the real and
complex setting is illustrated in~ \fig{resone}.
In the CV setting, the majority of the architecture in \cite{complexnn} remains identical to the one presented
in \cite{hesix} with a few subtle differences. Since all datasets that \cite{complexnn} work with have RV inputs, they present a way to learn their imaginary components to let the rest of the network
operate in the complex plane. They learn the initial imaginary component of the RV input by performing
the operations present within a single RV residual block as in :\\

BN$-->$  ReLU $-->$ Conv$-->$  BN$-->$  ReLU$-->$  Conv\\
Where BN means batch normalisation, ReLU is the activation function layer and Conv is a convolutional layer.
Using this learning block yielded better emprical results than assuming that the input image has a
null imaginary part. The parameters of this CV residual block are trained by back propagating
errors from the task specific loss function. Secondly, \cite{complexnn} perform a Conv, BN and activation
operation on the obtained complex input before feeding it to the first residual block. They also perform
the same operation on the RV network input instead of Conv Maxpooling as in \cite{hesix} . Inside, residual blocks, they subtly alter the way in which the  projection is performed. The complex models are tested with the CReLU, zReLU and modReLU. A cross entropy loss for both real and complex models are used. A global
average pooling layer followed by a single fully connected layer with a softmax function is used to
classify the input as belonging to one of 10 classes in the CIFAR-10 and SVHN datasets and 100
classes for CIFAR-100.

\section{Proposed CV-forward Residual Network}

We utilise the CV residual blocks from \cite{complexnn}, the residual blocks are such as \fig{restwo} which consists of $2$ layers of convolutions and average pooling, \cite{complexnn} adopts a few of the residual blocks and cascades them after each other, however as our intention is to reduce the computational time and the number of parameters due to the large dimensions of our CV samples in our CV datasets, we just use one residual block with $2$ feature maps at the first convolutional layer and $4$ feature maps at the second convolutional layer. The activation function is CRelU at the convolutional layers and softmax after the flatten layer. We utilise and modify the Python code which uses the CV-CNN, CV batch normalisation and CV weigh initialisation \cite{complexnn}, however we make adjustments to the code to be able to load our CV dataset instead of utilising a RV dataset and learn the initial imaginary component of the CV input. However,\cite{complexnn}utilise the RV dataset and learn the initial imaginary part of input samples through the~ BN$-->$  ReLU $-->$ Conv$-->$  BN$-->$  ReLU$-->$  Conv ~ operation.

In addition, in order for the results to be comparable with our other experiments such as fully CV-CNN and CV-forward CNN we do not use any data augmentation or regularisation techniques in our residual CV-forward experiments (unlike the \cite{complexnn}).      
Thus, we summarise our proposed CV-forward Residual network setting as:
\begin{itemize}
\item Only $1$ CV-forward residual block is utilised.
\item $2$ CV convolutional filters in the first convolutional layer and $4$ filters in the second convolutional layer of the residual block is utilised.
\item Simulated CV-CNN is utilised. 
\item CV batch normalisation and CV weigh initialisation are implemented.
\item CV dataset is used.
\item Each input sample's real and imaginary components are separated then concatenated together, so the input dataset dimensions are doubled.
\item The activation function used is CReLU on the convolutional layers and softmax after the flatten layer.    
\item Back propagation derivatives are all in RV domain.
\end{itemize}

\section{CV-forward Residual CV Datasets Experiments}

In this section we display the training and test accuracy results of implementing our proposed CV-forward residual model binary classification on two CV radar images datasets: complex-$1$ and complex-$2$. 

For each experiment there are two tables and two graphs attached. The first table demonstrates the architecture of the utilised residual network and the second table, displays the parameter settings. The first graph displays the test and training accuracy that is achieved in the CV experiment and the second graph is the baseline accuracy graph from the corresponding RV network. The architecture table shows every layer's name, output dimensions and number of parameters in that layer and total number of trainable parameters of the selected architecture for the corresponding experiment.

The parameter settings table, displays the utilised kernel's(feature maps) dimensions, number of filters in each convolutional layer, learning rate value, pooling window dimensions, batch size and the number of epochs. Furthermore, we have used the CReLU activation function after each convolutional layer and a tanh function after fully connected layer for all the experiments of this chapter.

\subsection{\texorpdfstring{Complex-$ 1$} Dataset Experiments}
In our first experiment we run a CV-forward residual network we train our complex-$1$ dataset that consists of $945$ samples of wave and swipe hand gesture radar images, the dimensions of each sample is equal to $800\times540\times2$. Tables \figg{resforone} and \figg{resfortwo} display the parameter settings for this experiment, the number of trainable parameters for complex-$1$ dataset is $13832$k and the batch size is $20$.

\begin{figure}
\centering\includegraphics[scale=0.76]{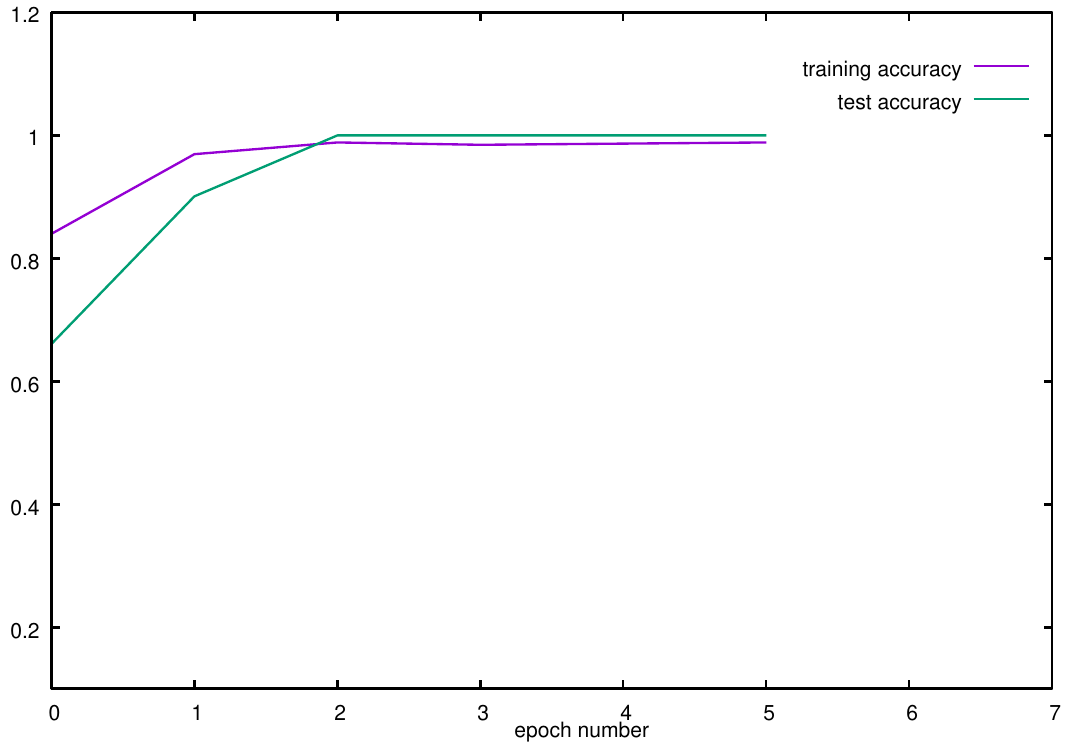}
\caption{Complex-$1$ dataset CV-forward Residual accuracy per epoch (setting:~\figg{resforone}, \figg{resfortwo})  }
\label{fig:resforthree}
\end{figure}

\begin{figure}
\centering\includegraphics[scale=0.78]{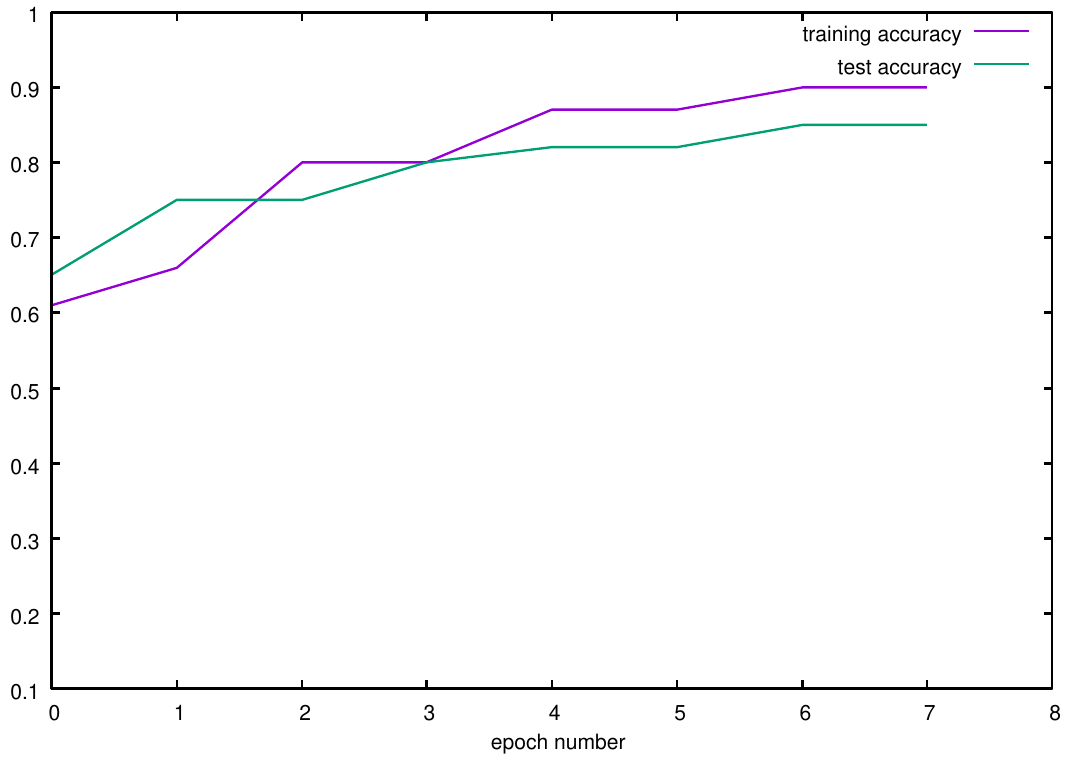}
\caption{Complex-$2$ dataset RV Residual }
\label{fig:rvrestwo}
\end{figure}

 We explored different combination of hyper parameters and as \fig{resforthree} shows, the result is very accurate in binary classification of the hand gestures on our complex-$1$ dataset. We achieve $100\%$ test and training accuracy after $3$epoch, which is outstanding result. However as \fig{rvrestwo} demonstrates, the equivalent RV residual network with similar parameter settings, converge slower (after $8$ epochs) and the test and training accuracy are both lower at $84\%$. Thus, for the case of hand gesture binary classification, the CV-forward residual model performs more accurately and converge much faster. 

\begin{table}
\centering\scalebox{0.65}{  
 \begin{tabular}{|l|l|l|l|}
 \hline
 \hline
 Layer(type) &Output shape & Param \# & Connected to \\
 \hline
 \hline
 input-1(InputLayer) & (None,800,540,2) & 0&  \\
 \hline
 conv1(ComplexConv2D) & (None,800,540,4) & 36 & input-1[0][0] \\
 \hline
 bn-conv1-2a(ComplexBatchNorm) & (None,800,540,4) & 20 &conv1[0][0] \\
 \hline
 activation-1(Activation) & (None,800,540,4) & 0 &  bn-conv1-2a[0][0]\\
 \hline
 bn20-branch-2a(ComplexBatchNorm) & (None,800,540,4) &20 & activation-1[0][0]\\
 \hline
 activation-2(Activation) & (None,800,540,4) & 0 &  bn20-branch-2a[0][0]\\
 \hline
 res20-branch2a(ComplexConv2D) & (None,800,540,4) & 72 & activation-2[0][0]\\
 \hline
 bn20-branch-2b(ComplexBatchNorm) & (None,800,540,4) &20 & res20-branch-2a[0][0]\\
 \hline
 activation-3(Activation) & (None,800,540,4) & 0 &  bn20-branch-2b[0][0]\\
 \hline
 res20-branch2b(ComplexConv2D) & (None,800,540,4) & 72 & activation-3[0][0]\\
 \hline
 add-1(Add) &(None,800,540,4) & 0 &  res20-branch2b , activation-1\\
 \hline
 bn30-branch-2a(ComplexBatchNorm) & (None,800,540,4) &20 & add-1[0][0]\\
 \hline
 activation-4(Activation) & (None,800,540,4) & 0 &  bn30-branch-2a[0][0]\\
 \hline
 res30-branch2a(ComplexConv2D) & (None,400,270,4) & 72 & activation-4[0][0]\\
 \hline
 bn30-branch-2b(ComplexBatchNorm) & (None,400,270,4) &20 &  res30-branch2a[0][0]\\
 \hline
 activation-5(Activation) & (None,400,270,4) & 0 &  bn30-branch-2b[0][0]\\
 \hline
 res30-branch1(ComplexConv2D) & (None,400,270,4) & 8 & add-1[0][0]\\
 \hline
 res30-branch2b(ComplexConv2D) & (None,400,270,4) & 72 & activation-5[0][0]\\
 \hline
 get-real-1(GetReal) & (None,400,270,2) & 0 & res30-branch1[0][0]\\
 \hline
 get-real-2(GetReal) & (None,400,270,2) & 0 & res30-branch2b[0][0]\\
 \hline
 get-imag-1(GetImag) &(None,400,270,2) & 0 & res30-branch1[0][0]\\
 \hline
 get-imag-2(GetImag) &(None,400,270,2) & 0 & res30-branch2b[0][0]\\
 \hline
 concatenate-1(Concatenate) & (None,400,270,4) & 0 & get-real-1 ,  get-real-2\\
 \hline
 concatenate-2(Concatenate) & (None,400,270,4) & 0 & get-imag-1 ,  get-imag-2\\
 \hline
 concatenate-3(Concatenate) & (None,400,270,8) & 0 &  concatenate-1 , concatenate-2  \\
 \hline
 bn40-branch-2a(ComplexBatchNorm) & (None,400,270,8) &40 & concatenate-3[0][0]\\
 \hline
 activation-6(Activation) & (None,400,270,8) & 0 &  bn40-branch-2a[0][0]\\
 \hline
 res40-branch2a(ComplexConv2D) & (None,200,135,8) & 288 & activation-4[0][0]\\
 \hline
 bn40-branch-2b(ComplexBatchNorm) & (None,200,135,8) &40 &  res40-branch2a[0][0]\\
 \hline
 activation-7(Activation) & (None,200,135,8) & 0 &  bn40-branch-2b[0][0]\\
 \hline
 res40-branch1(ComplexConv2D) & (None,200,135,8) & 32 & concatenate-3[0][0]\\
 \hline
 res40-branch2a(ComplexConv2D) & (None,200,135,8) & 288 & activation-7[0][0]\\
 get-real-3(GetReal) & (None,200,135,4) & 0 & res40-branch1[0][0]\\
 \hline
 get-real-4(GetReal) & (None,200,135,4) & 0 & res40-branch2b[0][0]\\
 \hline
 get-imag-3(GetImag) &(None,200,135,4) & 0 & res40-branch1[0][0]\\
 \hline
 get-imag-4(GetImag) &(None,200,135,4) & 0 & res40-branch2b[0][0]\\
 \hline
 concatenate-4(Concatenate) & (None,200,135,8) & 0 & get-real-3 ,  get-real-4\\
 \hline
 concatenate-5(Concatenate) & (None,200,135,8) & 0 & get-imag-3 ,  get-imag-4\\
 \hline
 concatenate-6(Concatenate) & (None,200,135,16) & 0 &  concatenate-4 , concatenate-5  \\
 \hline
 average-pooling2d-1(AveragePool) & (None,25,16,16) & 0 & concatenate6[0][0]\\
 \hline
 flatten-a(Flatten) & (None,6400) & 0 & average-pooling2d-1[0][0]\\
 \hline
 Dense-a(Dense) & (None,2) & 12802 & flatten-1[0][0]\\
 \hline
 \hline
 \end{tabular}}
\caption{Complex-$1$ dataset CV-forward Residual architecture  }  
\label{tbl:resforone}  
\end{table}


\begin{table}
\centering\includegraphics[scale=0.9]{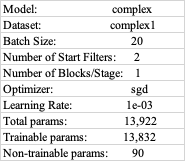}
\caption{Complex-$1$ dataset CV-forward Residual parameter settings. }
\label{tbl:resfortwo}
\end{table}

\begin{figure}
\centering\includegraphics[scale=0.78]{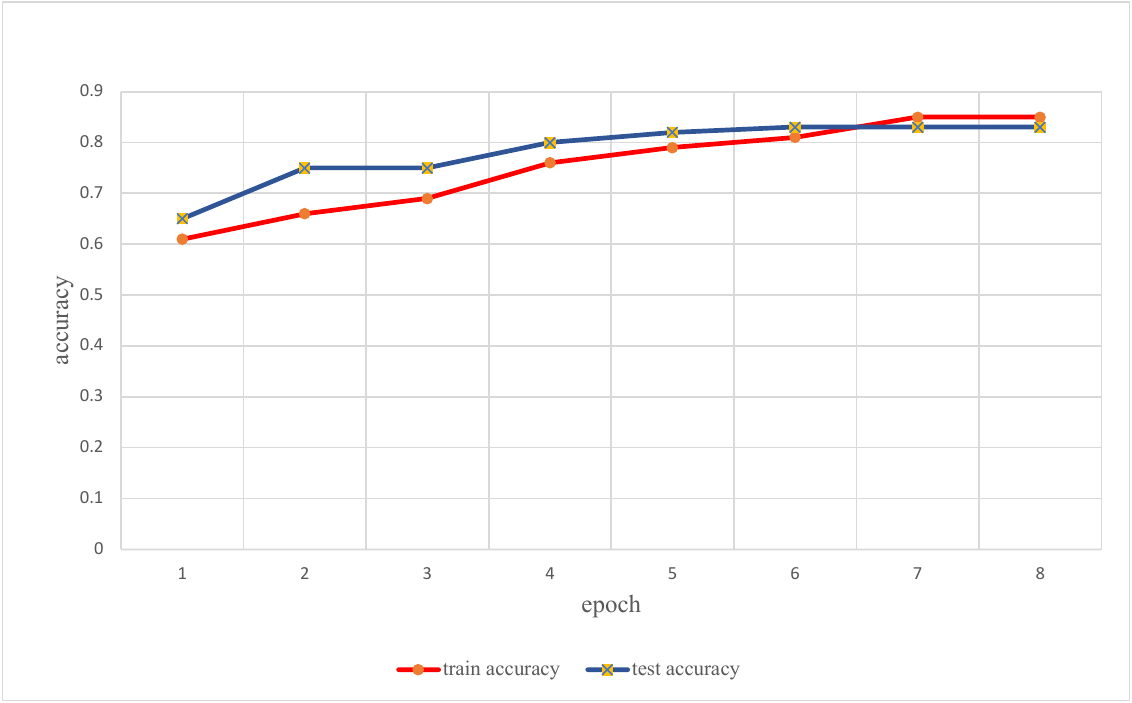}
\caption{Complex-$1$ dataset RV Residual. }
\label{fig:rvresone}
\end{figure}

\subsection{\texorpdfstring{Complex-$2$} Dataset Experiments}
\begin{figure}
\centering\includegraphics[scale=0.78]{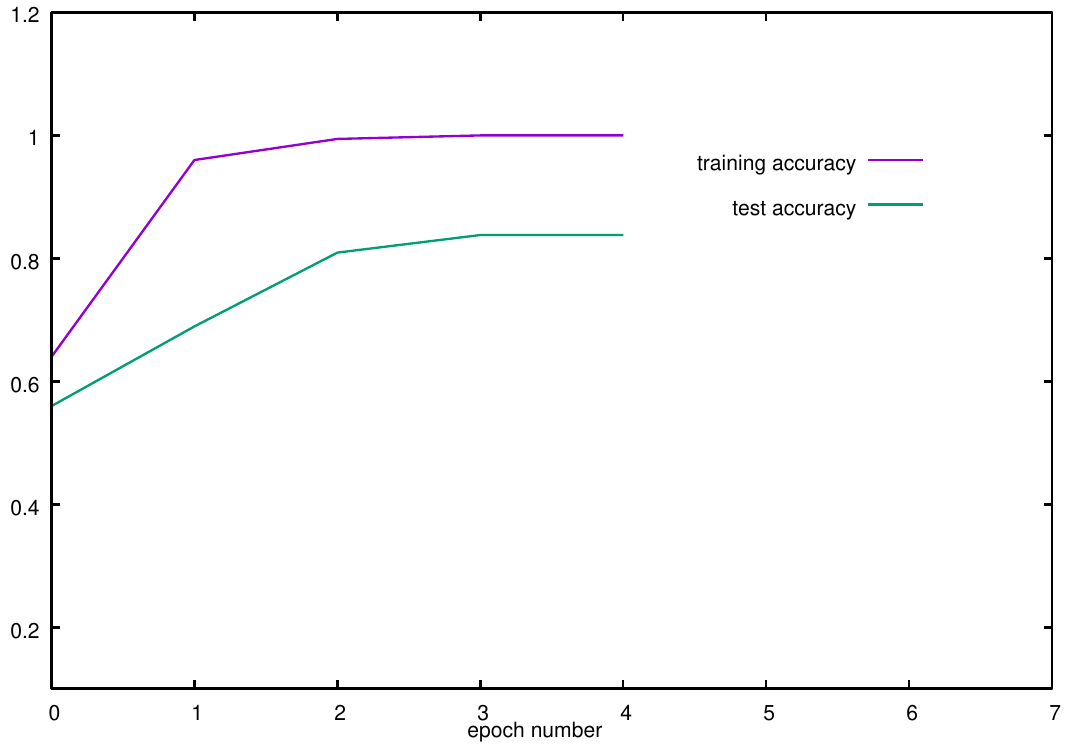}
\caption{First Experiment. Complex-$2$, CV-forward Residual accuracy per epoch (setting:~\figg{resforfour}, \figg{resforfive})  }
\label{fig:resforsix}
\end{figure}

We run $2$ experiments on our complex-$2$ dataset. In our first experiment we train CV-forward residual network training on our complex-$2$ dataset that consists of $1488$ samples of pinch and click hand gesture radar images, the dimensions of each sample is equal to $800\times231\times2$. \figg{resforfour},\figg{resforeight}and\figg{resforeight} display the parameter settings for our $2$ experiments, the number of trainable parameters for complex-$2$ dataset is $6632$k and it is same for both experiments with complex-$2$ as the network architecture is the same and the only difference is the learning rate of $1e^{-03}$ and $1e^{-03}$, the batch size is $20$ for both experiments. \fig{resforsix} shows a $100\%$ accurate test and about $82\%$ training accuracy results for both experiments with different learning rate.

\begin{figure}
\centering\includegraphics[scale=0.78]{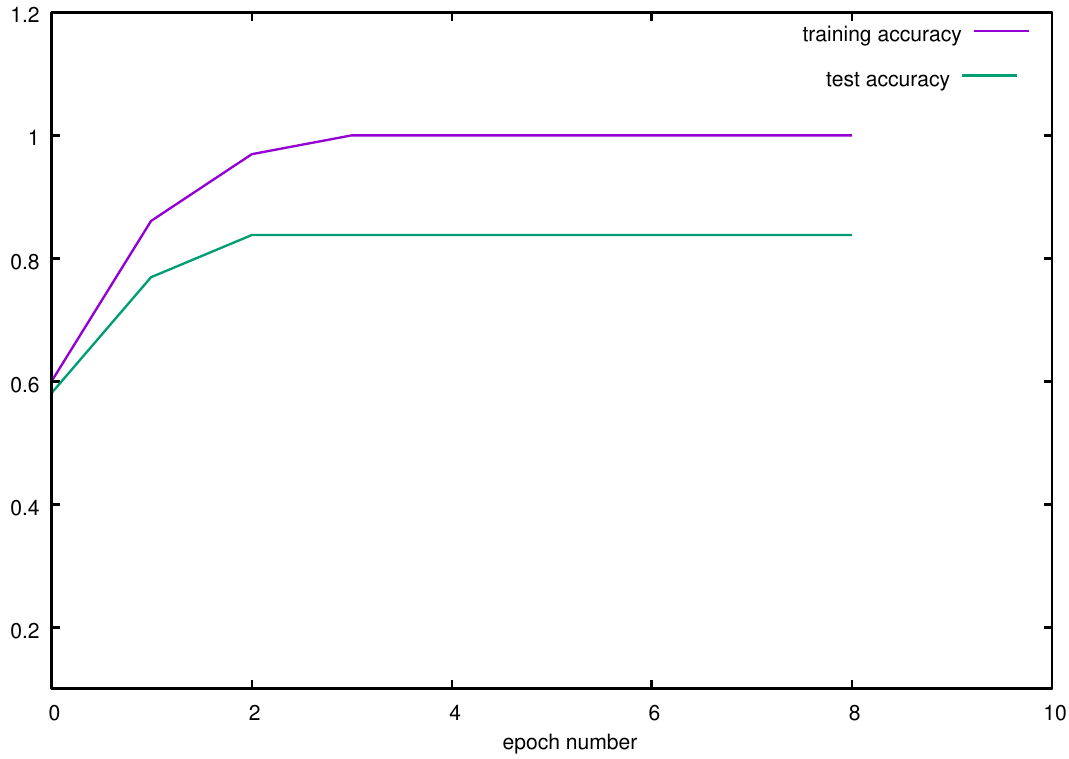}
\caption{Second Experiment, Complex-$2$, CV-forward Residual accuracy per epoch (setting:~\figg{resforfour}, \figg{resforeight})  }
\label{fig:resfornin}
\end{figure}

 We explored different combination of hyper parameters, \fig{resforsix} and  \fig{resfornin} show the results of binary classification of the hand gestures on our complex-$2$ dataset. We achieve $85\%$ test and $100\%$ in training accuracy after $4$epochs. However as \fig{rvresone} demonstrates, the equivalent RV residual network with similar parameter settings, converge a lot slower (after $8$ epochs) and the test and training accuracy are both lower at $85\%$. Thus, for the case of hand gesture binary classification, the CV-forward residual model performs more accurately and converge much faster. 

\begin{table}
\centering\scalebox{0.65}{  
 \begin{tabular}{|l|l|l|l|}
 \hline
 \hline
 Layer(type) &Output shape & Param \# & Connected to \\
 \hline
 \hline
 input-1(InputLayer) & (None,800,230,2) & 0&  \\
 \hline
 conv1(ComplexConv2D) & (None,800,230,4) & 36 & input-1[0][0] \\
 \hline
 bn-conv1-2a(ComplexBatchNorm) & (None,800,230,4) & 20 &conv1[0][0] \\
 \hline
 activation-1(Activation) & (None,800,230,4) & 0 &  bn-conv1-2a[0][0]\\
 \hline
 bn20-branch-2a(ComplexBatchNorm) & (None,800,230,4) &20 & activation-1[0][0]\\
 \hline
 activation-2(Activation) & (None,800,230,4) & 0 &  bn20-branch-2a[0][0]\\
 \hline
 res20-branch2a(ComplexConv2D) & (None,800,230,4) & 72 & activation-2[0][0]\\
 \hline
 bn20-branch-2b(ComplexBatchNorm) & (None,800,230,4) &20 & res20-branch-2a[0][0]\\
 \hline
 activation-3(Activation) & (None,800,230,4) & 0 &  bn20-branch-2b[0][0]\\
 \hline
 res20-branch2b(ComplexConv2D) & (None,800,230,4) & 72 & activation-3[0][0]\\
 \hline
 add-1(Add) &(None,800,230,4) & 0 &  res20-branch2b , activation-1\\
 \hline
 bn30-branch-2a(ComplexBatchNorm) & (None,800,230,4) &20 & add-1[0][0]\\
 \hline
 activation-4(Activation) & (None,800,230,4) & 0 &  bn30-branch-2a[0][0]\\
 \hline
 res30-branch2a(ComplexConv2D) & (None,400,116,4) & 72 & activation-4[0][0]\\
 \hline
 bn30-branch-2b(ComplexBatchNorm) & (None,400,116,4) &20 &  res30-branch2a[0][0]\\
 \hline
 activation-5(Activation) & (None,400,116,4) & 0 &  bn30-branch-2b[0][0]\\
 \hline
 res30-branch1(ComplexConv2D) & (None,400,116,4) & 8 & add-1[0][0]\\
 \hline
 res30-branch2b(ComplexConv2D) & (None,400,116,4) & 72 & activation-5[0][0]\\
 \hline
 get-real-1(GetReal) & (None,400,116,2) & 0 & res30-branch1[0][0]\\
 \hline
 get-real-2(GetReal) & (None,400,116,2) & 0 & res30-branch2b[0][0]\\
 \hline
 get-imag-1(GetImag) &(None,400,116,2) & 0 & res30-branch1[0][0]\\
 \hline
 get-imag-2(GetImag) &(None,400,116,2) & 0 & res30-branch2b[0][0]\\
 \hline
 concatenate-1(Concatenate) & (None,400,116,4) & 0 & get-real-1 ,  get-real-2\\
 \hline
 concatenate-2(Concatenate) & (None,400,116,4) & 0 & get-imag-1 ,  get-imag-2\\
 \hline
 concatenate-3(Concatenate) & (None,400,116,8) & 0 &  concatenate-1 , concatenate-2  \\
 \hline
 bn40-branch-2a(ComplexBatchNorm) & (None,400,116,8) &40 & concatenate-3[0][0]\\
 \hline
 activation-6(Activation) & (None,400,116,8) & 0 &  bn40-branch-2a[0][0]\\
 \hline
 res40-branch2a(ComplexConv2D) & (None,200,58,8) & 288 & activation-4[0][0]\\
 \hline
 bn40-branch-2b(ComplexBatchNorm) & (None,200,58,8) &40 &  res40-branch2a[0][0]\\
 \hline
 activation-7(Activation) & (None,200,58,8) & 0 &  bn40-branch-2b[0][0]\\
 \hline
 res40-branch1(ComplexConv2D) & (None,200,58,8) & 32 & concatenate-3[0][0]\\
 \hline
 res40-branch2a(ComplexConv2D) & (None,200,58,8) & 288 & activation-7[0][0]\\
 \hline
 get-real-3(GetReal) & (None,200,58,4) & 0 & res40-branch1[0][0]\\
 \hline
 get-real-4(GetReal) & (None,200,58,4) & 0 & res40-branch2b[0][0]\\
 \hline
 get-imag-3(GetImag) &(None,200,58,4) & 0 & res40-branch1[0][0]\\
 \hline
 get-imag-4(GetImag) &(None,200,58,4) & 0 & res40-branch2b[0][0]\\
 \hline
 concatenate-4(Concatenate) & (None,200,58,8) & 0 & get-real-3 ,  get-real-4\\
 \hline
 concatenate-5(Concatenate) & (None,200,58,8) & 0 & get-imag-3 ,  get-imag-4\\
 \hline
 concatenate-6(Concatenate) & (None,200,58,16) & 0 &  concatenate-4 , concatenate-5  \\
 \hline
 average-pooling2d-1(AveragePool) & (None,25,7,16) & 0 & concatenate6[0][0]\\
 \hline
 flatten-a(Flatten) & (None,2800) & 0 & average-pooling2d-1[0][0]\\
 \hline
 dense-a(Dense) & (None,2) & 5602 & flatten-1[0][0]\\
 \hline 
 \hline
 \end{tabular}}
\caption{First Experiment, Complex-$2$, CV-forward Residual architecture.  }  
\label{tbl:resforfour}  
\end{table}


\begin{table}
\centering\includegraphics[scale=0.9]{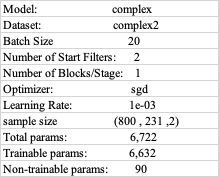}
\caption{First Experiment, Complex-$2$, CV-forward Residual parameter settings. }
\label{tbl:resforfive}
\end{table}

\begin{table}
\centering\includegraphics[scale=0.9]{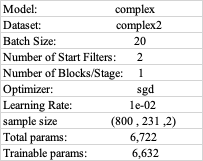}
\caption{Second Experiment, Complex-$2$, CV-forward Residual parameter settings. }
\label{tbl:resforeight}
\end{table}

\chapter{ CV-Forward CNN Network}
\label{cha:yfourcnnfor}
\section{CV-forward CNN Building Blocks}

In this chapter, we experiment the result of training a CV-forward CNN which utilises the simulated CV convolutional operation while separating the real component and imaginary component of the data as in 'CV-forward Residual' chapter of this document. We use the CV convolutional block from Python library that applies the same CV convolution technique as in CV-forward residual network (CV-forward Residual' chapter of this document). 

We produce a code with two layers of CV convolution and average pooling. The simulated convolution operation produce the same result as the CV convolution operation however the CV-forward CNN does not apply the CV back propagation and the derivatives are all RV as it is in CV-forward residual network. Each sample's real and imaginary components are separated then concatenated together, so the input dataset dimension is doubled, same as in the CV-forward residual network. The activation function used is CRelU on the convolutional layers and softmax after the flatten layer.     

\section {Proposed CV-forward CNN}
We Implemented the simulated CV-CNN block, the network consists of $2$ layers of convolutions and average pooling, with $2$ feature maps at the first convolutional layer and $4$ feature maps at the second convolutional layer. The activation function is CRelU at the convolutional layers and softmax after the flatten layer. We utilise and modify the Python code which uses the CV-CNN, CV batch normalisation and CV weigh initialisation \cite{complexnn}.

We summarise the proposed CV-forward CNN setting as :
\begin{itemize}
\item $2$ CV convolutional filters is utilised in the first convolutional layer and $4$ filters in the second convolutional layer.
\item Simulated CV-CNN is utilised. 
\item CV batch normalisation and CV weigh initialisation are implemented.
\item CV dataset is used.
\item Each input sample's real and imaginary components are separated then concatenated together, so the input dataset dimensions are doubled.
\item The activation function used is CRelU on the convolutional layers and softmax after the flatten layer.    
\item Back propagation derivatives are all in RV domain.
\end{itemize}

\section{CV Datasets experiments}

In this section we display the training and test accuracy results of implementing our proposed CV-forward CNN model binary classification on two CV radar images datasets: complex-$1$ and complex-$2$. 

For each experiment there are two tables and two graphs attached. The first table, demonstrates the architecture of the utilised CNN network and the second table, displays the parameter settings. The first graph displays the test and training accuracy that is achieved in the CV experiment and the second graph is the baseline accuracy graph from the corresponding RV network. The architecture table shows every layer's name, output dimensions and number of parameters in that layer and total number of trainable parameters of the selected architecture for the corresponding experiment.

The parameter setting table, displays the utilised kernel's(feature maps) dimensions, number of filters in each convolutional layer, learning rate value, pooling window dimensions, batch size and the number of epochs. Furthermore, we have used the CReLU activation function after each convolutional layer and a tanh function after fully connected layer for all the experiments of this chapter.

\subsection{\texorpdfstring{Complex-$1$} {Dataset Experiments}}

We train a CV-forward CNN with two convolutional layers with our complex-$1$ dataset. The input dimension is $800 \times540\times2$. We run $4$ experiments and explore how different pooling window size and different learning rate can effect the training and accuracy. The rest of the hyper parameters are set the same for all experiments.

 First of all, we compare the results of two experiments with different pooling window size. Afterwards, we compare the results of two experiments with different learning rate. Then, we explore the combination of most accurate achieved pooling window size and learning rate to achieve high accuracy with lowest number of parameters and network complexity. At the end, we compare the results of a RV equivalent network, with same architecture and parameter settings to explore the effect of CV-forward CNN over the RV-CNN.

\begin{table}
\centering\includegraphics[scale=0.9]{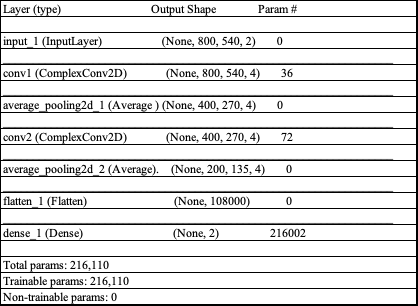}
\caption{First Experiment, Complex-$1$, CV-forward CNN architecture  }
\label{tbl:comforone}
\end{table}

\begin{table}
\centering\includegraphics[scale=0.9]{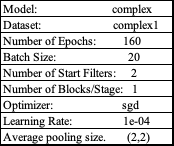}
\caption{First experiment, Complex-$1$, CV-forward CNN parameter settings. }
\label{tbl:comfortwo}
\end{table}

\begin{figure}
\centering\includegraphics[scale=0.8]{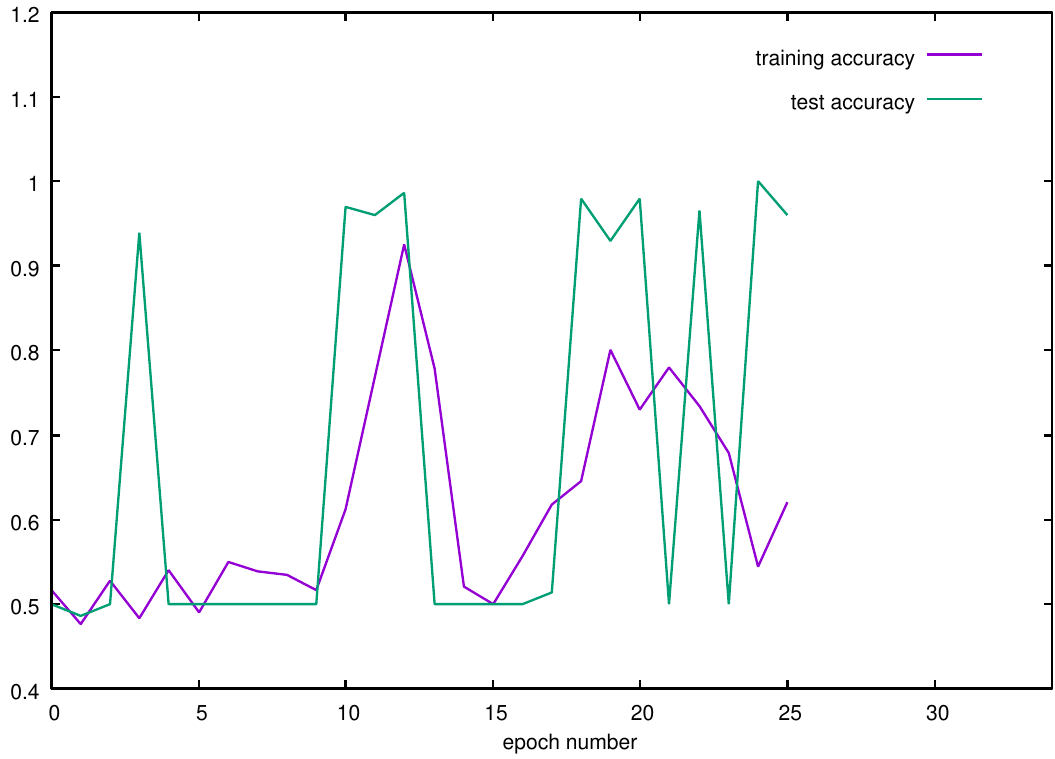}
\caption{First Experiment, Complex-$1$, CV-forward CNN accuracy per epoch (setting:\figg{comforone}, \figg{comfortwo}) . }
\label{fig:comforthree}
\end{figure}

\begin{figure}
\centering\includegraphics[scale=0.76]{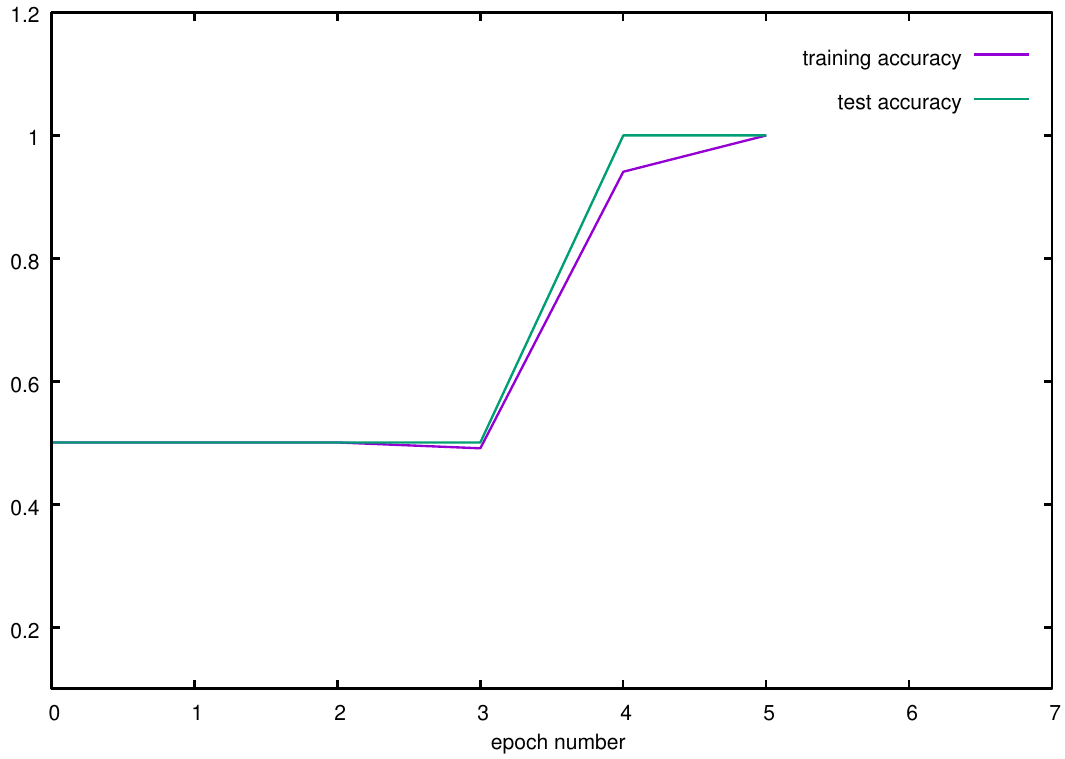}
\caption{Forth Experiment. Complex-$1$ dataset CV-forward CNN accuracy per epoch (setting: \figg{comforelev}, \figg{comfortwl} )  }
\label{fig:comforthrt}
\end{figure}

First experiment (as in \figg{comforone}, \figg{comfortwo}) and forth experiment (as in \figg{comforelev}, \figg{comfortwl}), share the same network architecture, pooling window size ($(2,2)$) parameter settings and the same number of parameters ($21$k), the only difference is their learning rate. As the results graphs show in \fig{comforthree} and \fig{comforthrt}, the smaller learning rate of $1e^{-05}$ setting improve the converging speed, forth experiment converges within $6$epochs however, first experiment with learning rate set into $1e^{-04}$ converges within $14$ epochs. Both these experiments achieve $100\%$ test and training accuracy.

\begin{table}
\centering\includegraphics[scale=0.9]{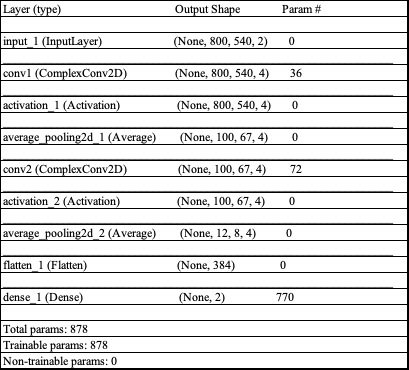}
\caption{Second experiment, Complex-$1$, CV-forward CNN architecture  }
\label{tbl:comforfour}
\end{table}

\begin{table}
\centering\includegraphics[scale=0.9]{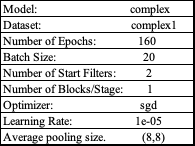}
\caption{Second experiment, Complex-$1$, CV-forward CNN parameter settings. }
\label{tbl:comforfive}
\end{table}

\begin{figure}
\centering\includegraphics[scale=0.76]{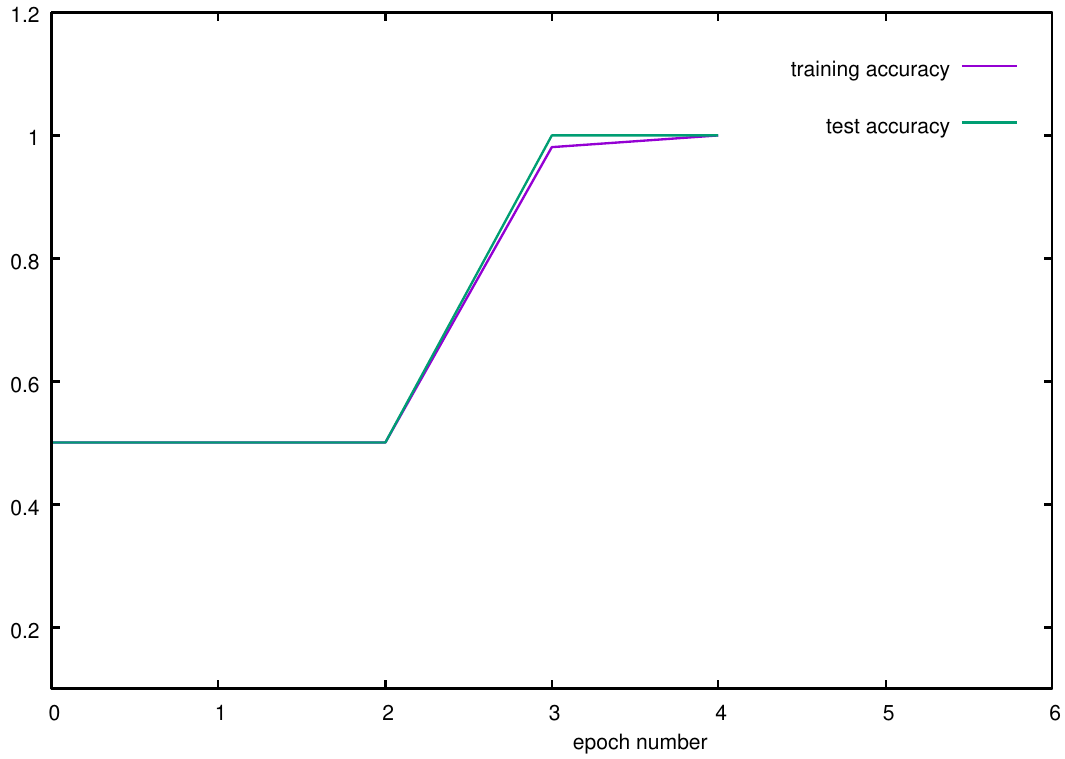}
\caption{Second Experiment, Complex-$1$, CV-forward CNN accuracy per epoch (setting:\figg{comforfive}, \figg{comforfour} )  }
\label{fig:comforsix}
\end{figure}

The forth experiment has the same hyper parameters as the first experiment \figg{comforone} \figg{comforseven} and the only difference is the value of the learning rate, the results shows that the results of training with learning rate value of $0.00001$ and $0.0001$ is the same $100\%$ accuracy but the $0.00001$ provides the accuracy faster and more stable, which proves that $0.0001$ is lare learning rate for this dataset.

\begin{table}
\centering\includegraphics[scale=0.9]{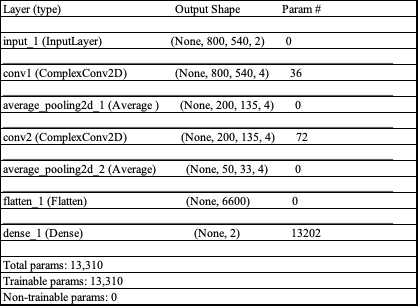}
\caption{Third experiment, Complex-$1$, CV-forward CNN architecture  }
\label{tbl:comforseven}
\end{table}

\begin{table}
\centering\includegraphics[scale=0.9]{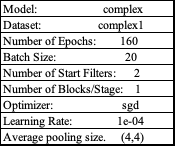}
\caption{Third experiment, Complex-$1$, CV-forward CNN parameter settings. }
\label{tbl:comforeight}
\end{table}

\begin{figure}
\centering\includegraphics[scale=0.8]{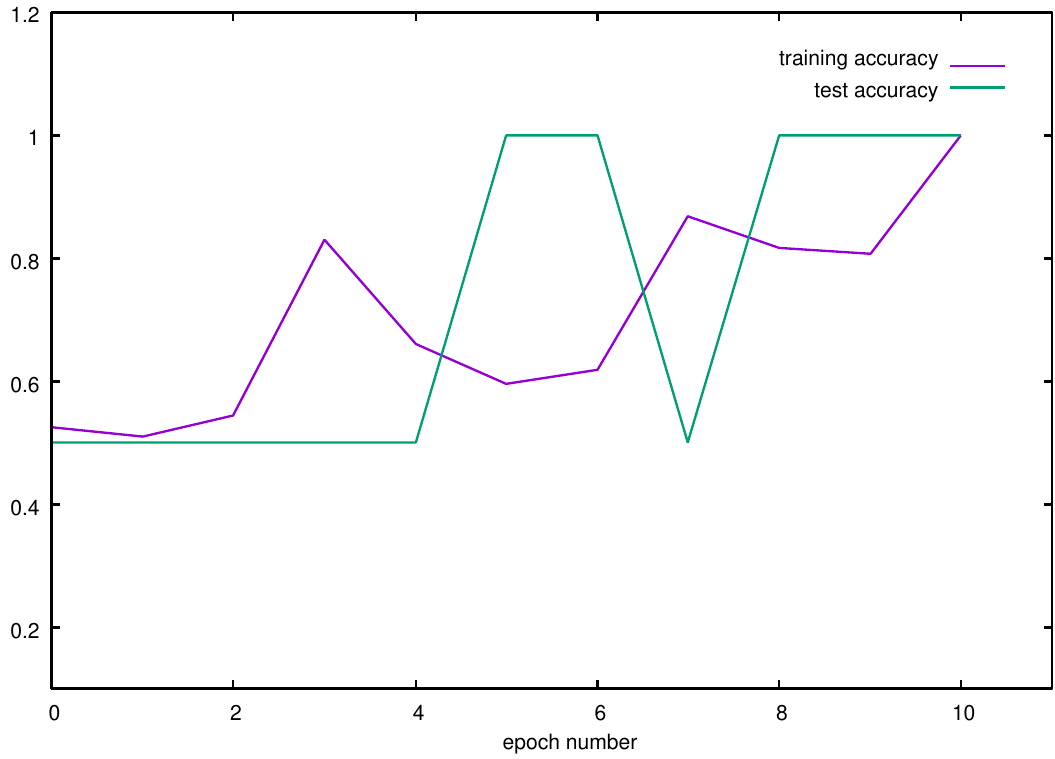}
\caption{Third Experiment, Complex-$1$, CV-forward CNN accuracy per epoch (setting: \figg{comforeight}, \figg{comforseven})  }
\label{fig:comforten}
\end{figure}

First (as in \figg{comforone}, \figg{comfortwo}) and third (as in \figg{comforseven}, \figg{comforeight}) experiments with complex-$1$ dataset, explore the effect of different pooling window size of $(2,2)$ and $(4,4)$. The first impact of larger average pooling window size is the reduction in the trainable parameters which results in faster computation as well, the number of trainable parameters reduce from $216$k with $(2,2)$ window size to $13$k with window size of $(4,4)$ which is a considerable reduction. Lower parameter numbers helps reducing the training complexity and time. In addition, \fig{comforthree} and \fig{comforten} demonstrate that both pooling window settings of $(2,2)$ and $(4,4)$ achieve a very high accuracy for test and training dataset, however in this case, the bigger pooling window size converges, over less number of epochs and higher accuracy.

\begin{table}
\centering\includegraphics[scale=0.9]{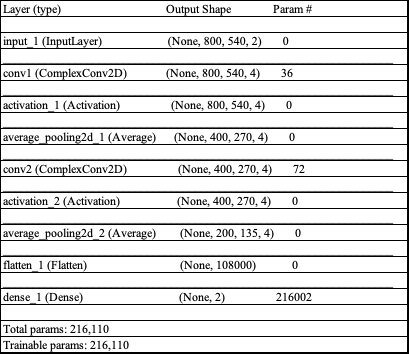}
\caption{Forth Experiment, Complex-$1$, CV-forward CNN architecture.  }
\label{tbl:comforelev}
\end{table}

\begin{table}
\centering\includegraphics[scale=0.9]{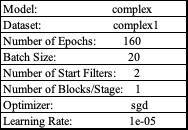}
\caption{Forth experiment, Complex-$1$, CV-forward CNN parameter settings. }
\label{tbl:comfortwl}
\end{table}

At last, we compare second (as in \figg{comforfour}, \figg{comforfive}) and forth (as in \figg{comforelev}, \figg{comfortwl}) experiments results, to explore the effect of bigger pooling window size in combination with smaller learning rate. Second and forth experiments,  share the same network architecture, same learning rate ($1e^{-05}$) and equal parameter settings, the only difference is their pooling window size and therefore  number of parameters. For the second experiment, the pooling window is $(8,8)$ and therefore $878$ parameters and for the forth one, pooling window size is set to $(2,2)$ and thus, $21$k parameters. Both experiments show the $100\%$ test and train accuracy as in \fig{comforsix} and \fig{comforthrt}. We found out that the bigger pooling window size performed the same accuracy faster.

\begin{figure}
\centering\includegraphics[scale=0.76]{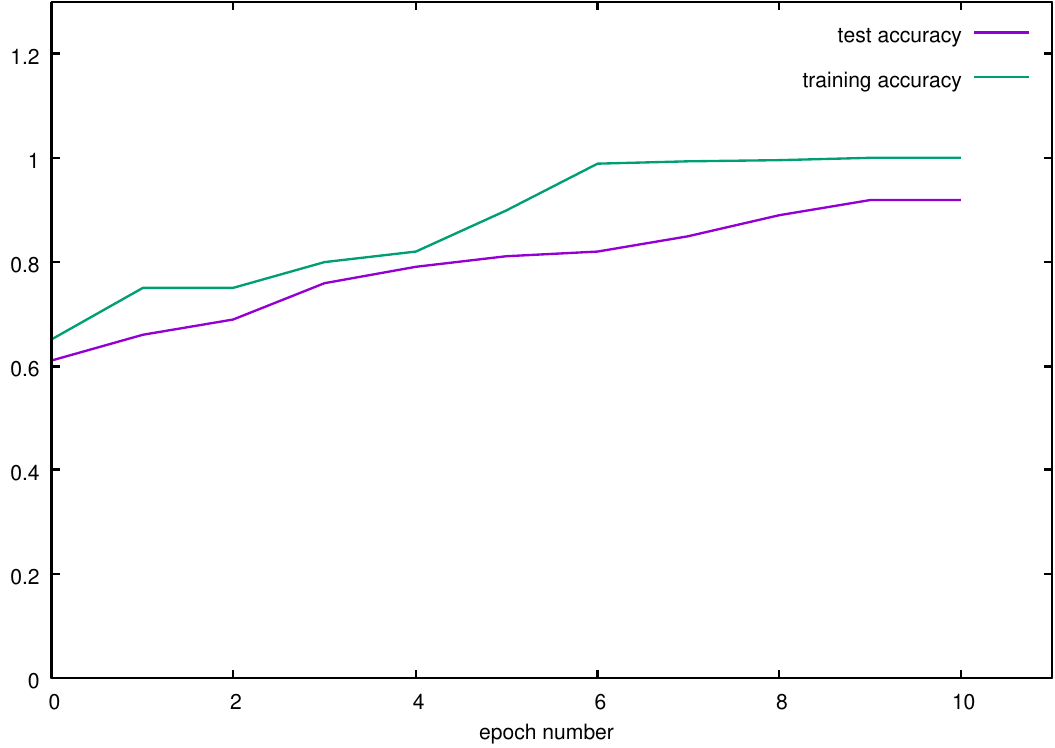}
\caption{Benchmark, Complex-$1$, RV-CNN accuracy. }
\label{fig:cvcnnbmonerecnn}
\end{figure}

\fig{cvcnnbmonerecnn} Displays the accuracy results of the equivalent RV-CNN network, with same architecture and parameter settings, as a benchmark. The equivalent RV-CNN network achieves the accuracy of $95\%$ training and $100\%$ test. In comparison with the CV-forward CNN, we observe that RV network takes longer time to converge and the final training accuracy remain lower than the $100\%$ corresponding CV network.

\subsection{\texorpdfstring{Complex-$2$} {Dataset Experiments}}

\begin{figure}
\centering\includegraphics[scale=0.8]{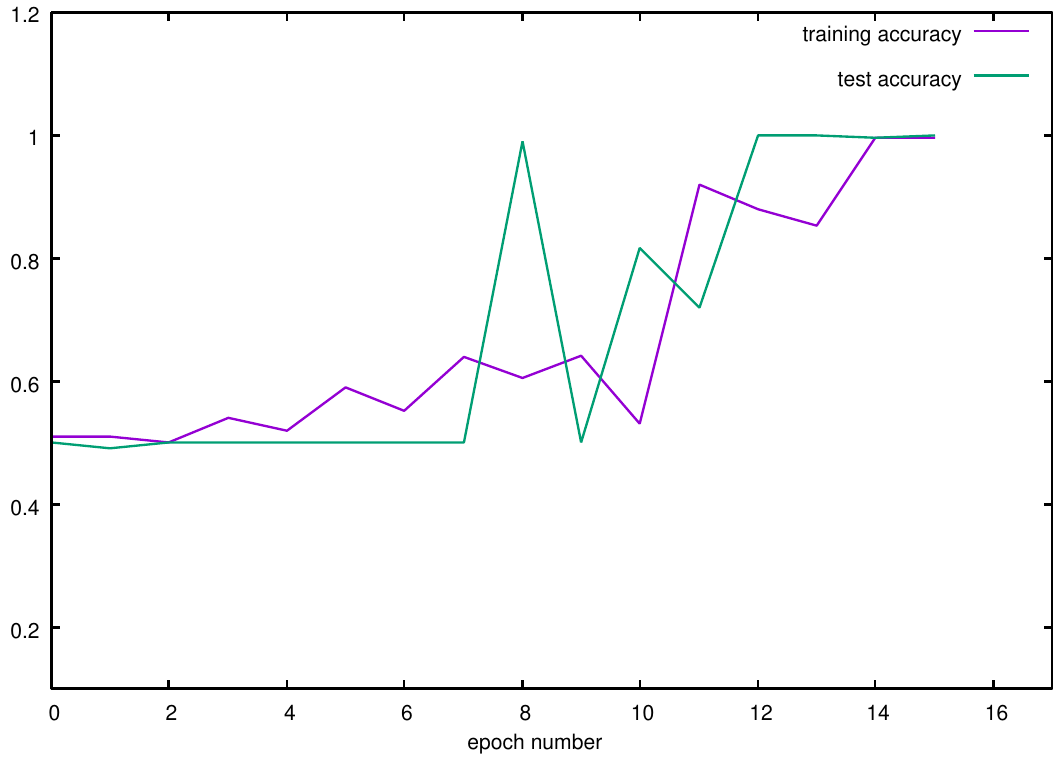}
\caption{First Experiment, Complex-$2$, CV-forward CNN accuracy per epoch (setting:\figg{comforfort}, \figg{comforfift} )  }
\label{fig:comforsixt}
\end{figure}

\begin{figure}
\centering\includegraphics[scale=0.7]{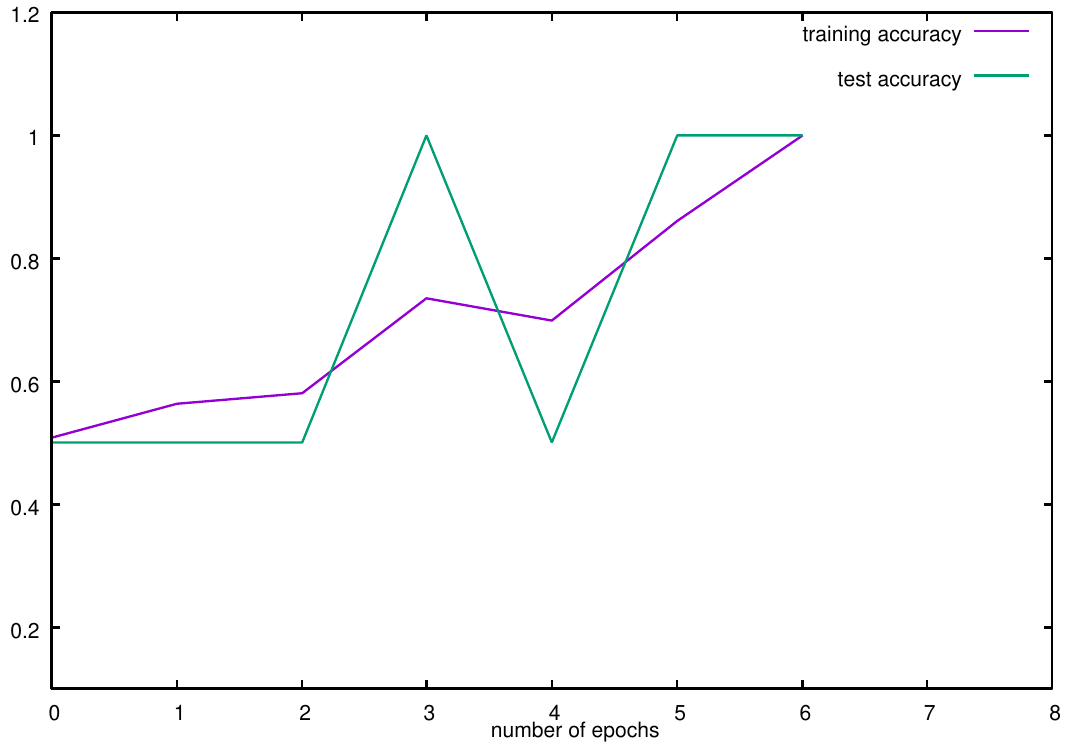}
\caption{Second Experiment, Complex-$2$, CV-forward CNN accuracy per epoch (\figg{comforeightt}, \figg{comforsevt})  }
\label{fig:comfornint}
\end{figure}

In this section we display the result of two experiments with training the CV-forward CNN with complex-$2$ dataset and we explore how the different average pooling window can effect the accuracy. \fig{comforsixt}\fig{comfornint} shows that the $(4,4)$ pooling window size provides faster ($7$ epoch) learning than the experiment with $(2,2)$. However in both experiments, the final accuracy is $100\%$. The number of parameters is less with $(4,4)$ widow as $5$k in comparison with $(2,2)$ window size with $91$k parameters which has a huge effect on the computational time.

\begin{table}
\centering\includegraphics[scale=0.9]{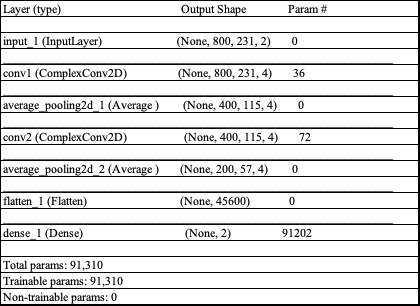}
\caption{First Experiment, Complex-$2$, CV-forward CNN architecture  }
\label{tbl:comforfort}
\end{table}

\begin{table}
\centering\includegraphics[scale=0.9]{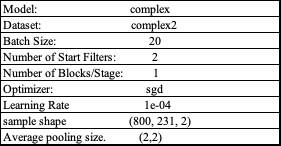}
\caption{First experiment, Complex-$2$, CV-forward CNN parameter settings. }
\label{tbl:comforfift}
\end{table}

\begin{table}
\centering\includegraphics[scale=0.9]{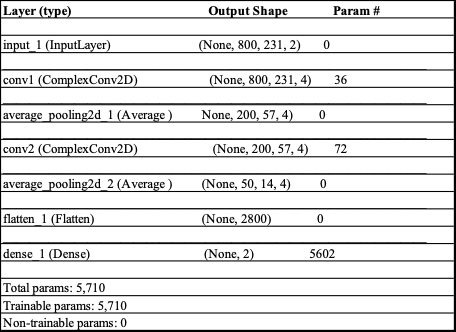}
\caption{Second Experiment, Complex-$2$, CV-forward CNN architecture  }
\label{tbl:comforsevt}
\end{table}

\begin{table}
\centering\includegraphics[scale=0.9]{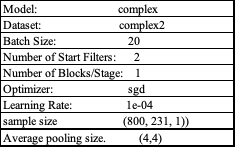}
\caption{Second experiment, Complex-$2$, CV-forward CNN parameter settings. }
\label{tbl:comforeightt}
\end{table}

At the end, we

\begin{figure}
\centering\includegraphics[scale=0.8]{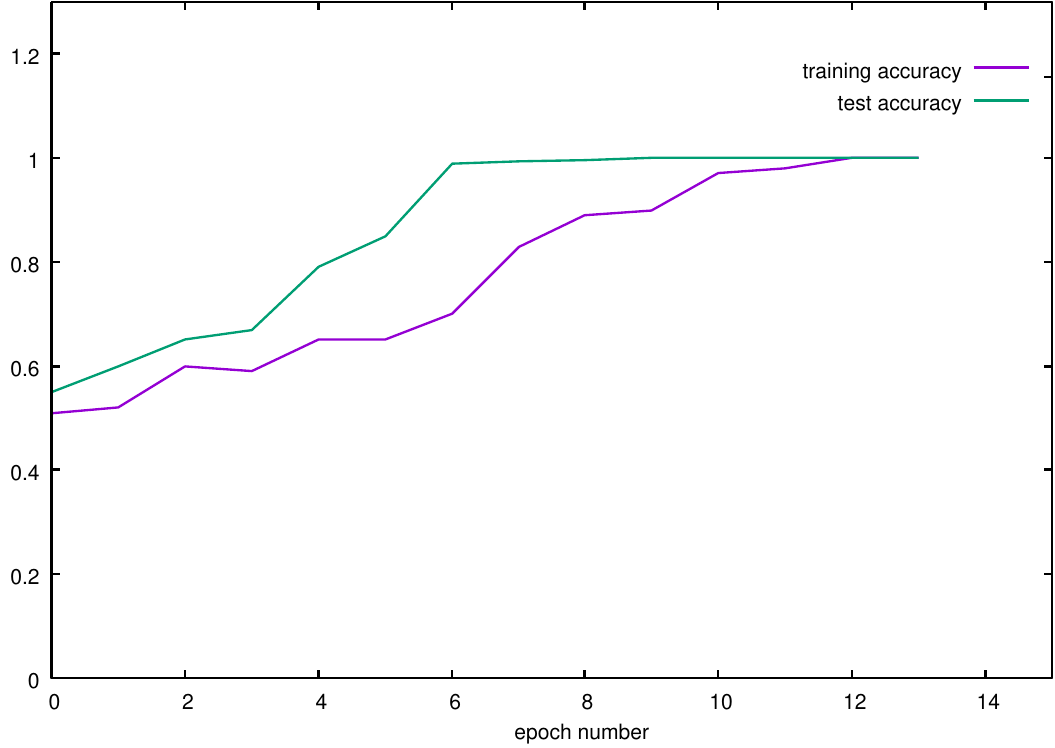}
\caption{Benchmark, Complex-$2$,  RV-CNN accuracy. }
\label{fig:cvcnnbmtworecnn}
\end{figure}

\fig{cvcnnbmtworecnn} Displays the equivalent RV-CNN network's accuracy, with same architecture and parameter settings, as a benchmark. The equivalent RV-CNN network achieves the accuracy of $100\%$ training and $100\%$ test similar to the CV-forward CNN network. In comparison with the CV-forward CNN, we observe that RV network takes longer time to converge, $12$ epochs in comparison with corresponding CV network with takes $7$ epochs to converge.

\chapter{Conclusion}
\section{The Conclusion}
 In this report we demonstrate three CV network architecture and their building blocks, in addition to the mathematical and computational details of each. We explore the accuracy of each network by training them with our hand gesture radar images of $2$ sets of CV datasets. Then we explore the efficiency of the proposed CV networks with the equivalent RV networks.

The three proposed CV networks are: fully CV-CNN, CV-forward residual network and CV-forward CNN. The details of each experiment's results, tables and graphs are shown at the end of each network's chapter. Overall the results of all three networks are remarkable and nearly $100\%$ accurate in binary classification of our radar hand gesture images.

We summarise the results of all the experiments on three CV networks as followed:

\subsection{Fully CV-CNN}

\begin{itemize}
\item The RV-CNN, converges slower than the corresponding fully CV-CNN. 
\item Back propagation and all derivatives are in CV domain.
\item In the RV-CNN network, the number of parameters are doubled, as it utilises two parallel CNNs (by separating real and imaginary components), this increases the complexity of the network and training time. We should note that the RV-CNN does not take the correlation between the real and imaginary part of the data into account. 
\item We achieve the binary classification accuracy of $100\%$ for test and over $83\%$ for training on CV-CNN for complex-$1$.
\item We achieve the binary classification accuracy of $100\%$ for test and $100\%$ for training on CV-CNN for complex-$2$.
 \item Smaller batch size demonstrates a better accuracy result and reduces the time of computation on CV-CNN experiments.
\item Very small learning rate slows the learning down and big learning rate and misses the minimum point of the loss and therefore misses the best accuracy.
  
\end{itemize}

\subsection{CV-forward Residual network}
\begin{itemize}
\item The number of parameters for the RV and CV networks are the same.
\item Back propagation and all derivatives are in RV domain.
\item The RV residual network, converges slower than the corresponding CV-forward residual one. 
\item We explored different combination of hyper parameters for complex-$1$ dataset shows, the result is very accurate in binary classification of the hand gestures on our complex-$1$ dataset. We achieve $100\%$ test and training accuracy after $3$ epoch, which is outstanding result. However the equivalent RV residual network with similar parameters setting, converge slower (after $8$ epochs) and the test and training accuracy are both lower at $84\%$. Thus, for the case of hand gesture binary classification, the CV-forward residual model performs more accurately and converge much faster.

 \item We explored different combination of hyper parameters on complex-$2$. We achieve $85\%$ test and $100\%$ in training accuracy after $4$epochs. However the equivalent RV residual network with similar parameters setting, converge a lot slower (after $8$ epochs) and the test and training accuracy are both lower at $85\%$. Thus, for the case of hand gesture binary classification, the CV-forward residual model performs more accurately and converge much faster. 
\end {itemize}

\subsection{CV-forward CNN}
\begin{itemize}
\item The number of parameters for the RV and CV networks are the same.
\item Back propagation and all derivatives are in RV domain.
\item The RV-CNN network, converges slower than the corresponding CV-forward CNN one. 
\item The equivalent RV-CNN network achieves the accuracy of $95\%$ training and $100\%$ test for complex-$1$ dataset. In comparison with the CV-forward CNN, we observe that RV network takes longer time to converge and the final training accuracy remain lower than the $100\%$ corresponding CV network.
\item  The equivalent RV-CNN network achieves the accuracy of $100\%$ training and $100\%$ test similar to the CV-forward CNN network for complex-$2$ dataset. In comparison with the CV-forward CNN, we observe that RV network takes longer time to converge, $12$ epochs in comparison with corresponding CV network with takes $7$ epochs to converge.
\end{itemize}

\section{Challenges and Future Work}

In the field of CV-CNN, there is still a gap of understanding the details of mathematical operations of the CV building blocks. Researches attempting to implement the CV-CNN, usually utilise the pre-written codes without analysing the mathematical details.

Researchers often implement a hybrid CV model and call it CV model, without paying attention to RV utilised building blocks, loss function, activation function or BP. Researchers, compare their proposed hybrid CV model with the equivalent RV network as a baseline. However in order to reach maturity in CV networks, we suggest similar CV networks to be compared against each other.

With the knowledge of the author today, there is no research or survey that paid attention to the CV- fully connected layer's functionality, which is implemented mathematically incorrect most of the time. Most researchers, implemented the CV simulated convolutional operation and applied average pooling in the CV domain, without paying any attention to the complex domain BP. There has been some research in the field of CV activation functions, there are a lot of CV activation functions that have been explored in ANNs, however only a few have been utilised in CV-CNNs.

Furthermore, the CV loss function characteristics and its CV differentiability, play important act on loss function selection, however it only received a small attention in literature. Researchers, have limited resources of programming languages pre-written libraries that implement the CV simulated mathematical operation for each block. 

Here, we summarise the recommended future works in the field of CV-CNN:

\begin{itemize}
\item A survey on the implemented CV blocks mathematical differences in the literature. In order to clarify and train the researchers on the options in the existing CV blocks.
\item Researchers in the field of CV-ANNs, need to gain information about CV functions differentiability. Thus the selection of loss function can be mathematically correct.
\item  Researchers in the field of CV-ANNs, should learn about CV functions differentiability. So they can implement the BP calculations fully CV.
\item The CV-CNN researchers need to gain information about numerous CV activation functions that have been implemented in various ANNs, but not in CV-CNN.
\item The researchers with interest in the field of CV-CNN should focus more on developing fully CV activation functions, rather than the splitted components activation functions only.
 
\item Researchers should pay attention to simulate the fully connected layer's operation mathematically correct in the CV domain.
\item There is a need for developers to prepare more pe-writtn libraries, which is aligned with the recent research achievements. In order to facilities the researchers to focus on innovating  ideas rather than spending long time on implementing a new idea on code from scratch.

\item In order for CV models reaching the next level of maturity, we recommend that it is time to compare various CV models with each other not only with a RV equivalent model as a baseline.

\end{itemize}

\bibliography{main.bib}             

\begin{thebibliography}{10}

\bibitem{g9841948}
K.~Alirezazad, , G.~Rhiel, , and L.~Maurer.
\newblock 2d cnn-gru model for multi-hand gesture recognition system using fmcw
  radar.
\newblock In {\em 2022 20th IEEE Interregional NEWCAS Conference (NEWCAS)},
  pages 158--162, 2022.

\bibitem{gshasht}
S.~Bharadwaj and A.~Nguyen.
\newblock Real-time multi-gesture recognition using 77 ghz fmcw mimo single
  chip radar.
\newblock In {\em IEEE International Conference on Consumer Electronics
  (ICCE)}, pages 1--4, 2019.

\bibitem{gstwl}
Z.~Chen, F.~Li, G.~Fioranelli, and H.~Griffiths.
\newblock Dynamic hand gesture classification based on multistatic radar
  micro-doppler signatures using convolutional neural network.
\newblock In {\em IEEE Radar Conference (RadarConf)}, pages 1--5, 2019.

\bibitem{gstten}
J.W Choi, S.J Ryu, and J.H. Kim.
\newblock Short-range radar based real-time hand gesture recognition using lstm
  encoder.
\newblock In {\em IEEE Access}, volume~7, pages 33 610--33 618, 2019.

\bibitem{gsften}
Y.~Wang, S.~Wang, M.~Zhou, Q.~Jiang, and Z.~Tian.
\newblock Ts-i3d based hand gesture recognition method with radar sensor.
\newblock In {\em IEEE Access}, volume~7, pages 22902--22913, 2019.

\bibitem{g9266565}
S.~Franceschini, M.~Ambrosanio, S.~Vitale, F.~Baselice, A.~Gifuni, G.~Grassini,
  and V.~Pascazio.
\newblock Hand gesture recognition via radar sensors and convolutional neural
  networks.
\newblock In {\em 2020 IEEE Radar Conference (RadarConf20)}, pages 1--5, 2020.

\bibitem{gsyek}
S.~Zhao, W.~Tan, C.~Wu, C.~Liu, and S~Wen.
\newblock A novel interactive method of virtual reality system based on hand
  gesture recognition.
\newblock In {\em Chinese Control and Decision Conference}, pages 5879--5882,
  2009.

\bibitem{gsdo}
S.S. Rani, K.J. Dhrisya, and M.~Ahalyadas.
\newblock Hand gesture control of virtual object in augmented reality.
\newblock In {\em International Conference on Advances in Computing
  Communications and Informatics (ICACCI)}, pages 1500--1505, 2017.

\bibitem{gsse}
Y.~Sun, T.~Fei, F.~Schliep, and N.~Pohl.
\newblock Gesture classification with handcrafted micro-doppler features using
  a fmcw radar.
\newblock In {\em EEE MTT-S International Conference on Microwaves for
  Intelligent Mobility (ICMIM)}, pages 1--4, 2018.

\bibitem{gschar}
Y.~Sun, T.~Fei, X.~Li, A.~Warnecke, E.~Warsitz, and N.~Pohl.
\newblock Real-time radar-based gesture detection and recognition built in an
  edge-computing platform.
\newblock In {\em EEE Sensors Journal}, volume~20, pages 10 706--10 716, 2020.

\bibitem{gsfften}
M.~Maghoumi and J.J. LaViola.
\newblock Deepgru: Deep gesture recognition utility.
\newblock \url{http://arxiv.org/abs/1810.12514}, 2018.
\newblock arXiv preprint 1810.12514.

\bibitem{g7152951}
Songchuan Zhang, Youshen Xia, and Jun Wang.
\newblock A complex-valued projection neural network for constrained
  optimization of real functions in complex variables.
\newblock {\em IEEE Transactions on Neural Networks and Learning Systems},
  26(12):3227--3238, 2015.

\bibitem{complexnn}
C.~Trabelsi, O.~Bilaniuk, Y.~Zhang, D.~Serdyuk, S.~Subramanian,
  J.~Felipe~Santos, S.~Mehri, N.~Rostamzadeh, Y.~Bengio, and C.J. Pal.
\newblock Deep complex networks.
\newblock {\em arXiv preprint arXiv:1705.09792}, 2017.

\bibitem{alma992975995682301631}
J.~Bell.
\newblock {\em Machine Learning}.
\newblock Wiley, 2014.

\bibitem{Kotsiantis}
S.B. Kotsiantis.
\newblock Supervised machine learning: A review of classification techniques.
\newblock {\em Informatica}, 31:249--268, 2007.

\bibitem{pythonbook}
OS~Randal.
\newblock {\em Python Machine Learning}.
\newblock PACKT, 2015.

\bibitem{regression1}
M.J. Flynn, S.~Sarkani, and T.A. Mazzuchi.
\newblock Regression analysis of automatic measurement systems.
\newblock {\em IEEE Transactions on Instrumentation and Measurement},
  58(10):3373--3379, Oct 2009.

\bibitem{ANN610}
J.J. Wang, S.G. Hu, S.T. Zhan, Q.~Luo, Q.~Yu, Z.~Liu, T.P. Chen, Y.~Yin,
  S.~Hosaka, and Y.~Liu.
\newblock Predicting house price with a memristor-based artificial neural
  network.
\newblock {\em IEEE Access}, 6:16523--16528, 2018.

\bibitem{ANN11}
J.G. Carbonell, R.S. Michalski, and T.M. Mitchell.
\newblock Machine learning: A historical and methodological analysis.
\newblock {\em AI Mag}, vol.4:69, 1983.

\bibitem{ANN12}
D.E. Rummelhart, G.E. Hinton, and R.J. Williams.
\newblock Learning internal representations by error propagation.
\newblock {\em eadings Cognit. Sci.}, vol.1:399--421, 1988.

\bibitem{ANN13}
D.E. Rummelhart, G.E. Hinton, and R.J. Williams.
\newblock Learning representations by back-propagating errors.
\newblock {\em Nature}, vol.323:533--536, 1986.

\bibitem{ANN16}
S.G. Hu, Z.~Liu, T.P. Chen, J.J. Wang, Q.~Yu, L.J. Deng, Y.~Yin, and S.~Hosaka.
\newblock Associative memory realized by a reconfigurable memristive hopfield
  neural network.
\newblock {\em Nature Commun}, vol.6, 2015.

\bibitem{ANN18}
T.~Serrano-Gotarredona, t.~Prodromakis, and B.~Linares-Barranco.
\newblock A proposal for hybrid memristor-cmos spiking neuromorphic learning
  systems.
\newblock {\em IEEE Circuits Syst. Mag.}, vol.13:74--88, 2013.

\bibitem{ANN19}
S.~Song, K.D. Miller, and L.F. Abbott.
\newblock Competitive hebbian learning through spike-timing-dependent synaptic
  plasticity.
\newblock {\em Nature Neurosci}, vol.3:919--926, 2000.

\bibitem{ANNone}
M.~Chu, B.~Kim, S.~Park, H.~Hwang, M.~Jeon, B.H. Lee, and B.~Lee.
\newblock Neuromorphic hardware system for visual pattern recognition with
  memrist\ or array and cmos neuron.
\newblock {\em IEEE Transactions on Industrial Electronics}, 62(4):2410--2419,
  April 2015.

\bibitem{ANN2}
S.~Park, M.~Chu, J.~Noh, M.~Jeon, B.~Hun~Lee, and B.G. Lee.
\newblock Electronic system with memristive synapses for pattern recognition.
\newblock {\em Sci. Rep.}, vol.5, 2015.

\bibitem{ANN3}
X.~Wu, V.~Saxena, and k.~Zhu.
\newblock Homogeneous spiking neuromorphic system for real-world pattern
  recognition.
\newblock {\em IEEE J. Emerg. Sel. Topics Circuits Syst.}, vol.5:pp. 254--266,
  Jun. 2015, 2015.

\bibitem{ANN4}
k.~Curran, X.~Li, and N~McCaughley.
\newblock Neural network face detection.
\newblock {\em Imag. Sci. J.}, vol.53, no.2:105--115, 2013.

\bibitem{ANN5}
S.S. Farfade, M.J. Saberian, and L.J. Li.
\newblock Multi-view face detection using deep convolutional neural networks.
\newblock {\em Proc. 5th ACM Int. Conf. Multimedia Retr}, pages 643--650, 2015.

\bibitem{ANN6}
Q.V. Le.
\newblock Building high-level features using large scale unsupervised learning.
\newblock {\em Proc. IEEE Int. Conf. Acoust. Speech Signal Process.}, pages
  8595--8598, 2013.

\bibitem{ANN7}
P.~A. Merolla, J.V. Arthur, R.~Alvarez-Icaza, A.S. Cassidy, and J.~Sawada.
\newblock A million spiking-neuron integrated circuit with a scalable
  communication network and interface.
\newblock {\em Science}, vol. 345, no. 6197,:668--673, 2014.

\bibitem{ANN8}
D.~Silver.
\newblock Mastering the game of go with deep neural networks and tree search.
\newblock {\em Nature}, 529:484--489, 2016.

\bibitem{datacamp}
K.~Willems.
\newblock Keras tutorial: Deep learning in python, 2019.

\bibitem{backprop}
Y.A. LeCun, L.~Bottou, G.B. Orr, and k.B. Muller.
\newblock {\em Efficient backprop}, pages 9--48.
\newblock Lecture Notes in Computer Science (including subseries Lecture Notes
  in Artificial Intelligence and Lecture Notes in Bioinformatics). Springer
  Verlag, 2012.

\bibitem{polyak}
B.T. Polyak.
\newblock Some methods of speeding up the convergence of iteration methods.
\newblock {\em USSR Computational Mathematics and Mathematical Physics},
  4(5):1--17, 1964.

\bibitem{Rumelhart}
D.~E. Rumelhart, G.~E. Hinton, and R.~J. Williams.
\newblock Parallel distributed processing.
\newblock {\em MIT Press, Cambridge, MA}, 1:318,362, 1986.

\bibitem{Sebastian}
S.~Ruder.
\newblock An overview of gradient descent optimization algorithms.
\newblock {\em Insight Centre for Data Analytics, NUI Galway Aylien Ltd.,
  Dublin}, 2017.

\bibitem{activationfunction}
7-types of neural network activation functions, 2019.

\bibitem{regimpact}
K.~Zeeshan.
\newblock The impact of regularisation on convolutional neural networks.
\newblock {\em semantic scholar}, 2018.

\bibitem{optcom}
O.~Yadanar and K.M. Soe.
\newblock Optimizer comparison with dropout for neural sequence labelling in
  myanmar stemmer.
\newblock In {\em IEEE International Conference on Industry 4.0, Artificial
  Intelligence, and Communications Technology (IAICT)}, 2019.

\bibitem{regsriv}
N.~Srivastava, G.~Hinton, A.~Krizhevsky, I.~Sutskever, and R.~Salakhutdinov.
\newblock A simple way to prevent neural networks from over-fitting.
\newblock {\em Machine Learning Research}, 15:1929--1958, 2014.

\bibitem{regsantur}
S.~Santurkar, D.~Tsipras, A.~Ilyas, and A.~Madry.
\newblock How does batch normalization help optimization?
\newblock {\em arXiv:1805.11604}, 2018.

\bibitem{reggong}
K.~Yu, W.~Xu, and Y.~Gong.
\newblock Deep learning with kernel regularization for visual recognition.
\newblock In {\em In Advances in Neural Information Processing Systems (NIPS)},
  2009.

\bibitem{NN1}
M.A. Nielsen.
\newblock {\em Neural networks and deep learning}.
\newblock Determination press, 2015.

\bibitem{gescnndo}
Gyutae Park, V.~K. Chandrasegar, JoongGun Park, and Jinhwan Koh.
\newblock Increasing accuracy of hand gesture recognition using convolutional
  neural network.
\newblock In {\em 2022 International Conference on Artificial Intelligence in
  Information and Communication (ICAIIC)}, pages 251--255, 2022.

\bibitem{Beltran-Hernandez2017}
C.~Beltran-Hernandez, J.~A. Chacon-Galindo, and L.~G. De-La-Fraga.
\newblock Radar-based hand gesture recognition using frequency modulated
  continuous wave radar.
\newblock In {\em 2017 IEEE International Autumn Meeting on Power, Electronics
  and Computing (ROPEC)}, pages 1--5. IEEE, 2017.

\bibitem{Anil2016}
K.~R. Anil and K.~N. Nandakumar.
\newblock Comparative study of vision-based and radar-based hand gesture
  recognition systems.
\newblock {\em International Journal of Advanced Research in Computer Science
  and Software Engineering (IJARCSSE)}, 6(12):127--131, 2016.

\bibitem{8835661}
Moeness~G. Amin, Zhengxin Zeng, and Tao Shan.
\newblock Hand gesture recognition based on radar micro-doppler signature
  envelopes.
\newblock In {\em 2019 IEEE Radar Conference (RadarConf)}, pages 1--6, 2019.

\bibitem{FeatureExtraction}
Yong Wang, Shasha Wang, Mu~Zhou, Wei Nie, Xiaolong Yang, and Zengshan Tian.
\newblock Two-stream time sequential network based hand gesture recognition
  method using radar sensor.
\newblock In {\em 2019 IEEE Globecom Workshops (GC Wkshps)}, pages 1--6, 2019.

\bibitem{gescnnyek}
Takuya Sakamoto, Xiaomeng Gao, Ehsan Yavari, Ashikur Rahman, Olga
  Boric-Lubecke, and Victor~M. Lubecke.
\newblock Hand gesture recognition using a radar echo i–q plot and a
  convolutional neural network.
\newblock {\em IEEE Sensors Letters}, 2(3):1--4, 2018.

\bibitem{gescnnsurv}
Sruthy Skaria, Akram Al-Hourani, and Robin~J. Evans.
\newblock Deep-learning methods for hand-gesture recognition using
  ultra-wideband radar.
\newblock {\em IEEE Access}, 8:203580--203590, 2020.

\bibitem{realtime}
Jun~Seuk Suh, Siiung Ryu, Bvunghun Han, Jaewoo Choi, Jong-Hwan Kim, and
  Songcheol Hong.
\newblock 24 ghz fmcw radar system for real-time hand gesture recognition using
  lstm.
\newblock In {\em 2018 Asia-Pacific Microwave Conference (APMC)}, pages
  860--862, 2018.

\bibitem{kumar2020radar}
M.~Kumar and A.~Kumar.
\newblock Radar-based hand gesture recognition using machine learning
  techniques: A survey.
\newblock {\em IEEE Sensors Journal}, 20(24):14582--14600, 2020.

\bibitem{hashtom}
S.~Yoo et~al.
\newblock Radar recorded child vital sign public dataset and deep
  learning-based age group classification framework for vehicular application.
\newblock {\em Sensors}, 21(7):2412, 2021.

\bibitem{nohom}
P.~Addabbo, M.~L. Bernardi, F.~Biondi, M.~Cimitile, C.~Clemente, and
  D.~Orlando.
\newblock Gait recognition using fmcw radar and temporal convolutional deep
  neural networks.
\newblock In {\em Proc. IEEE 7th Int. Workshop Metrol. AeroSp.
  (MetroAeroSpace)}, pages 171--175, 2020.

\bibitem{dahom}
P.~Molchanov, S.~Gupta, K.~Kim, and K.~Pulli.
\newblock Short-range fmcw monopulse radar for hand-gesture sensing.
\newblock In {\em Proc. IEEE Radar Conf. (RadarCon)}, pages 1491--1496, 2015.

\bibitem{yazdahom}
Z.~Yu, D.~Zhang, Z.~Wang, Q.~Han, B.~Guo, and Q.~Wang.
\newblock Sodar: Multitarget gesture recognition based on simo doppler radar.
\newblock {\em IEEE Trans. Human-Mach. Syst.}, 52(2):276--289, Apr. 2022.

\bibitem{davazdahom}
J.~Lien et~al.
\newblock Soli: Ubiquitous gesture sensing with millimeter wave radar.
\newblock {\em ACM Trans. Graphics}, 35(4):1--19, 2016.

\bibitem{gescnnse}
Desheng Liu and Hongfu Meng.
\newblock Application of fmcw radar for the recongnition of hand gesture using
  time series convolutional neural networks.
\newblock In {\em 2020 International Conference on Microwave and Millimeter
  Wave Technology (ICMMT)}, pages 1--3, 2020.

\bibitem{vchen}
V.C. Chen, L.~Fayin, S.S. Ho, and H.~Wechsler.
\newblock Micro-doppler effect in radar: phenomenon, model, and simulation
  study.
\newblock {\em IEEE\ Trans.\ Aerosp.\ Electron.\ Syst.}, 42, 2006.

\bibitem{dopnet}
M.~Ritchie, R.~Capraru, and F.~Fioranelli.
\newblock What is dopnet?

\bibitem{fiorone}
F.~Fioranelli, M.~Ritchie, and H.~Griffiths.
\newblock Centroid features for classification of armed/unarmed multiple
  personnel using multistatic human micro-doppler.
\newblock {\em IET Radar, Sonar Navig}, 10 no.9, 2016.

\bibitem{fiortwo}
F.~Fioranelli, M.~Ritchie, S.Z. Gürbüz, and H.~Griffiths.
\newblock Feature diversity for optimized human micro-doppler classification
  using multistatic radar.
\newblock {\em IEEE Trans. Aerosp. Electron. Syst.}, 53 no.2, 2017.

\bibitem{fiorthree}
F.~Fioranelli, M.~Ritchie, and H.~Griffiths.
\newblock Multistatic human micro-doppler classification of armed/unarmed
  personnel.
\newblock {\em IET Radar, Sonar Navig}, 9, no. 7, 2015.

\bibitem{tah}
D~Tahmoush and J.~Silvious.
\newblock Remote detection of humans and animals.
\newblock In {\em 2009 IEEE Applied Imagery Pattern Recognition Workshop (AIPR
  2009)}, pages 1--8, 2009.

\bibitem{survey11}
D.~Ciresan, A.~Giusti, L.M. Gambardekka, and J.~Schmidhuber.
\newblock Deep neural networks segment neural membranes in electron microscopy
  images.
\newblock {\em Advanced in neural information processing systems}, 2012.

\bibitem{survey}
A.~Khan, A.~Sohail, U.~Zahoora, and A.S. Qureshi.
\newblock A survey of the recent architecture of deep convolutional neural
  network.
\newblock {\em arXiv}, 2019.

\bibitem{activechar}
R.~Zahedinasab and H~Mohseni.
\newblock Using deep convolutional neural networks with adaptive activation
  functions for medical ct brain image classification.
\newblock In {\em 2018 25th National and 3rd International Iranian Conference
  on Bio-medical Engineering (ICBME)}, 2018.

\bibitem{activepanj}
S.~Qian, H.~Liu, C.~Liu, S.~Wu, and H.~S. Wong.
\newblock Adaptive activation functions in convolutional neural networks.
\newblock {\em Neurocomputing}, 272, 2017.

\bibitem{activeshish}
V.~Nair and G.~E. Hinten.
\newblock Rectified linear units improve restricted boltzmann machines.
\newblock In {\em Proc. 27th Int. Conf.}, 2010.

\bibitem{activeyek}
A.~Krizhevsky, I.~Sutskever, and G.E. Hinton.
\newblock Imagenet classification with deep convolutional neural networks.
\newblock {\em Proceedings of Advances in neural information processing
  systems}, 2012.

\bibitem{activese}
B.~Ding, Qian H., and J~Zhou.
\newblock Activation functions and their characteristics in deep neural
  networks.
\newblock In {\em Chinese Control And Decision Conference (CCDC)}, 2018.

\bibitem{activedo}
X.~Glorot and Y.~Bengio.
\newblock Understanding the difficulty of training deep feed forward neural
  networks.
\newblock {\em Journal of Machine Learning Research}, 9, 2010.

\bibitem{recactive}
Y.~Ying, J.~Su, P.~Shan, L.~Miao, X.~Wang, and S.~Peng.
\newblock Rectified exponential units for convolutional neural networks.
\newblock {\em IEEE Access}, 7, 2019.

\bibitem{rect8}
A.L Maas, A.~Y. Hannun, and A.Y. Ng.
\newblock Rectifier nonlinearities improve neural network acoustic models.
\newblock In {\em Proc. 30th Int. Conf. Mach. Learn. Workshop}, volume~28,
  2013.

\bibitem{rect9}
K.~He, X.~Zhang, S.~Ren, and J.~Sun.
\newblock Delving deep into rectifiers: Surpassing human-level performance on
  imagenet classification.
\newblock In {\em Proc. IEEE Int. Conf. Comput. Vis.}, 2015.

\bibitem{rect10}
D.~A. Clevert, T.~Unterthiner, and S.~Hochreiter.
\newblock Fast and accurate deep network learning by exponential linear units
  {(ELUs)}.
\newblock {\em arXiv:1511.07289}, 2015.

\bibitem{oncomplex}
N.~Guberman.
\newblock {\em On complex valued convolutional neural networks}.
\newblock PhD thesis, School of Computer Science and Engineering, The Hebrew
  University of Jerusalem, 2016.

\bibitem{deepcvNW}
C.~Trabelsi, O.~Bilaniuk, Y.~Zhang, D.~Serdyuk, S.~Subramanian, J.F Santos,
  S.~Mehri, N.~Rostamzadeh, Y.~Bengio, and C.J. Pal.
\newblock Deep complex networks.
\newblock In {\em International Conference on Learning Representations}, 2018.

\bibitem{hefift}
K.~He, X.~Zhang, S.~Ren, and J.~Sun.
\newblock Deep residual learning for image recognition.
\newblock {\em arXiv}, 1512.03385, 2015a.

\bibitem{hesix}
K.~He, X.~Zhang, S.~Ren, and J.~Sun.
\newblock Identity mappings in deep residual networks.
\newblock {\em arXiv}, 1603.05027, 2016.

\bibitem{iof}
S.~Ioffe and C.~Szegedy.
\newblock Batch normalization: Accelerating deep network training by reducing
  internal covariate shift.
\newblock {\em arXiv preprint arXiv:1502.03167}, 2015.

\bibitem{hirose}
S.~Hirose, A.~andYashida.
\newblock Generalization characteristics of complex-valued feed forward neural
  networks in relation to signal coherence.
\newblock {\em IEEE Trans. Neural Netw.}, 23(4): 541-551, 2012.

\bibitem{arjovsky}
M.~Arjovsky, A~Shah, and Y.~Bengio.
\newblock unitary evolution recurrent neural networks.
\newblock {\em arXiv:1511.06464}, 2015.

\bibitem{danihelka}
I.~Danihelka, G.~Wayne, B.~Uria, N.~Kalchbrenner, and A.~Graves.
\newblock Associative long short term memory.
\newblock {\em arXiv:1602.03032}, 2016.

\bibitem{wisdom}
S.~Wisdom, J.~POWERS, T.~Hershey, J.L. Rouz, and L.~atlas.
\newblock Full capacity unitary recurrent neural networks.
\newblock In {\em In advances in neural information processing system}, pages
  4880--4888, 2016.

\bibitem{ciften}
B.~Recht, R.~Roelofs, L.~Schmidt, and V.~Shankar.
\newblock Do cifar-10 classifiers generalize to cifar-10.
\newblock {\em arXiv}, 1806.00451, 2018.

\bibitem{cifhamner}
B.~Hamner.
\newblock Popular datasets over time - kaggle notebook.
\newblock
  \url{https://www.kaggle.com/behamner/popular-datasets-over-time/code}, 2021.
\newblock Accessed: 2024-09-11.

\bibitem{test}
John Doe.
\newblock A sample article.
\newblock {\em Sample Journal}, 1(1):1--10, 2024.

\end{thebibliography}

\bibliographystyle{unsrt}       

\appendix

\chapter{abbreviations}
\begin{tabular}{ll}
{ML} & machine learning\\
{FMCM} & frequency modulated continuous wave\\
{CNN} & convolutional neural network\\
{CN-CNN} & Complex value CNN\\
{ANN} & artificial neural network\\
{2D} & two dimensional\\
{3D} & three dimensional\\
{POLSAR} & polarimetric synthetic aperture radar\\
{BP} & back propagation\\
{SGD} & stochastic gradient descent\\
{RelU} & rectified linear unit\\
{LRelU} & leaky rectified linear unit\\
PRelU & parametric rectified linear unit\\
ELV & exponential linear unit\\
{AF} & activation function\\
{LOOCV} & leave one out cross validation\\
{Conv1} & first convolutional layer\\
{Conv2} & second convolutional layer\\
MSE & mean square error\\
ICS & internal covariate shift\\
\end{tabular}

\chapter{notations}
\begin{tabular}{ll}
$\mym$ & number of samples in the dataset\\
${\mym}_{tr}$ & number of samples in the training dataset\\
${\mym}_{val}$ & number of samples in the validation dataset\\
${\mym}_{tst}$ & number of samples in the test dataset\\
${\myvec{X}}^{(m)}$ & the m-th input sample matrix\\
${\myvec{Y}}^{(m)}$ & the m-th label sample vector\\
$y$ & scalar output\\
$\haty$ & the predicted output\\
$\Ivec$ & one input sample matrix\\
$\alpha \times \beta$ & input sample matrix's dimensions\\
$\myvec{W}_{1}$ & the first convolutional layer's weight tensor\\ 
$\dl{1} \times \dl{1} \times \kl{1}$ & $\myvec{W}_{1}$ 's dimensions\\
$\wonekone$ & the $\kpl{1}$th plane of $\myvec{W}_{1}$ \\
$\myvec{b}_{1}$ & the first convolutional layer's bias vector\\
$\myvec{O}_{1}$ & first convolutional layer's result after the activation function\\
$\oonekone$ & the $\kpl{1}$th plane of $\myvec{O}_{1}$\\
$\myvec{V_{1}}$ & the first convolutional layer's weighted matrix\\
$\alphl{V_{1}} \times \betl{V^{1}} \times \kl{1}$ & the dimensions of $\vonekone$ \\
$\vonekone$ & the $\kpl{1}$th plane of $\myvec{V}_{1}$\\
$\myvec{S_{1}}$ & the first convolutional layer's pooling result\\
$\myvec{W}_{2}$ & the second convolutional layer's weight tensor\\
$\dl{2} \times \dl{2} \times (\kl{1} \times \kl{2})$  & $\myvec{W}_{2}$ 's dimensions\\
$\wtwoktwo$ & the $\kpl{2}$th plane of $\myvec{W}_{2}$ \\
\end{tabular}
\\
\\
\\
\begin{tabular}{ll}
$\myvec{b}_{2}$ & the first convolutional layer's bias vector\\
$\myvec{O}_{2}$ & second convolutional layer's result after the activation function\\
$\otwoktwo$ & the $\kpl{2}$th plane of $\myvec{O}_{2}$\\
$\myvec{V_{2}}$ & the second convolutional layer's weighted matrix\\
$\alphl{V_{2}} \times \betl{V_{2}} \times (\kl{1} \times \kl{2})$ & the dimensions of $\vtwoktwo$ \\
$\vtwoktwo$ & the $\kpl{2}$th plane of $\myvec{V}_{2}$\\
$\myvec{S_{2}}$ & the second convolutional layer's pooling result\\
$\sigma$ & the activation function\\
$\kl{1}$ & number of kernel maps in the first convolutional layer\\
$\kl{2}$ & number of kernel maps in the second convolutional layer\\
$\myvec{f}$ & vectorized $\myvec{S_{2}}$\\
$\kl{fc}$ & the dimensions of $\myvec{f}$\\
${\vecwthree}$ & (layer three) the fully connected layer's weight vector\\
$b_{3}$ & the fully connected layer's bias scalar value\\
$\vecvthree$ & the weighted vector of fully connected layer\\
$\Re$ & real part\\
$\Im$ & imaginary part\\
$\mathbb{C}$ & complex numbers\\
$\mathbb{R}$ & real numbers\\
$\otimes$ & convolutional operation\\
$\odot$ & element-wise multiplication operaxtion\\ 
$\myL$ & loss function\\
$\nabll{\haty}$ & loss gradient with respect to $\haty$\\
$\nabll{\vecwthree}$ & loss gradient with respect to $\vecwthree$\\
$\nabll{b_{3}}$ & loss gradient with respect to $b_{3}$\\
$\nabll{\myvec{\wtwoktwo}}$ & loss gradient with respect to $\myvec{\wtwoktwo}$\\
$\nabll{\myvec{f}}$ & loss gradient with respect to $\myvec{f}$\\
$\nabll{\myvec{\otwoktwo}}$ & loss gradient with respect to $\myvec{\otwoktwo}$\\
\end{tabular}
\\
\\
\\
\begin{tabular}{ll}
$\nabll{\myvec{\vtwoktwo}}$ & loss gradient with respect to $\myvec{\vtwoktwo}$\\
$\nabll{\myvec{\btwoktwo}}$ & loss gradient with respect to $\myvec{\btwoktwo}$\\
$\nabll{\myvec{\wtwoktwo}}$ & loss gradient with respect to $\myvec{\wtwoktwo}$\\
$\nabll{\myvec{\oonekone}}$ & loss gradient with respect to $\myvec{\oonekone}$\\
$\nabll{\myvec{\vonekone}}$ & loss gradient with respect to $\myvec{\vonekone}$\\
$\nabll{\myvec{\bonekone}}$ & loss gradient with respect to $\myvec{\bonekone}$\\
$\nabll{\myvec{\wonekone}}$ & loss gradient with respect to $\myvec{\wonekone}$\\
$\eta$ & learning rate\\
$t$ & iteration number\\
${\myL}_{tr}$ & training loss\\
${\myL}_{val}$ & validation loss\\ 
${\myL}_{tst}$ & test loss\\
${\myL}_{1}$ & ${\myL}_{1}$ norm regularised loss function\\
${\myL}_{2}$ & ${\myL}_{2}$ norm regularised loss function\\ 
$\lambda$ & regularisation parameter\\
$\cal{W}$ & all network's weights parameters vector\\
$\cal{b}$ & all network's biases parameters vector\\
\end{tabular}

\chapter{Real Value Convolutional Neural network }
\label{cha:ytwoappreal}

\begin{figure}
\centering\includegraphics[scale=0.9]{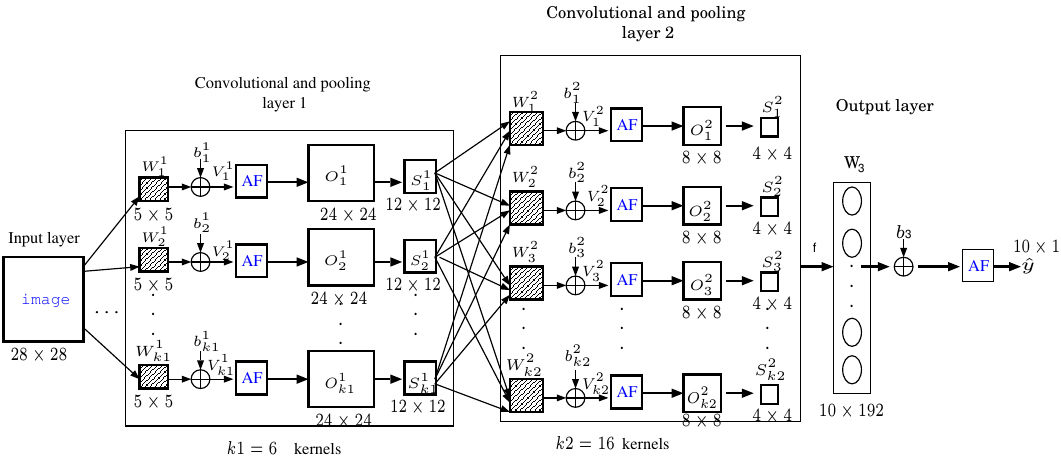}
\caption{The real value-CNN architecture }
\label{fig:realconv}
\end{figure}

This document explains all mathematical equations for a three layer real value CNN network. The architecture of the real value CNN is illustrated in \fig{realconv}, we have 2 convolutional layers and one fully connected layer, the 2D input image dimensions is $28 \times 28$. First convolutional layer (Conv1) has $6$ kernel maps  $\vecwonekone$ with $5\times5$ dimensions and $6$ bias values ($\bonekone$), where $\kpl{1} = 1,2,3...6$. Second convolutional layer (Conv2) has $12$ kernel maps $(\vecwtwoktwo)$ with $5 \times 5$ dimensions and $12$ bias values ($\btwoktwo$), where $\kpl{2} = 1,2,3...12$. Whereas fully connected layer's weight $({\vecW}_{3})$ dimensions is  $10 \times 192$ and the bias $(\myvec{b_{3}})$ is  $10 \times 1$.

\section{Initialisation of the parameters}
We initialise the ${\vecW}_{3}$,${\vecW}_{1}$ and ${\vecW}_{2}$ weight parameters with random numbers and the ${\myvec{b}}_{3}$, ${\myvec{b}}_{1}$ and ${\myvec{b}}_{2}$ biases with zero.

\section{First convolutional layer (Conv1)}
 The output of the first convolutional layer $(\vecoonekone)$ is computed as a convolution between the input image ($\myvec{I}$) and the first layer's kernel maps $\vecwonekone$,

\begin{eqnarray}
\label{eq:apyekcony}
\vecoonekone   & = &  \sigma (\vecvonekone) \nonumber \\
&=& \sigma (\Ivec \otimes {\vecwonekone} + \bonekone)
\end{eqnarray}
\\
where the activation function is

\begin{eqnarray}
\sigma (x) &=& \relu (x) \nonumber \\
   &=& max (x,0)
\end{eqnarray}
 
and $\vecvonekone$ is the weighted vector of first convolutional network before the activation function.

\begin{eqnarray}
\oonekone (i,j)  & = &  \sigma \sum _{u=0}^{4}  \sum _{v=0}^{4} {I}(i-u,j-v) \cdot {\wonekone}(u,v) + \bonekone
\end{eqnarray}
\\
Where $\kpl{1}= 1,2,...6$ because there $6$ kernels in the fist layer, the $\otimes$ denotes the convolution function and $\cdot$ denotes the element-wise multiplication. The size of each $\vecoonekone$ is $24 \times 24$ with zero padding.

\section{\texorpdfstring{First pooling layer $({\myvec{S}}_{1})$} {First pooling layer}}

In this stage we replace each $2 \times 2$ window of the convoluted matrix with the scalar value of the average of the window as in
\begin{eqnarray}
  {S_{1}}_{\kpl{1}}(i,j) &=& \frac{1}{4}  \sum _{u=0}^{1}  \sum _{v=0}^{1} \oonekone (i\times u+i,j \times v+j)
\end{eqnarray}
\\
where $i,j = 1,2,...12$.

\section{Second convolutional layer (Conv2)}
 The output of the second convolutional layer ($\vecotwoktwo$) is computed as a convolution between the $\myvec{S}^{1}_{\kpl{1}}$ and the second layer's kernel maps $\vecwtwoktwo$, as in

\begin{eqnarray}
\vecotwoktwo   & = &  \sigma (\vecvtwoktwo) \nonumber \\
&=& \sigma (\sum ^{6}_{\kpl{1}=1} {{\vecs}_{1}}_{\kpl{1}} \otimes {\vecwtwoktwo} + \btwoktwo)
\end{eqnarray}
\\
The $\vecvtwoktwo$ is the weighted vector of second convolutional network before the activation function.

\begin{eqnarray}
\otwoktwo (i,j)  & = &  \sigma \sum ^{6}_{\kpl{1}=1} \sum _{u=0}^{1}  \sum _{v=0}^{1} S^{1}_{\kpl{1}} (i-u,j-v) \cdot {\wtwoktwo}(u,v) + \btwoktwo)
\end{eqnarray}
\\

Where $\kpl{2}= 1,2,...12$ because there $12$ kernels in the second convolutional layer. The size of each $\vecotwoktwo$ is $8 \times 8$ with zero padding.

\section{\texorpdfstring{Second pooling layer $(S_{2})$}{Second pooling layer}}

In this stage we replace each $2 \times 2$ window of the convoluted matrix with the scalar value of the average of the window as in
\begin{eqnarray}
  S^{2}_{\kpl{2}} &=& \frac{1}{4}  \sum _{u=0}^{1}  \sum _{v=0}^{1} \otwoktwo (i \times u+i,j \times v+j)
\end{eqnarray}

where $i= 1,2,...4$ and $j=1,2,\cdots 4$.

\section{Vectorisation and concatenation}

There are $12$ ${{\vecs}_{2}}_{\kpl{2}}$ and each of them is a $4 \times 4$ matrix. First each ${\vecs_{2}}_{\kpl{2}}$ is vectorised by column scan then all $12$
vectors are concatenated to form a vector with the length of $4 \times 4 \times 12 = 192$. We denote the process of the vectorising and the concatenating as a function f, thus we have

\begin{eqnarray}
\label{eq:vectorisationre}
\vecf &=& F (\{{{ \vecs}_{2}}_{\kpl{2}} \} ) \nonumber \\
F^{-1}(\vecf) &=& \{ {{\vecs}_{2}}_{\kpl{2}} \}
\end{eqnarray}
\\
where $\kpl{2}= 1,2 ,\cdots 12$.

\section{Fully connected layer}
$\haty$ the network output, is the predicted value of its correspondent input, so we have

\begin{eqnarray}
\hat{\vecy} &=& \sigma ({\myvec{V}}_{3})\nonumber \\
\hat {\vecy} &=& \sigma ({\vecW}_{3} \times \vecf + {\vecb}_{3})
\end{eqnarray}

\section{Back propagation}

The network weights and bias parameters are updated using the SGD technique by computing the loss function's gradient with respect to each parameter. We define ${\myL}^{(m)}$ as the MSE loss function of the $m$th training sample as in
\begin{eqnarray}
{\myL}^{(m)} &=& \frac{1}{2}|\myvec{y}^{(m)} -\myvec{\haty}^{(m)}|^{2}
\end{eqnarray}
\\
where $\myvec{y}^{(m)}$ and $\myvec{\haty}^{(m)}$ are the $m$-th label and predicted output respectively.We compute the average loss for all the sample pairs in the training dataset as 

\begin{eqnarray}
{\myL} &=& \frac{1}{2 \mym}|\myvec{y} -\myvec{\haty}|^{2}
\end{eqnarray}
\\
In the BP, we update the parameters backward from the last layer to the first layer, so that we will compute the gradient of each weigh and bias vector of $\nabll{W}$, $\nabll{b}$, $\nabll{W^{2}_{\kl{1},\kl{2}}}$, $\nabll {b^{2}_{k2}}$, $\nabll{W^{1}_{1,k1}}$ then $\nabll{b^{1}_{\kl{1}}}$. Where $\nabla^{\zeta}_{\theta}(a) =\frac {\partial \zeta (a)}{\partial \theta (a)}$.

\section{\texorpdfstring{Loss Gradient with Respect to ${\vecW}_{3}$ ($\nabll{{\vecW}_{3}}$)}{Loss Gradient with Respect to}}
Dimensions of the  $\nabll{{\vecW}_{3}}$ is $10 \times 192$, we have

\begin{eqnarray}
\nabll{W_{3}} (i,j) &=& \frac {\partial \myL}{\partial W_{3} (i,j)} \nonumber \\
  &=& \frac {\partial \myL}{\partial \hat{y}(i)}\cdot\frac {\partial \hat{y}(i)} {\partial \wthree (i,j)}  \nonumber \\
  &=& \mid \hat{y}(i)- y(i) \mid \cdot  \frac {\partial} {\partial \wthree (i,j)} \sigma \bigl ( \sum_{j=1}^{192} \wthree (i,j) \times f^{T}(j) +b(i) \bigr)  \nonumber \\
  &=&  \mid \hat{y}(i)- y(i)\mid \cdot \frac{\partial \sigma {(\vthree (i))}}{\partial \vthree(i)} \cdot f^{T}(j)
\end{eqnarray}
\\
We have
\begin{eqnarray}
 \nabll{\vecwthree} &=& \mid \hat{\vecy} - \vecy \mid \cdot  \frac{\partial \sigma {(\myvec{\vthree})}}{\partial \myvec{\vthree}} \cdot {\vecf}^{T}
\end{eqnarray}

\section{\texorpdfstring{Loss Gradient with Respect to ${\vecb}_{3}$ ($\nabll{{\vecb}_{3}}$)}{Loss Gradient with Respect to}}
Dimensions of the  $\nabll{{\vecb}_{3}}$ is $10 \times 1$, we have

\begin{eqnarray}
  \nabll{{\vecb}_{3}} &=& \frac {\partial \myL}{\partial {\vecb}_{3}} \nonumber \\
  &=& \frac {\partial \myL}{\partial \myvec{\haty}} \cdot \frac {\partial \myvec{\haty}}{\partial \vecvthree} \cdot \frac {\partial \vecvthree}  {\partial {\vecb}_{3}}  \nonumber \\
  &=& \mid \myvec{\haty}- \myvec{y} \mid \cdot  \frac {\partial \sigma (\vecvthree)}{\partial \vecvthree} \cdot\frac {\partial \myvec{V}} {\partial \vecb}  \nonumber \\
  &=&  \mid \myvec{\haty}- \myvec{y} \mid \cdot  \frac {\partial \sigma (\vecvthree)}{\partial \vecvthree}
\end{eqnarray}

\section{\texorpdfstring{Loss Gradient with Respect to $\vecwtwoktwo$ ($\nabll{\vecwtwoktwo}$)}{Loss Gradient with Respect to}}
Dimensions of the  $\nabll{\vecwtwoktwo}$ is $5 \times 5$. Because of concatenation, vectorisation and pooling, we need to compute the BP error on conv2 layer $(\nabll{ \vecotwoktwo})$ before calculating the $\nabll{\vecwtwoktwo}$. Therefore first we compute $\nabll{ \vecf}$ so that we can calculate the $\nabll {{{\vecs}_{2}}_{\kpl{2}}}$ and $\nabll{\vecwtwoktwo}$ accordingly.

\begin{eqnarray}
   \nabll{f} (j) &=& \frac {\partial \myL}{\partial f (j)} \nonumber \\
  &=& \sum _{i=1}^{10} \frac {\partial \myL}{\partial \hat{y}(i)} \cdot \frac {\partial \hat{y}(i)}{\partial \vthree(i)} \cdot \frac{\partial \vthree(i)} {\partial f (j)}  \nonumber \\
  &=& \sum _{i=1}^{10} \mid \hat{y}(i)- y(i) \mid \cdot  \frac {\partial \sigma (\vthree(i))}{\partial \vthree(i)} \cdot \frac {\partial \vthree(i)} {\partial f (j)}
\end{eqnarray}
\\
So we have
\begin{eqnarray}
  \nabla{\vecf}  &=& \vecW \cdot \nabll{ \hat{\vecy}} \cdot \frac {\partial \sigma (\vecvthree)}{\partial \vecvthree} 
\end{eqnarray}
\\
and considering \eq{vectorisationre},  we reshape the 1D error vector $\nabll{\vecf}$ (size $192 \times 1$) by

\begin{eqnarray}
  \vecf &=& F ({\vecs^{2}_{k2}}) \nonumber \\
  F^{-1}(\nabll{ \vecf}) &=& \{ \nabll{ \vecs^{2}_{\kpl{2}}} \} 
\end{eqnarray}
\\
where $ {\kpl{2}=1,2, \cdots 12}$
thus we get 12 error maps on $S_{2}$ layer with the $4 \times 4$ dimensions. Because there is no parameters in the $S_{2}$ layer, we do not need to compute the derivation of second pooling layer. In order to obtain the gradient on the Conv2 layer, we perform up-sampling on $S_{2}$ error maps, so we have

\begin{eqnarray}
\nabll{ \otwoktwo} (i,j) &=& \frac {1}{4} \nabll{ {{\vecs}_{2}}_{\kpl{2}}} (\lceil \frac {i}{2} \rceil , \lceil \frac {j}{2} \rceil) 
\end{eqnarray}
\\
where ${i=1,2,\cdots 8}$ and ${j=1,2,\cdots 8}$ and $\lceil. \rceil$ denotes the ceiling function and the dimensions of $ \nabll{ {{\vecs}_{2}}_{\kpl{2}}}$ and $\nabll{\vecotwoktwo}$ are $ 4\times 4$ and $8 \times 8$ respectively. In this stage as we have already computed $\nabll{ \vecotwoktwo}$ we can finally calculate the $\nabll{\vecwtwoktwo}$ as in

\begin{eqnarray}
\nabll {\wtwoktwo}(u,v) &=&  \frac {\partial \myL}{\partial \wtwoktwo(u,v)} \nonumber \\
&=& \sum_{i=1}^{8} \sum_{j=1}^{8} \frac {\partial \myL}{\partial \otwoktwo(i,j)}\cdot \frac {\partial \otwoktwo(i,j)}{\partial \vtwoktwo (i,j)} \cdot \frac {\partial \vtwoktwo(i,j)}{\partial \wtwoktwo(u,v)} \nonumber \\
&=& \sum_{i=1}^{8} \sum_{j=1}^{8} \nabll{ \otwoktwo}(i,j) \cdot \frac {\partial \sigma(\vtwoktwo(i,j))}{\partial \vtwoktwo (i,j)} \cdot  \frac {\partial \vtwoktwo(i,j)}{\partial \wtwoktwo(u,v)} \nonumber \\
&=& \sum_{i=1}^{8} \sum_{j=1}^{8} \nabll{ \otwoktwo}(i,j) \cdot \frac {\partial \sigma(\vtwoktwo(i,j))}{\partial \vtwoktwo (i,j)} \cdot S^{1}_{\kpl{1}} (i-u,j-v)
\end{eqnarray}
\\
we have
\begin{eqnarray}
\nabll{\myvec{\vtwoktwo}}&=& \nabll{\myvec{\otwoktwo}} \odot {\sigma}^{\prime} {(\myvec{\vtwoktwo)}}
\end{eqnarray}
which is the loss gradient before the activation function on Conv2 layer, we have

\begin{eqnarray}
\vtwoktwo (i,j)&=&  \sum_{\kl{1}=1}^{6} \sum_{u=-2}^{2} \sum_{v=-2}^{2}  S^{1}_{\kpl{1}} (i-u,j-v) \cdot \wtwoktwo (u,v) + \btwoktwo 
\end{eqnarray}
\\
and we know that $ {S_{1}}_{\kpl{1}, rot80} (u-i,v-j)= {S_{1}}_{\kpl{1}} (i-u,j-v)$, so

\begin{eqnarray}
\nabla \wtwoktwo(u,v) &=& \sum_{i=1}^{8} \sum_{j=1}^{8}  {S_{1}}_{\kpl{1}, rot80} (u-i,v-j)\cdot \nabll{ \vtwoktwo} (i,j)
\end{eqnarray}
\\
thus, we have
\begin{eqnarray}
\nabll{ \vecwtwoktwo} &=&  {{\vecs} _{1}}_{\kpl{1}, rot80} \otimes  \nabll{ \vecvtwoktwo}
\end{eqnarray}

\section{\texorpdfstring{Loss Gradient with Respect to $\vecbtwoktwo$ ($\nabll{ \vecbtwoktwo} $)}{Loss Gradient with Respect to}}
The dimensions of $\nabll{ \vecbtwoktwo} $ is $12 \times 1$. We have

\begin{eqnarray}
\nabll{ \btwoktwo} &=&  \frac {\partial \myL}{\partial \btwoktwo} \nonumber \\
&=& \sum_{i=1}^{8} \sum_{j=1}^{8} \frac {\partial \myL}{\partial \otwoktwo(i,j)} \cdot \frac {\partial \otwoktwo(i,j)}{\partial \vtwoktwo (i,j)} \cdot \frac {\partial \vtwoktwo(i,j)}{ \btwoktwo} \nonumber \\
&=& \sum_{i=1}^{8} \sum_{j=1}^{8} \nabll{ \otwoktwo}(i,j) \cdot \frac {\partial \sigma(\vtwoktwo(i,j))}{\partial \vtwoktwo (i,j)} \cdot \frac {\partial \vtwoktwo(i,j)}{ \btwoktwo} \nonumber \\
&=& \sum_{i=1}^{8}\sum_{j=1}^{8}  \nabll{\vtwoktwo}(i,j)
\end{eqnarray}
\\
thus, we have
\begin{eqnarray}
\nabla \vecbtwoktwo &=&\sum_{j=1}^{8} \sum_{j=1}^{8}  \nabll{ \vecvtwoktwo}
\end{eqnarray}
\\
\section{\texorpdfstring{Loss Gradient with Respect to $\vecwonekone$ ($\nabll{ \vecwonekone} $)}{Loss Gradient with Respect to}}
Dimensions of the  $\nabll{ \vecwonekone} $ is $5 \times 5$.  In order to compute the $\nabll{ \vecwonekone} $ first we need to obtain $\nabll{ {{\vecs}_{1}}_{\kl{1}}}$ which is the error on S1 layer then we require to compute the $ \nabll{ \vecoonekone}$ the error in conv1 layer. Therefore, we have 

\begin{eqnarray}
\nabla  {S_{1}}_{\kpl{1}}(i,j) &=&  \frac {\partial \myL}{\partial {S_{1}}_{\kpl{1}}(i,j) } \nonumber \\
&=& \sum_{\kpl{2}=1}^{12}  \sum_{u=0}^{4} \sum_{v=0}^{4}  \frac {\partial \myL}{\partial \vtwoktwo (i+u, j+v)} \cdot \frac {\partial \vtwoktwo (i+u, j+v)}{\partial {S_{1}}_{\kpl{1}}(i,j) } \nonumber \\
&=& \sum_{\kpl{2}=1}^{12}  \sum_{u=0}^{4} \sum_{v=0}^{4} \nabll{ \vtwoktwo} (i+u, j+v) \cdot \frac {\partial}{\partial {S_{1}}_{\kpl{1}}(i,j)}\nonumber \\
&& \bigl( \sum_{\kpl{1}=1}^{6}  \sum_{u=0}^{4} \sum_{v=0}^{4}  {S_{1}}_{\kpl{1}}(i,j) \cdot \wtwoktwo (u,v) + \btwoktwo\bigr) \nonumber \\
&=& \sum_{\kpl{2}=1}^{12}  \sum_{u=0}^{4} \sum_{v=0}^{4} \nabll{ \vtwoktwo} (i+u, j+v) \cdot \wtwoktwo (u,v)
\end{eqnarray}
\\
We know that $ {{W}_{\kpl{1},\kpl{2},rot180 }}_{2} (-u,-v) = \wtwoktwo (u,v)$, therefore we have

\begin{eqnarray}
\nabla  {S_{1}}_{\kpl{1}}(i,j) &=&  \sum_{\kpl{2}=1}^{12}  \sum_{u=-2}^{2} \sum_{v=-2}^{2} \nabll{ \vtwoktwo} (i-(-u),j-(-v)) \nonumber \\
&& \cdot {{W}_{\kpl{1},\kpl{2},rot180 }}_{2} (-u,-v)
\end{eqnarray}
\\
so we have

\begin{eqnarray}
\nabla  \vecs ^{1}_{\kpl{1}} &=&  \sum_{\kpl{2}=1}^{12} \nabll{\vecvtwoktwo} \otimes \vecwtwoktwo _{,rot180}
\end{eqnarray}
\\
In order to obtain the loss gradient with respect to the output of Conv1 layer we need to up-sample the pooling layer's error maps, so

\begin{eqnarray}
\nabll{ \oonekone} (i,j) &=& \frac {1}{4} \nabll{ {{\vecs}_{1}}_{\kpl{1}}} (\lceil \frac {i}{2} \rceil , \lceil \frac {j}{2} \rceil)
\end{eqnarray}
\\
Where ${i=1,2,...24}$ and ${j=1,2,...24}$, we can compute the $\nabll{ \vecwonekone} $, therefore

\begin{eqnarray}
\nabll{ \wonekone}(u,v) &=&  \frac {\partial \myL}{\partial \wonekone(u,v)} \nonumber \\
&=& \sum_{i=1}^{24} \sum_{j=1}^{24} \frac {\partial \myL}{\partial \oonekone(i,j)} \cdot \frac {\partial \oonekone(i,j)}{\partial \vonekone(i,j)} \cdot \frac {\partial \vonekone(i,j)}{\partial \wonekone(u,v)} \nonumber \\
&=& \sum_{i=1}^{24} \sum_{j=1}^{24} \nabll{ \oonekone} (i,j) \cdot \frac {\partial \sigma(\vonekone (i,j))}{\partial \vonekone (i,j)} \cdot \frac {\partial \vonekone(i,j)}{\partial \wonekone(u,v)}  \nonumber \\
&=& \sum_{i=1}^{24} \sum_{j=1}^{24} \nabll{ \oonekone}(i,j) \cdot  \frac {\partial \sigma(\vonekone (i,j))}{\partial \vonekone (i,j)} \cdot \Ivec (i-u,j-v)
\end{eqnarray}
\\
We rotate $I$, 180 degrees and we know that

\begin{eqnarray}
\nabll{ \vonekone} (i,j) &=& \nabll{ \oonekone}(i,j) \cdot  \frac {\partial \sigma(\vonekone (i,j))}{\partial \vonekone (i,j)}
\end{eqnarray}
\\
thus,

\begin{eqnarray}
\nabll{ \wonekone}  (u,v) &=&   \sum_{i=1}^{24} \sum_{j=1}^{24} I_{rot180} (u-i,v-j) \cdot \nabll{ \vonekone} (i,j)
\end{eqnarray}
\\
so we have

\begin{eqnarray}
\nabll{\myvec{ \wonekone}} &=& {\Ivec}_{rot180} \otimes \nabll{ \myvec{\vonekone}}
\end{eqnarray}
\\

\section{\texorpdfstring{Loss Gradient with Respect to $\vecbonekone$ ($\nabll{ \vecbonekone} $)}{Loss Gradient with Respect to}}
The dimensions of $\nabll{ \vecbonekone} $ is $6 \times 1$. We have

\begin{eqnarray}
\nabll{ \bonekone} &=&  \frac {\partial \myL}{\partial \bonekone} \nonumber \\
&=& \sum_{i=1}^{24} \sum_{j=1}^{24} \frac {\partial \myL}{\partial \oonekone(i,j)} \cdot \frac {\partial \oonekone(i,j)}{\partial \vonekone (i,j)} \cdot \frac{\partial \vonekone (i,j)}{\partial \bonekone} \nonumber \\
&=& \sum_{i=1}^{24} \sum_{j=1}^{24} \nabll{ \oonekone}(i,j) \cdot \frac {\partial \sigma (\vonekone(i,j))}{\partial \vonekone(i,j)} \cdot \frac {\partial \vonekone (i,j)}{\partial \bonekone}  \nonumber \\
&=& \sum_{i=1}^{24} \sum_{j=1}^{24} \nabll{ \oonekone}(i,j) \cdot \frac {\partial \sigma (\vonekone(i,j))}{\partial \vonekone(i,j)}\nonumber \\
&=& \sum_{i=1}^{24} \sum_{j=1}^{24} \nabll{ \vonekone}(i,j)
\end{eqnarray}
\\
thus, we have
\begin{eqnarray}
\nabll{ \vecbonekone} &=& \sum_{i=1}^{24} \sum_{j=1}^{24} \nabll{ \vecvonekone}
\end{eqnarray}

\section{Parameter update}

Following computing the partial derivative of loss function with respect to weights and bias in each layer, we can update the parameters after each iteration, we need to set the value of the learning rate ($\lr$) so we can update parameters accordingly as in

\begin{eqnarray}
\vecwthree [t+1] & = & \vecwthree [t] + \Delta \vecwthree [t] \nonumber \\
{\vecb}_{3} [t+1] & = & {\vecb}_{3} [t] + \Delta {\vecb}_{3} [t] \nonumber \\
\vecwonekone [t+1] & = &  \vecwonekone[t] + \Delta \vecwonekone[t] \nonumber \\
\vecbonekone [t+1] & = & \vecbonekone[t] + \Delta \vecbonekone[t] \nonumber \\
\vecwtwoktwo [t+1] & = & \vecwtwoktwo [t]+ \Delta \vecwtwoktwo [t] \nonumber \\
\vecbtwoktwo [t+1] & = & \vecbtwoktwo [t]+ \Delta \vecbtwoktwo [t]
\end{eqnarray}

where $[t+1]$ and $[t]$ denote the iteration numbers.

\begin{eqnarray}
 \Delta \vecwthree [t] &=& - \lr \nabll{ \vecwthree} [t] \nonumber \\
 \Delta {\vecb}_{3} [t] & =& - \lr \nabll{ {\vecb}_{3}} [t] \nonumber \\
 \Delta \vecwonekone[t] &=& - \lr \nabll{ \vecwonekone}[t] \nonumber \\
 \Delta \vecbonekone[t] &=& - \lr \nabll{\vecbonekone}[t] \nonumber \\
 \Delta \vecwtwoktwo [t]& =& - \lr \nabll{ \vecwtwoktwo} [t] \nonumber \\
 \Delta \vecbtwoktwo [t]& =& - \lr \nabll{ \vecbtwoktwo} [t]
\end{eqnarray}

\chapter{Complex differentiability}
\label{cha:ytwoappcomplex}

\subsubsection{Holomorphism and Couchy - Riemann equations} 
Holomorphism also is called analyticity, ensures that a complex-valued function is complex differentiable in the neighbourhood of every point in its domain. This means that the derivative $ f^{\prime} (z)\equiv \lim_{\Delta z \rightarrow 0}[\frac{f(z) + \Delta z) - f(z)}{\Delta z}]$ of $f$ exists at every point $z$ of the domain of complex-valued function $f$ of complex variable $z = r + \jmath q$ thus, $f(z) = u(r,q) + \jmath v(r,q)$. $u$ and $v$ are real-valued functions. We can express $\Delta z = \Delta r + \jmath \Delta q$, $\Delta z$ can approach zero along the real axis, imaginary or in-between, however for a complex function to be complex differentiable $ f^{\prime} (z)=\frac{\partial f}{\partial z}$ must be the same complex value regardless of the direction of approach. When $\Delta z$ approaches zero along the real axis, $ f^{\prime} (z)$ will be calculated as
\begin{eqnarray}
\label{eq:couchyek}
 f^{\prime} (z) & = & [ \frac{(f(z) + \Delta z) - f(z)}{\Delta z} ]\nonumber \\
 & = & \lim_{\Delta r \rightarrow 0}\lim_{\Delta q \rightarrow 0}[\frac{\Delta u (r,q) + \jmath \Delta v(r,q)}{\Delta r + \jmath \Delta q}]\nonumber \\
& = & \lim_{\Delta r \rightarrow 0}[\frac{\Delta u (r,q) + \jmath \Delta v(r,q)}{\Delta r + \jmath 0}]
\end{eqnarray}

when $\Delta z$ approaches $0$ along the imaginary axis, $ f^{\prime} (z)$ will be calculated as
\begin{eqnarray}
\label{eq:couchydo}
 f ^{\prime} (z) & = & \lim_{\Delta r \rightarrow 0}\lim_{\Delta q \rightarrow 0}[\frac{\Delta u (r,q) + \jmath \Delta v(r,q)}{\Delta r + \jmath \Delta q}]\nonumber \\
& = & \lim_{\Delta q \rightarrow 0}[\frac{\Delta u (r,q) + \jmath \Delta v(r,q)}{0 + \jmath \Delta q}]
\end{eqnarray}

\eq{couchyek} and \eq{couchydo} are equivalent so we have $\frac {\partial f}{\partial z} = \frac {\partial u}{\partial r} + \jmath \frac {\partial v}{\partial r} = - \jmath \frac {\partial u}{\partial q} + \frac {\partial v}{\partial q}$. In order for $f$ to be complex differentiable, it must satisfy both these conditions of first: $\frac {\partial u}{\partial r} = \frac{\partial v}{\partial q}$  and second:  $\frac {\partial u}{\partial q} = - \frac {\partial v}{\partial r}$, which are called Couchy - Riemann equations.
\cite{deepcvNW} holomorphic functions can be leveraged for computational efficiency purposes. Using holomorphic functions allows us to share gradient values, so instead of computing and back propagating $4$ different gradients, only $2$ are required. However \cite{hirose} \cite{arjovsky} \cite{danihelka} and \cite{wisdom} used non-holomorphic activation functions and optimised their networks.

\chapter{Real-value Dataset}

\section{Real-value Dataset}
\cite{complexnn} designed and implemented a complex-forward residual network and they only test the performance of the network with real-value datasets. In order to create a comparison we experiment training all three fully complex CNN, complex-forward residual and the complex-forward CNN with two class of Cifar-$10$. We copy the real component to create the imaginary component for the real-value Cifar-$10$ dataset, therefore  the sample dimension is doubles from$32\times32\times3$.  

\subsection{\texorpdfstring{Cifar-$10$} class dataset}

The Canadian Institute for Advanced Research (Cifar) is a Canadian-based global research organisation that brings together teams of top researchers from around the world to address important and complex questions. The Cifar-$10$ dataset is one of the most widely used datasets for machine learning research~\cite{ciften}. Cifar-$10$ is currently one of the most widely used datasets in machine learning and serves as a test ground for many computer vision methods. A concrete measure of popularity is the fact that Cifar-10 was the second most common dataset in NIPS 2017 (after MNIST) ~\cite{cifhamner}~\cite{ciften}.

The CIFAR-$10$ dataset consists of $60000$ $32\times32\times 3$colour images in $10$ classes, with $6000$ images per class. There are $50000$ training images and $10000$ test images.

The dataset is divided into five training batches and one test batch, each with $10000$ images. The test batch contains exactly $1000$ randomly-selected images from each class. The training batches contain the remaining images in random order, but some training batches may contain more images from one class than another. Between them, the training batches contain exactly $5000$ images from each class.

\begin{figure}
\centering\includegraphics[scale=0.8]{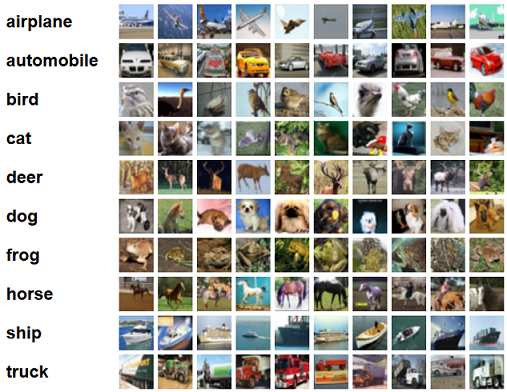}
\caption{Cifar-10 classes samples}
\label{fig:cifar}
\end{figure}

The Cifar-$10$ dataset consists of $10$ classes of airplane, automobile, bird, cat, deer, dog, frog, horse, ship and truck~\fig{cifar}

We have used $2$ random classes of the Cifar-$10$ dataset, so that we have got $10000$ training and $2000$ test samples for the binary classification experiments. ~\fig{cifar} illustrates $10$ sample images of each class.

\subsection{Our Real-value Dataset}

We have randomly selected $2$ classes of Cifar-10 datasets as our real-value dataset, which consist of $10000$ training and $2000$ test samples in total. In order to test our complex network architectures we have added a copy of the real-value samples to create the imaginary value for each sample, therefore we create a new complex-value alike test and training datasets that each sample dimensions are $32\times32\times6$   instead of $60000$ $32\times32$ in the original Cifar-10 dataset.

However, \cite{complexnn} that creates the imaginary part for each sample by applying two layer of batch normalisation, activation function and a $2$D convolution on each sample of the dataset and apply the imaginary part alongside the real part of each sample to the residual network with complex blocks.

\section{Real-value Dataset Results}

\subsection{Complex-forward Residual Network}
This section displays some experiment result of training the complex-forward residual network with $2$ classes of Cifar-$10$.
\begin{table}
\centering\includegraphics[scale=0.9]{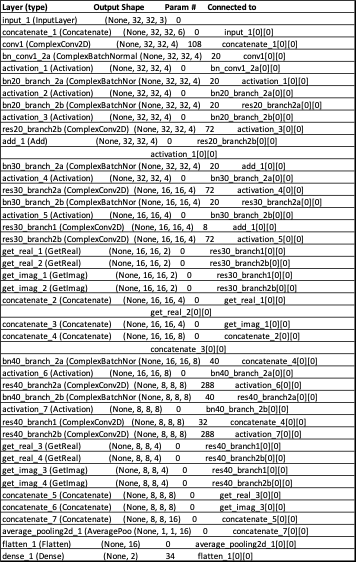}
\caption{Cifar$10$ dataset Complex-forward Residual architecture  }
\label{tbl:cifarresone}
\end{table}

\begin{table}
\centering\includegraphics[scale=0.9]{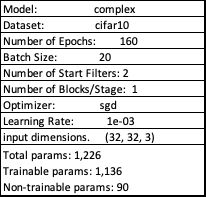}
\caption{Cifar-$10$ dataset Complex-forward Residual parameters setting }
\label{tbl:cifarrestwo}
\end{table}

\begin{figure}
\centering\includegraphics[scale=0.9]{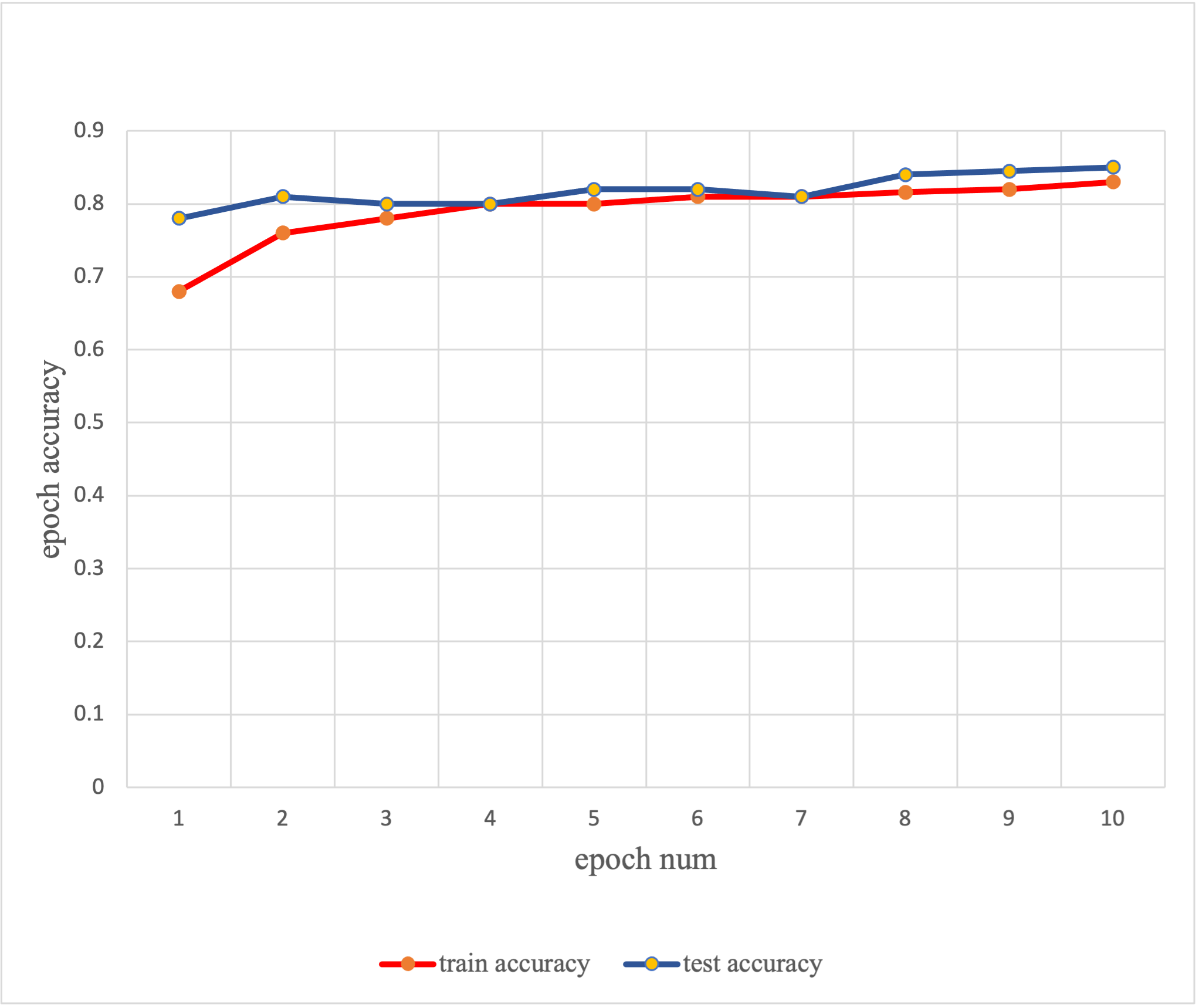}
\caption{Cifar-$10$ dataset Complex-forward Residual accuracy per epoch (for Figure~\figg{cifarresone}, \figg{cifarrestwo} setting)  }
\label{fig:cifarresthree}
\end{figure}

\subsection{Complex-forward CNN Network}

This section displays some experiment result of training the complex-forward CNN network with $2$ classes of Cifar-$10$.
\begin{table}
\centering\includegraphics[scale=0.9]{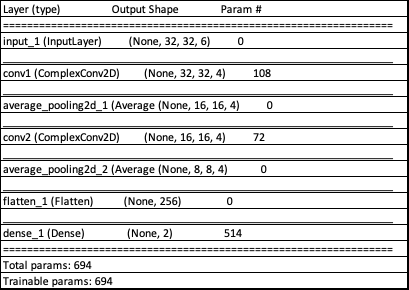}
\caption{Cifar$10$ dataset Complex-forward CNN architecture  }
\label{tbl:cifarconvone}
\end{table}

\begin{table}
\centering\includegraphics[scale=0.9]{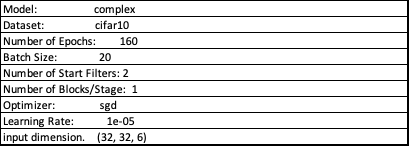}
\caption{Cifar-$10$ dataset Complex-forward CNN parameters setting }
\label{tbl:cifarconvtwo}
\end{table}

\begin{figure}
\centering\includegraphics[scale=0.9]{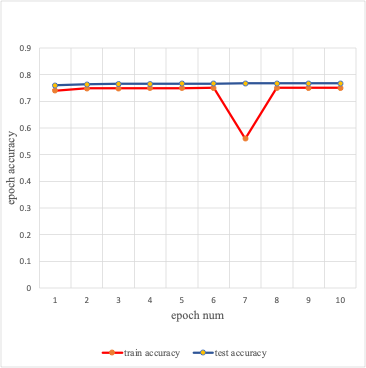}
\caption{Cifar-$10$ dataset Complex-forward CNN accuracy per epoch (for Figure~\figg{cifarconvone}, \figg{cifarconvtwo} setting)  }
\label{fig:cifarconvthree}
\end{figure}

\subsection{Complex CNN Network}

This section displays some experiment result of training the fully complex CNN network with $2$ classes of Cifar-$10$.

\begin{table}
\centering\includegraphics[scale=0.9]{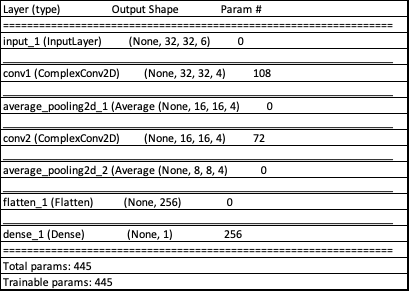}
\caption{Cifar$10$ dataset Complex CNN architecture  }
\label{tbl:cifarmyone}
\end{table}

\begin{table}
\centering\includegraphics[scale=0.9]{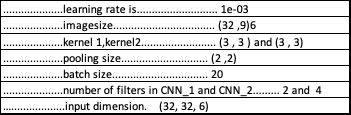}
\caption{Cifar-$10$ dataset Complex CNN parameters setting }
\label{tbl:cifarmytwo}
\end{table}

\begin{figure}
\centering\includegraphics[scale=0.9]{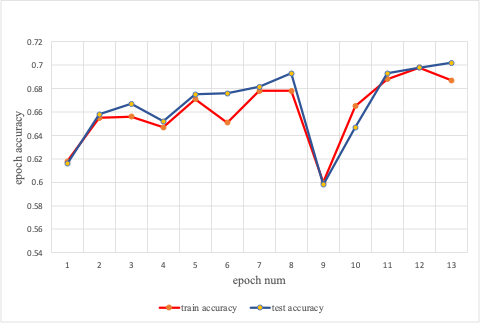}
\caption{Cifar-$10$ dataset Complex CNN accuracy per epoch (for Figure~\figg{cifarmyone}, \figg{cifarmytwo} setting)  }
\label{fig:cifarmythree}
\end{figure}

\chapter{Complex BN and Weight Initialisation}
 This Appendix displays the detail of complex batch normalisation and complex weight initialisation by \cite{complexnn}, which we utilised in our residual complex forward experiments.

\begin{figure}
\centering\includegraphics[scale=0.7]{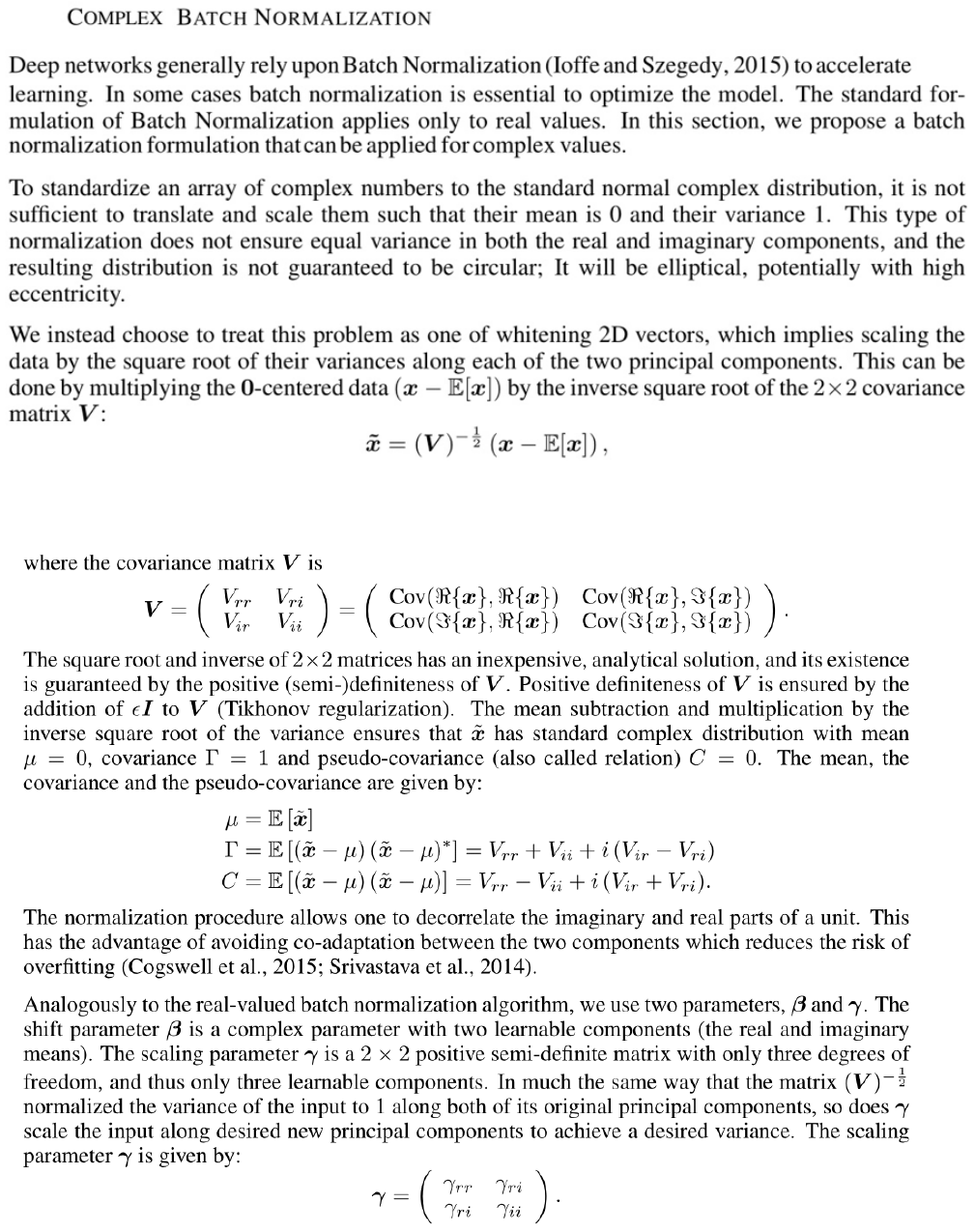}
\caption{  }
\label{fig:apppiceman}
\end{figure}

\begin{figure}
\centering\includegraphics[scale=0.7]{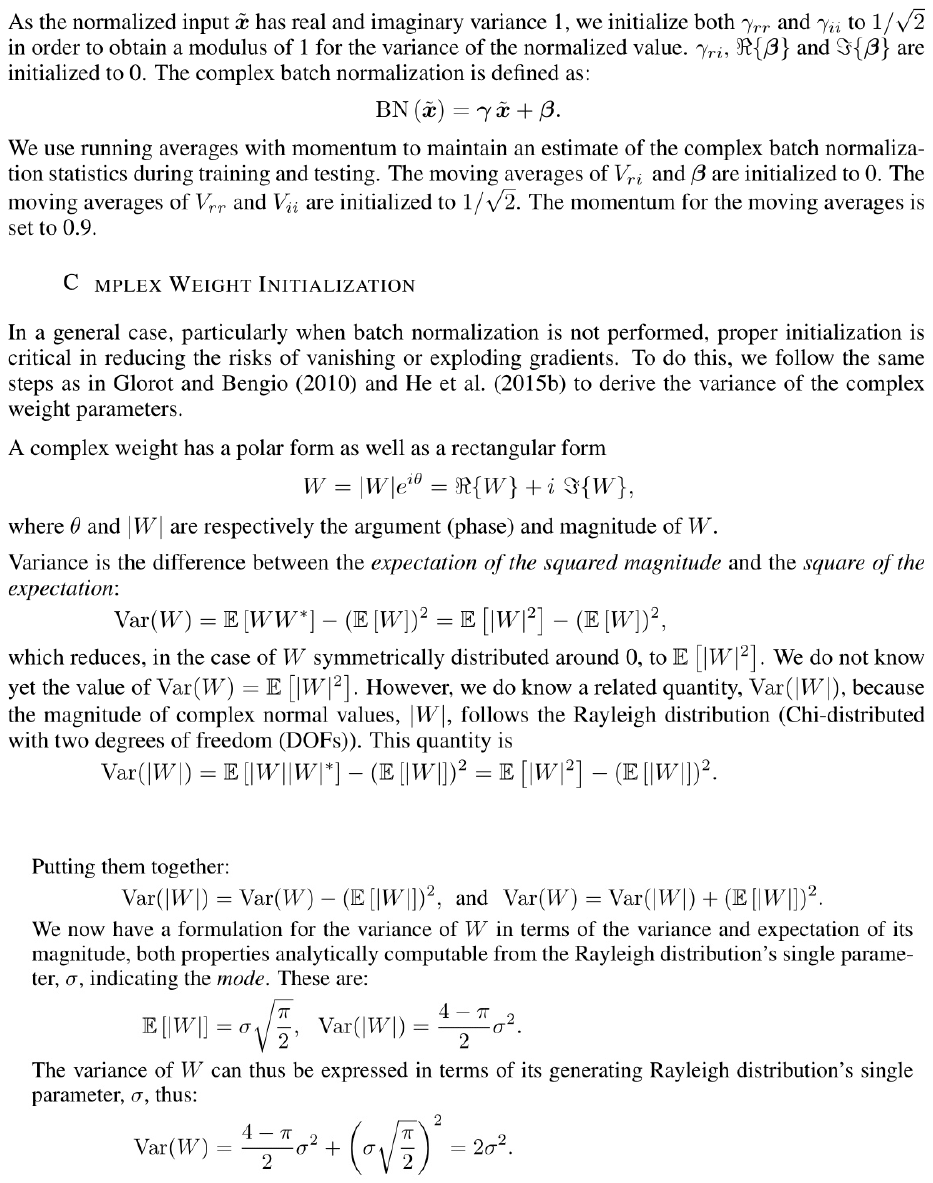}
\caption{  }
\label{fig:mapman}                                                                                                                                                                   
\end{figure}

\begin{figure}
\centering\includegraphics[scale=0.7]{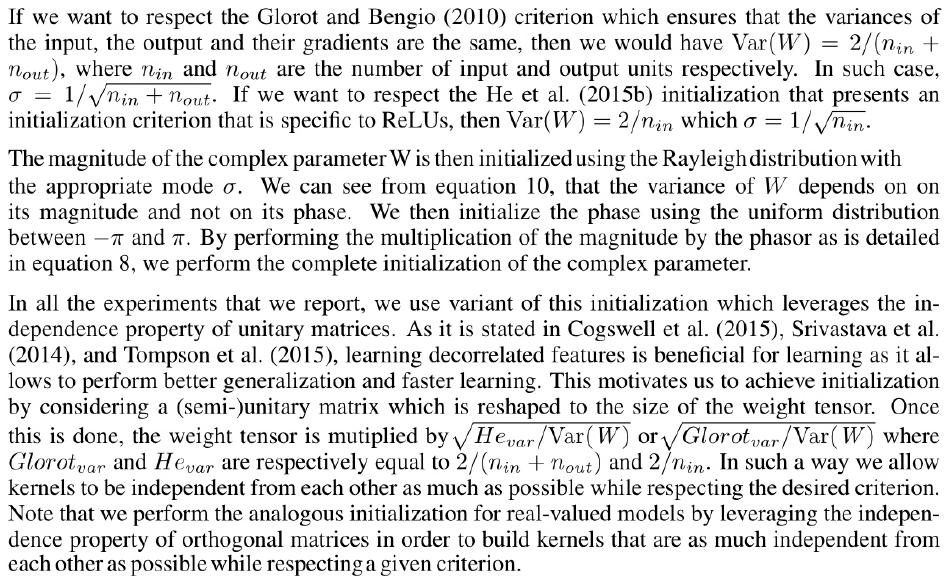}
\caption{       }                                                                                                                                                             
\end{figure}

\end{document}